\documentclass{article}

    \PassOptionsToPackage{numbers, compress}{natbib}

 \usepackage[eandd, final]{neurips_2026}

\usepackage[utf8]{inputenc} 
\usepackage[T1]{fontenc}    
\usepackage{hyperref}       
\usepackage{url}            
\usepackage{booktabs}       
\usepackage{amsfonts}       
\usepackage{nicefrac}       
\usepackage{microtype}      
\usepackage{xcolor}         

\usepackage{array,multirow}
\usepackage{graphicx}
\usepackage{enumitem}
\usepackage{bbm}
\usepackage{soul}
\usepackage{booktabs}
\usepackage{subcaption}
\usepackage[utf8]{inputenc}
\usepackage{amsmath}
\usepackage{listings}
\usepackage{xcolor}

\definecolor{codegreen}{rgb}{0,0.6,0}
\definecolor{codegray}{rgb}{0.5,0.5,0.5}
\definecolor{codepurple}{rgb}{0.58,0,0.82}
\definecolor{backcolour}{rgb}{0.95,0.95,0.92}
\title{

An Assessment of Human vs. Model Uncertainty in Soft-Label Learning and Calibration} 

\author{
  Maja Pavlovic \\
  Queen Mary University London \\
  \texttt{m.pavlovic@qmul.ac.uk} \\
  \And
  Silviu Paun\thanks{Work done outside of Amazon.}   \\
  Amazon \\ 
  \texttt{spaun3691@gmail.com} \\
  \And
  Massimo Poesio \\
  Queen Mary University London - University of Utrecht\\
  \texttt{m.poesio@\{qmul.ac.uk,uu.nl\} } \\
}

\begin{document}

\maketitle

\begin{abstract}
    Central to human-aligned AI is understanding the benefits of human-elicited labels over synthetic alternatives. While human soft-labels improve calibration by capturing uncertainty, prior studies conflate these benefits with the implicit correction of mislabeled data (mode shifts), obscuring true effects of soft-labels. We present a controlled audit of soft-label learning across MNIST and a synthetic variant, re-annotating subsets to extract human uncertainty. By decoupling soft-label supervision from underlying label mode shifts, we show that while human soft-labels do provide accuracy gains, their larger value lies in acting as a regularizer that improves model calibration on difficult samples and promotes stable convergence across training runs. Dataset cartography reveals models trained on human soft-labels mirror human uncertainty, whereas those trained on synthetic labels fail to align with humans. Broadly, this work provides a diagnostic testbed for human-AI uncertainty alignment. Dataset 
    available at \href{https://huggingface.co/datasets/majapavlo/image-level-soft-label-digits}{HuggingFace}.\footnote[1]{  \href{https://huggingface.co/datasets/majapavlo/image-level-soft-label-digits}{https://huggingface.co/datasets/majapavlo/image-level-soft-label-digits}}
\end{abstract}

\section{Introduction }
Model calibration ensures that predicted probabilities accurately reflect the true likelihood of outcomes. Traditionally, models are trained and evaluated using one-hot encoded vectors. Within this paradigm, smaller models have been shown to produce well-calibrated predictions \cite{niculescu2005predicting}; however, as model complexity increases, models often become less calibrated \cite{guo_2017_on_calibration, achiam2023gpt}. While early approaches demonstrate that soft-labels enhance calibration \cite{muller2019does} and generalization \cite{hinton2015distilling}, recent work suggests that synthetically generated data can further refine calibration by exposing models to a wider variety of edge cases \cite{thulasidasan2019mixup, bouniot2025tailoring, ba2024fillcalibration}. 
However, using teacher model soft-labels \cite{hinton2015distilling} or otherwise algorithmically generated soft-labels risks propagating synthetic biases rather than reflecting the nuances of human uncertainty. Although such targets provide a softer signal than one-hot vectors, recent analyses find why algorithmic soft-labels can actually harm uncertainty ranking \cite{zhu2022rethinking,xia2024towards}. 

In contrast to these synthetic soft-labels, human-elicited soft-labels encode the diverse origins of uncertainty that characterize real-world data and have been shown to directly improve both model calibration and robustness \cite{peterson_2019_iccv, collins_eliciting_2022}. Human  uncertainty is multifaceted, originating from varied interpretations of annotation schemes \cite{parrish_2024_picture_bird, uma2021learning}, differences in cultural or background knowledge \cite{sanders2022ambiguous, parrish_2024_picture_bird}, cognitive difficulties in separating similar concepts \cite{collins_human_2023, uma2021learning}, or data-specific issues like image occlusion and noise \cite{peterson_2019_iccv, schwirten2024ambiguous, collins_human_2023}.
Additionally, difficulty arises when instances legitimately belong to multiple categories \cite{basile_2021_need, sanders2022ambiguous}, or are inherently vague, allowing for multiple valid interpretations due to ambiguity rather than overlapping categories \cite{liu_2023_afraid, schwirten2024ambiguous}.
Whereas classic approaches treat conflicting annotations as noise for which a single, recoverable ground truth can be found \cite{dawidskene1979, natarajan2013learning}, recent efforts have begun integrating individual human uncertainty into annotated datasets \cite{aroyo_dices_2023, parrish_2024_picture_bird, collins_eliciting_2022}.

While the utility of soft-labels is thought to depend on their alignment with human label variation \cite{plank_2022_problem, falk2025mining, uma2021learning}, our fundamental understanding of how and why human-derived soft-labels influence model training dynamics remains limited. 
Prior studies demonstrating accuracy gains from human soft-labels \cite{peterson_2019_iccv} conflate the soft-supervision with the implicit correction of the underlying label mode.
Collecting multiple annotations per instance can shift the label mode ($\operatorname{argmax}$) from a wrong or somewhat likely class to a more agreed upon one. Several works have shown that common benchmarks contain label errors in test sets \cite{northcutt2021pervasive, muller2019identifying, bauerle2020classifier}. Thus, reported accuracy gains in soft-training may simply reflect an improvement in majority labels rather than the benefits of learning from soft-labels.
Decoupling these effects requires a controlled testbed, akin to the use of MNIST to validate hypotheses in generalization \cite{zhang2016understanding} or challenge assumptions regarding adversarial robustness \cite{szegedy2013intriguing}.

To provide a similarly principled investigation into human soft-labels, we conduct experiments on two digit datasets (MNIST and a synthetic equivalent), re-annotating subsets to explicitly capture individual human class uncertainty.  Within these testbeds, we restrict model capacity across three tractable baselines to capture fundamental learning differences. By evaluating these models across strata of human agreement and comparing soft-labels to standard training paradigms, we offer the following contributions: 

\begin{itemize}[itemsep=1.pt, topsep=0pt]
    \item \textbf{An audit of two fundamental datasets with respect to human-centric soft-labels:} 
    Specifically, we show that simply re-labeling the data contributes to a substantial accuracy increase.
    Furthermore, re-annotation of the synthetic digits reveals that model-generated labels in our testbed fail to align with human visual intuition across much of the dataset; consequently, we observe that models easily fit these synthetic labels but struggle to model human uncertainty.
    \item \textbf{Refining the soft-label advantage:}  We provide a thorough evaluation of human soft-labels by disentangling their benefits from mode shifts during re-labeling; a control for isolating true effects absent in previous evaluations. While we confirm accuracy gains, we show that the {benefit} of human uncertainty lies in improving model calibration {particularly across difficult samples} and facilitating more stable convergence across runs. We further corroborate that these gains hold even when using an order of magnitude fewer labels per image than prior work.
    \item \textbf{Aligning training dynamics with human uncertainty:} We find that models supervised by human soft-labels mirror human uncertainty throughout the learning process; leading to models whose predictions are more aligned with human-centric uncertainty than those trained on standard one-hot labels, and notably more than those trained on synthetic labels.
\end{itemize}

\section{Dataset \& annotation scheme}
\label{sec:dataset_annotation_scheme}

\subsection{Baseline datasets} 
By deliberately using simple, well-defined digit datasets, we create a controlled environment to evaluate how human soft-labels influence model performance without the confounding variables of natural imagery. MNIST \cite{lecun2002gradient} and Ambiguous-MNIST \cite{mukhoti_2023_ddu} are our two baseline datasets for further annotation of human uncertainty as they preserve a natural distribution across the difficulty spectrum, central to our interest in ambiguous/hard samples. 
Our intent is not to subvert the original labels, but to augment them by capturing human perceptual uncertainty on intrinsically ambiguous edge cases. 
For further details on the dataset selection, deduplication, and construction see Appendix \ref{appendix_data_collection}. In MNIST the images and labels were directly extracted via the writing process itself. 
In Ambiguous-MNIST, images are generated by decoding a linear combination of latent representations from two different digits using a pre-trained Variational Autoencoder \cite{mukhoti_2023_ddu}. Each image is assigned a label by sampling from the softmax probabilities of a pre-existing MNIST model, then filtered by mutual information and stratified by entropy to ensure a diverse and high-quality range of ambiguity \cite{mukhoti_2023_ddu}. Prior to sampling, we deduplicated all candidate datasets: while MNIST had 0 duplicates, Ambiguous-MNIST contained 1,392 duplicates, leaving 10,608 unique images (see Appendix \ref{appendix_data_collection} for more details \& other candidates).

\subsection{Annotation scheme and human judgments}
\textbf{Annotation scheme: } To effectively capture human uncertainty, we require both a categorical label and a reliable confidence judgment from annotators. Prior efforts in machine learning have generated soft-labels either by aggregating single categorical judgments across multiple annotators \cite{peterson_2019_iccv} or by explicitly asking annotators to provide numerical likelihood estimates for a given label \cite{sanders2022ambiguous, collins_eliciting_2022, collins2023human_mixup}. However, cognitive science literature demonstrates that humans are poorly calibrated when it comes to providing numerical estimates \cite{fleming2024metacognition, keren1987facing}. For instance, whilst humans show little general overconfidence on two-choice questions, they exhibit pronounced overconfidence when asked to produce subjective numerical confidence intervals \cite{klayman1999overconfidence}.

Recent work suggests that allowing multiple label selections to capture uncertainty achieves results close to those capturing single judgments \cite{collins_eliciting_2022}. However, survey methodology research finds that standard "select-all-that-apply" interfaces are vulnerable to satisficing: respondents tend to pick the first plausible option and fail to deeply consider subsequent choices \cite{eckman_2024_survey, galesic2008eye}. Conversely, prompting annotators with a forced-choice "yes/no" format for each individual class encourages them to process every option separately \cite{eckman_2024_survey}. 
Recent studies on data collection in machine learning demonstrate its practical superiority. "Yes/no" approaches yield higher inter-rater reliability and are significantly more sensitive at detecting subtle flaws than the Likert scales \cite{mallinar2026scalable} traditionally used for granular uncertainty feedback \cite{maladry2024likert}.

Motivated by these findings, we adopt a forced-choice yes/no approach over "select-all-that-apply" formats or Likert scales. To accurately capture human uncertainty without forcing arbitrary numerical guesses, we augment this binary choice with an explicit "unsure" option, aligning with recent best practices in data collection \cite{parrish_2024_picture_bird, aroyo_dices_2023}. Providing an "unsure" option does not encourage satisficing; rather, it provides highly informative signals about intrinsically difficult or ambiguous instances where models also tend to struggle \cite{eckman_2024_survey}. Unlike choosing a multiple-choice approach and collecting an overarching confidence score, this setup isolates annotator uncertainty per class, providing highly granular feedback.
Finally, we omit examples of "properly" labeled uncertainty from our instructions to prevent instruction bias. Over-reliance on examples artificially constrains data diversity and overestimates downstream model performance \cite{parmar2023don}; withholding them ensures we collect distinct annotator behavior \cite{eckman_2024_survey}.

\textbf{Human judgments:} We recruited $N=480$ unique annotators via Prolific \cite{palan2018prolific}, with each participant completing an average workload of approximately 70 images. The annotation task was developed and hosted using the Gorilla Experiment Builder \cite{anwyl2020gorilla}. For each image, annotators performed a multi-class categorization, specifying "Yes", "No" or "Unsure" for each digit class ($0, \dots, 9$). To ensure data quality, we embedded three \textit{gold standard} digit attention checks; any participant failing a single check was excluded from the final dataset. This resulted in a detailed annotation set with an average of 6 annotations per image (further details in Appendix \ref{appendix_annotation}).

\subsection{Aggregation to image-level}
\label{aggregation_method}
To consolidate individual human judgments into image-level representations, we derive both soft-labels and uncertainty metrics from the raw "Yes/No/Unsure" annotations. We define two weighting schemes for these aggregated soft-labels:

\textbf{Equally-weighted ($soft_e$):} Both "Yes" and "Unsure" selections receive a weight of 1 to form each annotator's individual probability distribution. These individual distributions are then averaged to create the final image-level soft-label.

\textbf{Uncertainty-weighted ($soft_w$):} "Unsure" selections are down-weighted to 0.5 at the annotator level to reflect a lower class-likelihood, while "Yes" remains 1, before averaging the distributions.

If an annotator rejects all ten digits, the sample is assigned to a "NaN" class (non-digit).
To quantify collective human doubt for a given sample, we use two image-level uncertainty proxies:

\textbf{Mean uncertainty ($u^{mean}$):} The average magnitude of doubt across all annotators for an image.

\textbf{Unsure proportion ($u^{prop}$):} The fraction of annotators who expressed uncertainty at least once for a given image, capturing the consensus of doubt.

For full definitions and mathematical formulations of the aggregation, refer to Appendix \ref{appendix_annotation}.

\subsection{Dataset statistics}
Following \cite{plank_2022_problem}, we explore the prevalence of human label variation (HLV) by partitioning the data into HLV and NoHLV samples, where the latter represents unanimous annotator agreement (i.e. no human label variation). As shown in Table \ref{tab:dataset_stats}, there is a stark divergence in label complexity between the two datasets. While MNIST remains highly consistent with 60.37\% NoHLV samples and only has about 3\% change in mode (i.e. original label changes), Mukhoti is defined by pronounced human label variation. Concretely, 80.52\% of Mukhoti samples show HLV, and about 33\% of the samples change mode (the majority label) under human re-annotation. This aligns with prior preliminary observations in \cite{collins2023human_mixup}, which show that human annotators do not perceive or label mixup-generated images according to their mathematical mixing ratios. Our re-annotation results extend this evidence by showing that model-based methods for generating synthetic target-distributions can also differ from human visual perception, reinforcing preliminary observations from \cite{collins2023human_mixup, peterson_2019_iccv}. Quantitatively, the mean uncertainty ($\mu_{u^{\text{mean}}}=0.076$) and proportional uncertainty ($\mu_{u^{\text{prop}}}=0.418$) are $\sim$3x that of  MNIST, justifying its use as a benchmark for high-ambiguity cases. Further details are provided in Appendix \ref{appendix_dataset_stats}.

\begin{table}[h]
\vspace{-0.5em}
\centering
\small
\caption{Comparative statistics for re-annotated MNIST and Mukhoti datasets. We report the prevalence of human label variation (HLV), the mode change from original ground-truth labels, Krippendorf's $\alpha$, and mean uncertainty metrics ($\mu_{u^{\text{mean}}}$, $\mu_{u^{\text{prop}}}$).
}
\label{tab:dataset_stats}
\vspace{0.4em}
\renewcommand{\arraystretch}{.8} 
\begin{tabular}{lccccccc}
\toprule
{Dataset} & {\textit{NoHLV} (\%)} & {\textit{HLV} (\%)} & Mode change (\%) & { Krippendorf's} $\alpha$ & {$\mathbf{\mu_{u^{mean}}}$} & {$\mathbf{\mu_{u^{prop}}}$}   \\
\midrule
MNIST   & 60.37 & 39.63 & $\sim$3  & 0.973  & 0.025 & 0.142 \\
Mukhoti & 19.48 & 80.52 & $\sim$33 &  0.835 & 0.076 & 0.418 \\
\bottomrule
\end{tabular}
\end{table}

\section{Analysis }
\label{sec:dataset_analysis}

\subsection{Learning from human-centric soft-targets}
Our analysis centers on investigating the impact of different target encodings on model accuracy, calibration, and training dynamics. We use the re-annotated MNIST and Mukhoti subsets to explore how models behave in the presence of human-centric soft-labels versus a lack thereof across four label regimes:

\begin{itemize}
    \item \textbf{Original} ($orig.$) and \textbf{synthetic} ($synth.$) \textbf{labels}: one-hot labels for MNIST and the model-generated label-distributions in Mukhoti
    \item \textbf{Equally-weighted} ($soft_e$) and \textbf{uncertainty-weighted} ($soft_w$) \textbf{labels}: human soft-labels as described in Section \ref{aggregation_method}
    \item \textbf{New majority voted labels} ($maj._n$): One-hot version of the soft-labels obtained via majority voting. This provides a controlled comparison that preserves the modal class, allowing us to isolate the specific utility of soft-labels.
\end{itemize}

\textbf{Training protocols:} 
To prevent the memorization characteristic of large regimes from obscuring fundamental learning differences, we deliberately restrict model capacity, allowing us to 
measure the effects of human uncertainty across three tractable baselines:
a Simple FFN (one hidden layer, 128 units), a Deeper FFN (two hidden layers, 256 and 128 units), and LeNet \cite{lecun2002gradient}. All models are trained with a batch size of 64 using the Adam optimizer \citep{kingma2014adam}, and results are aggregated across 6 random seeds. To ensure our results support the evaluative focus in Section \ref{subsection-eval-focus}, we employ two training regimes:
\begin{enumerate}
    \item \textbf{Test-set performance:} To capture peak accuracy and calibration, we utilize early stopping alongside a ReduceLROnPlateau learning rate scheduler on the validation set. The final model is then evaluated on the held-out test set. By comparing performance of the new majority (mode) targets against the original targets ($orig. \rightarrow maj._n$) and the new majority against the soft-labels ($orig. \rightarrow soft_w$), we aim to determine whether observed accuracy gains stem directly from the rich supervisory signal of soft-labels, or simply a shift in sample mode within the evaluation set.
    \item \textbf{Training behavior (dynamics):} To analyze model confidence and stability during the late stages of training, we utilize a fixed period of 40 epochs and visualize the results with respect to human uncertainty. This fixed-horizon ensures temporal alignment across all seeds allowing for an observation of how soft-labels influence the stability of the learning trajectory throughout the training process. See Appendix \ref{app:experimental_details} and supplementary code for more reproducibility details. 
\end{enumerate}

\subsection{Evaluative focus: beyond aggregate metrics}
\label{subsection-eval-focus}
Although aggregate metrics provide a coarse overview of model performance, they can mask performance disparities across data subsets. We, thus, stratify our datasets into NoHLV (one-hot) and HLV (soft-target) subsets.  This prevents performance on `easy' consensus samples from skewing the results of complex, ambiguous cases, and vice versa. Our evaluation is structured around:

\textbf{(1) Testset performance:} We evaluate both accuracy and calibration to understand how well models trained on soft-labels align with human majority labels and human label variation. 
Expected Calibration Error (ECE) \cite{naeini2015obtaining, guo_2017_on_calibration} is frequently used for evaluating calibration, however it is not suited for tasks characterized by inherent human label variation as it only looks at the most likely  probability disregarding the full distribution \citep{baan2022stop, pavlovic2025calibration}. We therefore adopt the Kullback-Leibler Divergence (KLD) as our main diagnostic to measure the discrepancy between the human-annotated soft-labels and the model predictions: 
$D_{KL}(P_{\text{human}} \| P_{\text{model}})$. By using the human distribution as the reference ($P_{\text{human}}$), we measure the expected excess surprisal when the model's predictive distribution is used to approximate the collective human judgment. While we focus on KLD, to corroborate our findings we additionally report the Brier Score (Appendix \ref{appendix_evaluation_calibration}), a frequently employed alternative to compensate for the well-documented flaws in ECE \cite{pavlovic2025calibration, kull2019beyond, kumar2018trainable, chidambaram2024reassessing}. 

\textbf{(2) Training dynamics:} 
Following \cite{swayamdipta2020dataset}, we utilize Dataset Cartography to map samples based on their training dynamics, specifically tracking model confidence and variability. The confidence of sample $i$ is defined as the mean model probability of the true label $y_i^*$ across $E$ epochs, while its variability $\hat{\sigma}_i$ measures the spread across epochs via the standard deviation: 
\vspace{-0.13cm}
$$
\hat{\mu}_i = \frac{1}{E}\sum_{e=1}^E p_i^{(e)}, \qquad \hat{\sigma}_i = \sqrt{\tfrac{1}{E}\textstyle\sum_{e=1}^E (p_i^{(e)} - \hat{\mu}_i)^2}
\vspace{-0.3cm}
$$
where $p_i^{(e)} = p_{\boldsymbol{\theta}^{(e)}}(y_i^* \mid \mathbf{x}_i)$ denotes the model probability at epoch $e$.
Because we are interested in convergence behaviors, we look at the last 5 epochs of training across 6 seeds. This multi-seed, late-stage evaluation allows us to 
analyze how different target representations influence the stability 
of the learning trajectory, as well as to measure the correlation between model confidence and human uncertainty signals ($u^{prop}$, $u^{mean}$).
We further stratify these maps by human label variation. Building on the observation in \cite{swayamdipta2020dataset} that annotator agreement correlates with model confidence, we hypothesize that \textit{NoHLV} instances will cluster in the "easy" region (high confidence, low variability). Conversely, instances with \textit{HLV} should distribute across the "ambiguous" (medium confidence, high variability) and "hard" (low confidence, low variability) regions.

\begin{table}[ht]
\vspace{-.7em}
\caption{$\uparrow$ Accuracy - models trained on MNIST and Mukhoti with different targets and \textbf{evaluated on original ($orig.$) labels}; accuracy with respect to the argmax of the new human soft-labels; 
 $\mu\,_{ \pm\; \sigma}$ over 6 seeds.}
\label{tab:acc_hlvsplit}
\vspace{0.5em}
\centering
\small
\renewcommand{\arraystretch}{.8}
\begin{tabular}{crcccccc}
\toprule
 & & \multicolumn{2}{c}{\textbf{SimpleFFN}} & \multicolumn{2}{c}{\textbf{DeeperFFN}} & \multicolumn{2}{c}{\textbf{LeNet}} \\
\cmidrule(lr){3-4} \cmidrule(lr){5-6} \cmidrule(lr){7-8}
& \textbf{target} & \textbf{$No HLV$} & \textbf{$HLV$} & \textbf{$No HLV$} & \textbf{$HLV$} & \textbf{$No HLV$} & \textbf{$HLV$} \\ 
\midrule
\multirow{4}{*}{\rotatebox[origin=c]{90}{MNIST}} 
    & $orig.$   & 84.52$_{\pm 0.71}$ & 56.61$_{\pm 2.16}$ & 85.14$_{\pm 0.73}$ & 57.94$_{\pm 0.90}$ & 93.91$_{\pm 1.39}$ & 72.49$_{\pm 1.50}$ \\ 
    & $maj._{n}$ & 85.08$_{\pm 1.18}$ & 58.03$_{\pm 0.85}$ & 84.27$_{\pm 2.28}$ & 56.35$_{\pm 1.36}$ & 93.53$_{\pm 1.73}$ & 69.84$_{\pm 1.98}$ \\
    & $soft_{w}$ & \textbf{86.26}$_{\pm 0.79}$ & \textbf{58.91}$_{\pm 0.95}$ & 86.13$_{\pm 1.25}$ & 60.23$_{\pm 2.39}$ & \textbf{95.15}$_{\pm 1.10}$ & 73.72$_{\pm 0.50}$ \\ 
    & $soft_{e}$ & 85.63$_{\pm 0.83}$ & 57.14$_{\pm 0.81}$ & \textbf{86.44}$_{\pm 1.19}$ & \textbf{61.02}$_{\pm 2.15}$ & 95.03$_{\pm 0.80}$ & \textbf{74.69}$_{\pm 2.23}$ \\
\midrule
\multirow{4}{*}{\rotatebox[origin=c]{90}{Mukh.}} 
    & $synth.$   & 74.58$_{\pm 3.20}$ & \textbf{77.08}$_{\pm 1.70}$ & \textbf{82.50}$_{\pm 2.50}$ & \textbf{80.03}$_{\pm 1.08}$ & \textbf{89.58}$_{\pm 1.86}$ & \textbf{88.00}$_{\pm 1.03}$ \\ 
    & $maj._{n}$ & 73.33$_{\pm 2.13}$ & 48.92$_{\pm 0.51}$ & 71.88$_{\pm 1.73}$ & 48.46$_{\pm 1.01}$ & 78.13$_{\pm 2.47}$ & 56.83$_{\pm 4.91}$ \\ 
    & $soft_{w}$ & \textbf{75.21}$_{\pm 1.12}$ & 50.80$_{\pm 2.78}$ & 75.42$_{\pm 0.93}$ & 50.40$_{\pm 1.33}$ & 80.21$_{\pm 1.33}$ & 59.39$_{\pm 2.73}$ \\ 
    & $soft_{e}$ & 74.79$_{\pm 1.12}$ & 50.00$_{\pm 1.62}$ & 74.79$_{\pm 1.12}$ & 50.17$_{\pm 1.55}$ & 80.00$_{\pm 1.44}$ & 57.57$_{\pm 1.85}$ \\ 
\bottomrule
\end{tabular}
\end{table}

\subsection{Test-set performance on human variation strata}

Performance is assessed across two dimensions: an evaluation label shift, meaning comparing the same model against a changing evaluation set ($orig.$  vs. $new$); and a training target shift, comparing the effect of training on original, majority-vote, or soft distributions ($orig.$ vs $maj._n$ vs $soft_w$ vs  $soft_e$) evaluated against the same evaluation set ($new$, i.e. the re-annotated human ground truth) .

\textbf{Evaluation label shift:} 
On MNIST, as shown in Table \ref{tab:acc_hlvsplit} and Table \ref{tab:acc_hlvsplit2}, performance on the NoHLV subset is identical across both evaluation sets ($orig.$ and $new$) for all training targets and architectures, confirming that global performance differences stem from the more challenging HLV subset. For each training target, evaluation on the $new$ labels (Table \ref{tab:acc_hlvsplit2}) shows a consistent $\sim$3\% accuracy gain over the corresponding $orig.$ labels, indicating that test-set label noise accounts for a measurable performance loss on harder/ambiguous items. A marginal portion of this discrepancy ($<1$\% of the total dataset) is attributable to "NaN categories".
Mukhoti, on the other hand show inverted behavior: When evaluated on the original synthetic labels ($orig.$, Table \ref{tab:acc_hlvsplit}), models trained on synthetic targets appear dominant, achieving $\sim$28\% higher accuracy on the HLV subset than models trained on human annotated targets ($\sim$77-88\% vs. $\sim$48-59\%). However, switching the evaluation to the human HLV test-set ($new$, Table \ref{tab:acc_hlvsplit2}) shows that performance of the synthetically trained models degrades by $\sim$23-32\%; while models trained on human annotated targets gain up to $\sim$14\% on NoHLV and around $\sim$5-11\% on HLV.

\begin{table}[ht]
\vspace{-0.7em}
\caption{$\uparrow$ Accuracy - models trained on MNIST and Mukhoti with different targets and \textbf{evaluated on re-labeled data} ($new$); accuracy with respect to the argmax of the new human soft-labels; 
 $\mu\,_{ \pm\; \sigma}$ over 6 seeds. }
\label{tab:acc_hlvsplit2}
\vspace{0.5em}
\centering
\small
\renewcommand{\arraystretch}{.8} 
\begin{tabular}{crcccccc}
\toprule
 & & \multicolumn{2}{c}{\textbf{SimpleFFN}} & \multicolumn{2}{c}{\textbf{DeeperFFN}} & \multicolumn{2}{c}{\textbf{LeNet}} \\
\cmidrule(lr){3-4} \cmidrule(lr){5-6} \cmidrule(lr){7-8}
& \textbf{target} & \textbf{$No HLV$} & \textbf{$HLV$} & \textbf{$No HLV$} & \textbf{$HLV$} & \textbf{$No HLV$} & \textbf{$HLV$} \\ 
\midrule
\multirow{4}{*}{\rotatebox[origin=c]{90}{MNIST}} 
    & $orig.$   & 84.52$_{\pm 0.71}$ & 59.79$_{\pm 1.93}$ & 85.14$_{\pm 0.73}$ & 60.32$_{\pm 1.56}$ & 93.91$_{\pm 1.39}$ & 73.90$_{\pm 1.04}$ \\ 
    & $maj._{n}$ & 85.08$_{\pm 1.18}$ & 61.46$_{\pm 1.31}$ & 84.27$_{\pm 2.28}$ & 59.97$_{\pm 1.25}$ & 93.53$_{\pm 1.73}$ & 73.72$_{\pm 1.13}$ \\
    & $soft_{w}$ & \textbf{86.26}$_{\pm 0.79}$ & \textbf{62.35}$_{\pm 1.60}$ & 86.13$_{\pm 1.25}$ & \textbf{64.90}$_{\pm 2.76}$ & \textbf{95.15}$_{\pm 1.10}$ & 76.63$_{\pm 0.48}$ \\ 
    & $soft_{e}$ & 85.63$_{\pm 0.83}$ & 60.49$_{\pm 1.95}$ & \textbf{86.44}$_{\pm 1.19}$ & 64.82$_{\pm 2.64}$ & 95.03$_{\pm 0.80}$ & \textbf{77.16}$_{\pm 2.23}$ \\
\midrule
\multirow{4}{*}{\rotatebox[origin=c]{90}{Mukh.}}  
    &  $synth.$   & 69.58$_{\pm 4.49}$ & 53.87$_{\pm 1.21}$ & 75.00$_{\pm 1.02}$ & 55.29$_{\pm 0.71}$ & 77.71$_{\pm 2.54}$ & 56.26$_{\pm 1.37}$ \\
    & $maj._{n}$ & 83.54$_{\pm 2.09}$ & 55.18$_{\pm 1.27}$ & 83.96$_{\pm 4.11}$ & 56.03$_{\pm 2.96}$ & 92.50$_{\pm 1.44}$ & 60.69$_{\pm 1.48}$ \\
    & $soft_{w}$ & \textbf{89.58}$_{\pm 1.38}$ & \textbf{59.78}$_{\pm 2.76}$ & \textbf{91.04}$_{\pm 1.33}$ & \textbf{62.34}$_{\pm 2.08}$ & \textbf{96.46}$_{\pm 1.97}$ & 65.07$_{\pm 2.15}$ \\ 
    & $soft_{e}$ & 88.96$_{\pm 1.83}$ & 58.59$_{\pm 2.74}$ & 90.63$_{\pm 1.40}$ & 61.32$_{\pm 1.98}$ & 95.83$_{\pm 1.86}$ & \textbf{65.47}$_{\pm 2.38}$ \\
\bottomrule
\end{tabular}
\end{table}

\textbf{Training target shift:} 
Training on $orig.$ and $maj_n$ one-hot targets produces comparable performance across both HLV and NoHLV strata. Transitioning from $maj_n$ to $soft_{w/e}$ targets provides more noticeable improvements, with increases of 1–4\% on HLV samples across architectures compared to a more modest 1–2\% improvement on the NoHLV split. These gains are statistically significant for the two more capable architectures on the HLV split; however, they otherwise remain modest relative to the variance across runs. 
In contrast, improvements on Mukhoti are substantially more pronounced for all architectures across the HLV split, with a clear tiered improvement in accuracy: on the HLV subset, transitioning from $synth.$ to $maj._n$ training shows a 1-4\% gain, with further $\sim$4-6\% improvements across the three architectures when training on $soft_{w/e}$. This progression (from $synth.$ to $maj_n$ to $soft_{w/e}$) is consistent across the NoHLV split albeit with higher accuracy gains in the transitions. 

To better isolate these effects, we decompose the effect for every architecture on both datasets by splitting the accuracy contributed by relabeling alone (original one-hot label $\rightarrow$ new majority label) from the accuracy contributed by soft-label training (new majority label $\rightarrow$ soft-label), see Table \ref{tab:acc_decomp}.
On MNIST, the benefit from training on new modes is near zero or slightly negative across all three architectures. Almost all the gain comes from the soft-label information itself, which is consistent with MNIST's original labels already being highly reliable. 
On Mukhoti, the benefit from training on new modes is large on NoHLV items (+9.0 to +14.8 points), as most of these items needed their argmax corrected. However, the shift is smaller on HLV items (+0.7 to +4.4), where there is rarely a clean majority to shift to. Notably, soft-label information adds a further 4.0 to 7.1 points across every architecture and group on Mukhoti, showing that there is some benefit to soft-labels on hard/ambiguous data beyond just fixing the argmax.

\begin{table}[htbp!]
\vspace{-0.7em}
\centering
\caption{Decomposition of accuracy gains on the \textbf{re-annotated evaluation set ($new$)} from  Table \ref{tab:acc_hlvsplit2}. "New majority benefit" isolates the gain from training on newly relabeled data alone ($orig. \rightarrow maj._n$). "Soft-benefit" isolates the accuracy gain by training on soft-labels opposed to one-hot labels ($maj._n \rightarrow soft_w$).
}
\label{tab:acc_decomp}
\vspace{0.5em}
\centering
\small
\renewcommand{\arraystretch}{.8} 
\begin{tabular}{llrrrrrr}
\toprule
\multirow{2}{*}{
} & \multirow{2}{*}{\textbf{Model}} & \multicolumn{2}{c}{\textbf{New majority-benefit}} & \multicolumn{2}{c}{\textbf{Soft-benefit}} & \multicolumn{2}{c}{\textbf{Total ($soft_w-orig.$)}} \\
\cmidrule(lr){3-4} \cmidrule(lr){5-6} \cmidrule(lr){7-8}
 & & NoHLV & HLV & NoHLV & HLV & NoHLV & HLV \\
\midrule
\multirow{3}{*}{\rotatebox[origin=c]{90}{\scriptsize MNIST}} & SimpleFFN  & +0.6  & +1.7 & +1.2 & +0.9 & +1.7  & +2.6 \\
     & DeeperFFN  & -0.9  & -0.4 & +1.9 & +4.9 & +1.0  & +4.6 \\
     & LeNet      & -0.4  & -0.2 & +1.6 & +2.9 & +1.2  & +2.7 \\
\midrule
\multirow{3}{*}{\rotatebox[origin=c]{90}{Mukh.}} & SimpleFFN  & +14.0 & +1.3 & +6.0 & +4.6 & +20.0 & +5.9 \\
      & DeeperFFN  & +9.0  & +0.7 & +7.1 & +6.3 & +16.0 & +7.1 \\
      & LeNet      & +14.8 & +4.4 & +4.0 & +4.4 & +18.8 & +8.8 \\
\bottomrule
\end{tabular}%
\vspace{-0.5em}
\end{table}

While our findings align with \cite{peterson_2019_iccv} regarding the benefits in accuracy of soft-label training, the difference in magnitude between MNIST and Mukhoti suggests that soft-labels offer greater utility as inherent dataset ambiguity increases. Beyond corroborating these directional trends, our evaluation extends \cite{peterson_2019_iccv} by isolating the effects of label mode shifts introduced during re-annotation, a confounding factor unaddressed in prior work.
However, the greater utility of human soft-labels emerges in calibration. 
As detailed in Table \ref{tab:klddiv}, models supervised by soft-labels achieve significantly lower KLD on HLV subsets compared to those trained on $orig./synth.$ or $maj_{n}$ targets. 
On the NoHLV subsets, soft-labels mostly maintain a KLD comparable to $maj_{n}$ training; especially when using the more capable LeNet.  Without stratifying the evaluation data, this detail goes unnoticed in \cite{peterson_2019_iccv}. 
Soft-label supervision enables models to capture fine-grained information that is unrecoverable from collapsed one-hot representations. This is further supported by finding that soft-labels also consistently improve HLV Brier scores across both datasets (Appendix \ref{appendix_evaluation_calibration}).
This underlines the value of incorporating human label variation into evaluation frameworks, as one-hot or synthetic targets may inadvertently encourage models to learn confidences that fail to capture human ambiguity.

\begin{table}[ht]
\vspace{-.7em}
\caption{$\downarrow$ KLDivergence -  models trained on MNIST and Mukhoti with different targets and evaluated on our newly re-labeled data, more specifically $soft_{w}$; $\mu\,_{ \pm\; \sigma}$ over 6 seeds.
}
\label{tab:klddiv}
\vspace{0.5em}
\centering
\small
\renewcommand{\arraystretch}{.8}
\setlength{\tabcolsep}{7pt} 
\begin{tabular}{crcccccc}
\toprule
 & & \multicolumn{2}{c}{\textbf{SimpleFFN}} & \multicolumn{2}{c}{\textbf{DeeperFFN}} & \multicolumn{2}{c}{\textbf{LeNet}} \\
\cmidrule(lr){3-4} \cmidrule(lr){5-6} \cmidrule(lr){7-8}
& \textbf{target} & \textbf{$No HLV$} & \textbf{$HLV$} & \textbf{$No HLV$} & \textbf{$HLV$} & \textbf{$No HLV$} & \textbf{$HLV$} \\ 
\midrule
\multirow{4}{*}{\rotatebox[origin=c]{90}{MNIST}} & $orig.$ & 0.487$_{ \pm 0.007 }$ & 1.917$_{ \pm 0.100 }$ & 0.453$_{ \pm 0.022 }$ & 2.049$_{ \pm 0.181 }$ & 0.178$_{ \pm 0.041 }$ & 2.214$_{ \pm 0.073 }$ \\ 
    & $maj._{n}$ & 0.457$_{ \pm 0.010 }$ & 1.543$_{ \pm 0.049 }$ & 0.443$_{ \pm 0.025 }$ & 1.484$_{ \pm 0.107 }$ & \textbf{0.175}$_{ \pm {0.031} }$ & 1.661$_{ \pm 0.061 }$ \\
    & $soft_{w}$ & \textbf{0.453}$_{ \pm 0.019 }$ & 1.028$_{ \pm 0.025 }$ & 0.408$_{ \pm 0.022 }$ & 0.989$_{ \pm 0.021 }$ & 0.181$_{ \pm 0.014 }$ & 0.669$_{ \pm 0.023 }$ \\ 
    & $soft_{e}$ & 0.454$_{ \pm 0.021 }$ & \textbf{1.021}$_{ \pm 0.019 }$ & \textbf{0.407}$_{ \pm 0.023 }$ & \textbf{0.983}$_{ \pm 0.027 }$ & 0.183$_{ \pm 0.016 }$ & \textbf{0.660}$_{ \pm {0.029} }$ \\
\midrule
\multirow{4}{*}{\rotatebox[origin=c]{90}{Mukh.}} & $synth.$ & 0.804$_{ \pm 0.036 }$ & 1.525$_{ \pm 0.069 }$ & 0.667$_{ \pm 0.024 }$ & 1.525$_{ \pm 0.019 }$ & 0.529$_{ \pm 0.040 }$ & 1.612$_{ \pm 0.044 }$ \\
    & $maj._{n}$ & 0.565$_{ \pm 0.028 }$ & 0.922$_{ \pm 0.037 }$ & 0.516$_{ \pm 0.069 }$ & 0.915$_{ \pm 0.040 }$ & 0.266$_{ \pm 0.031 }$ & 0.905$_{ \pm 0.024 }$ \\
    & $soft_{w}$ & \textbf{0.425}$_{ \pm 0.018 }$ & \textbf{0.550}$_{ \pm 0.012 }$ & \textbf{0.351}$_{ \pm 0.030 }$ & \textbf{0.537}$_{ \pm 0.011 }$ & \textbf{0.262}$_{ \pm {0.034} }$ & 0.479$_{ \pm 0.013 }$ \\ 
    & $soft_{e}$ & 0.451$_{ \pm 0.022 }$ & 0.556$_{ \pm 0.014 }$ & 0.368$_{ \pm 0.013 }$ & 0.543$_{ \pm 0.010 }$ & 0.265$_{ \pm 0.019 }$ & \textbf{0.468}$_{ \pm {0.011} }$ \\
\bottomrule
\end{tabular}
\end{table}

\subsection{Training dynamics on human variation strata}

\begin{figure}[htbp]
    \centering
    \begin{subfigure}[b]{0.49\textwidth}
        \centering
        \includegraphics[width=\textwidth]{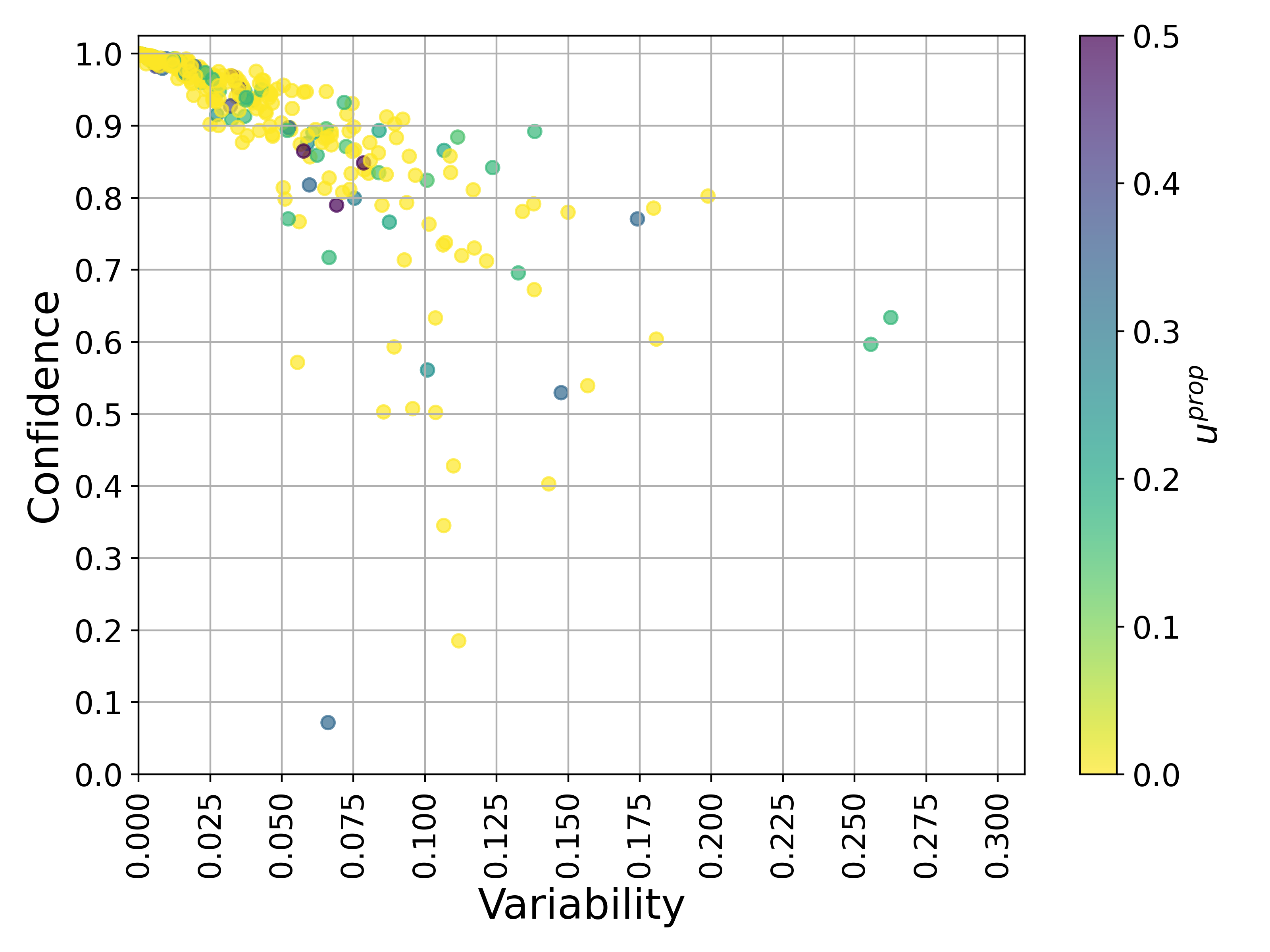}
        \vspace{-1.7em}
        \caption{Mukhoti - \textit{NoHLV} - $maj._{n}$ }
        \label{fig:row1_left}
    \end{subfigure}
    \begin{subfigure}[b]{0.49\textwidth}
        \centering
        \includegraphics[width=\textwidth]{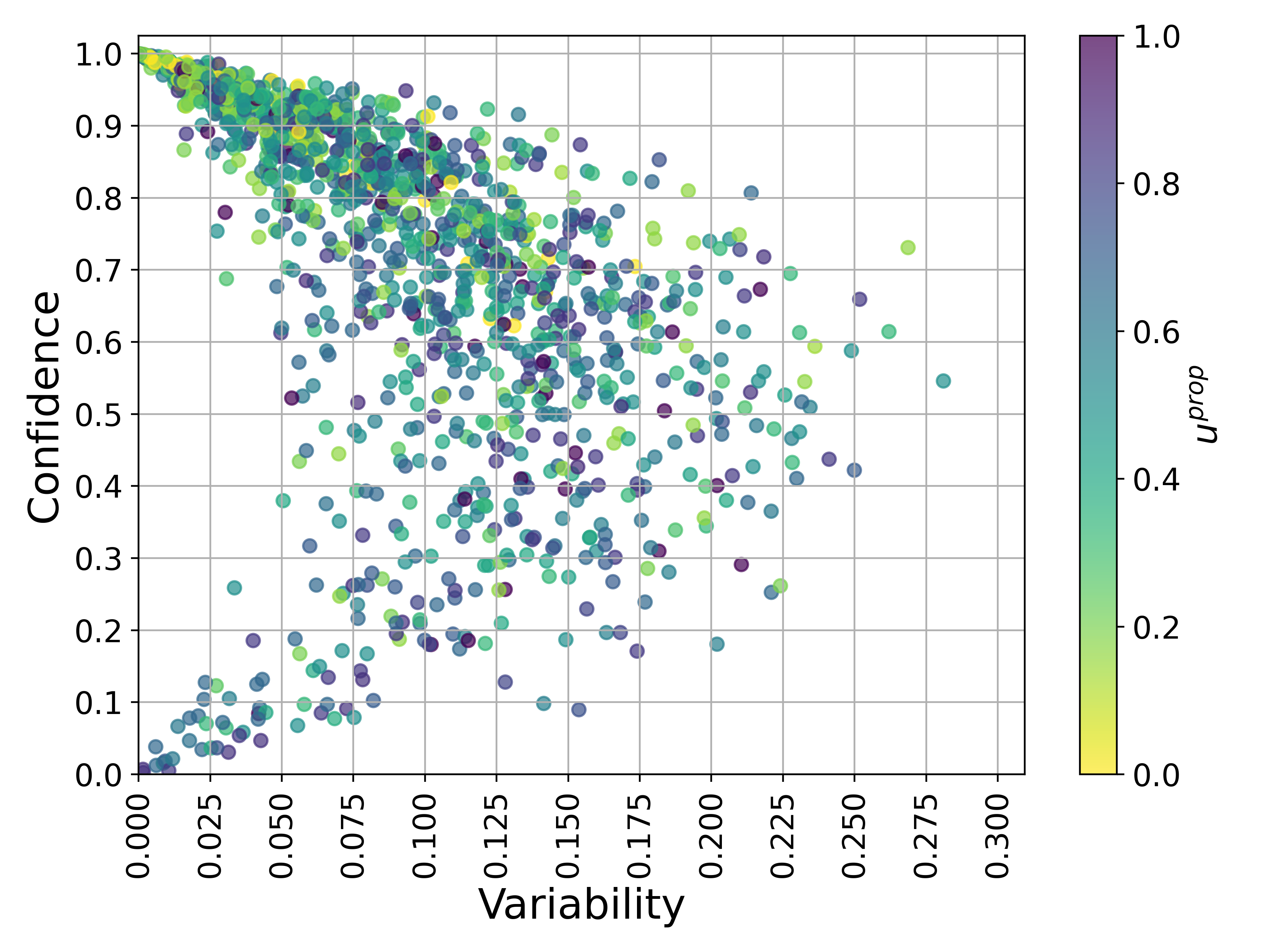}
        \vspace{-1.7em}
        \caption{Mukhoti - \textit{HLV} - $maj._{n}$ }
        \label{fig:row1_right}
    \end{subfigure}

    \vspace{.5ex} 

    \begin{subfigure}[b]{0.49\textwidth}
        \centering
        \includegraphics[width=\textwidth]{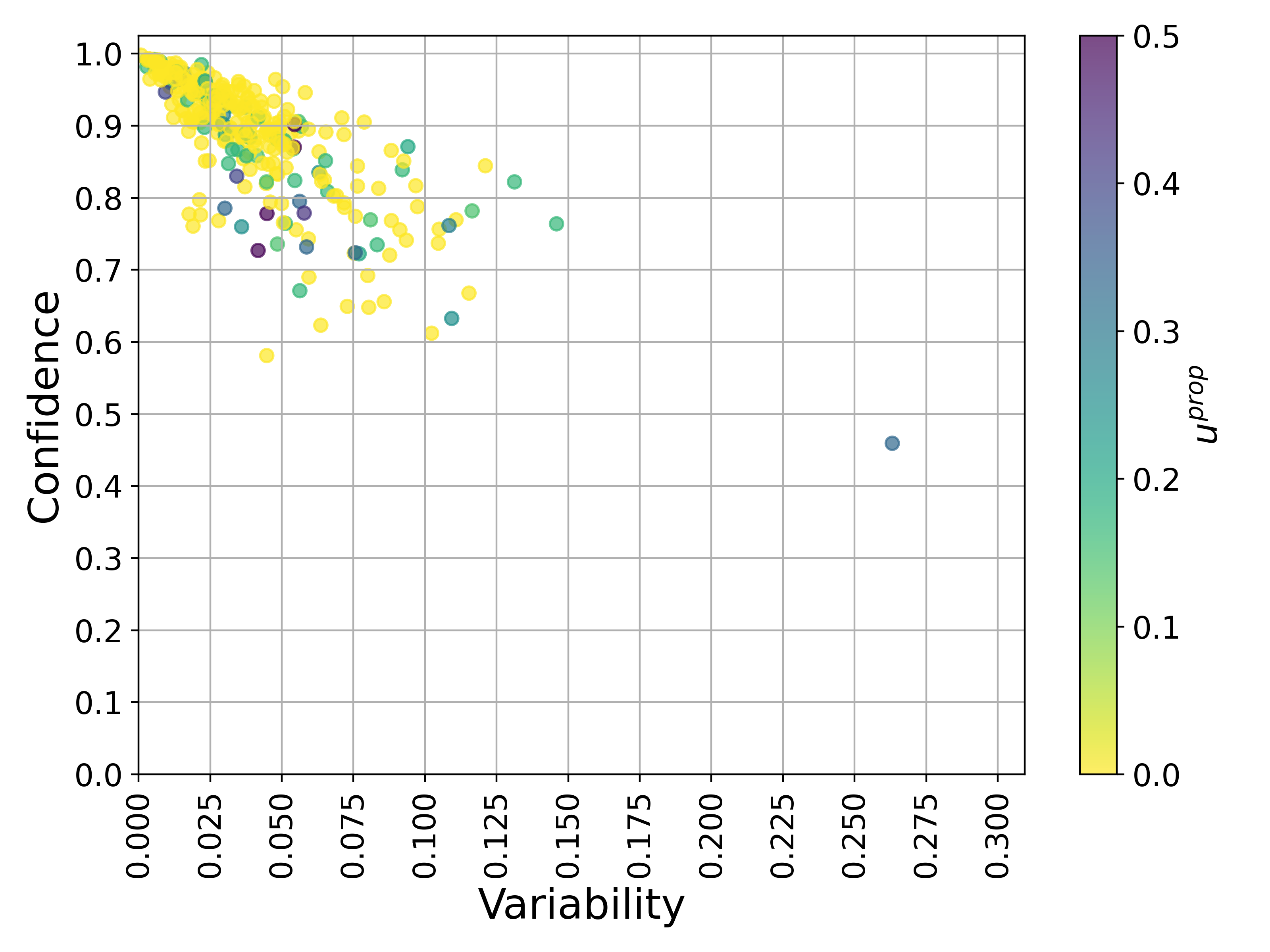}
        \vspace{-1.7em}
        \caption{Mukhoti - \textit{NoHLV} - $soft_{w}$ }
        \label{fig:row2_left}
    \end{subfigure}
    \begin{subfigure}[b]{0.49\textwidth}
        \centering
        \includegraphics[width=\textwidth]
        {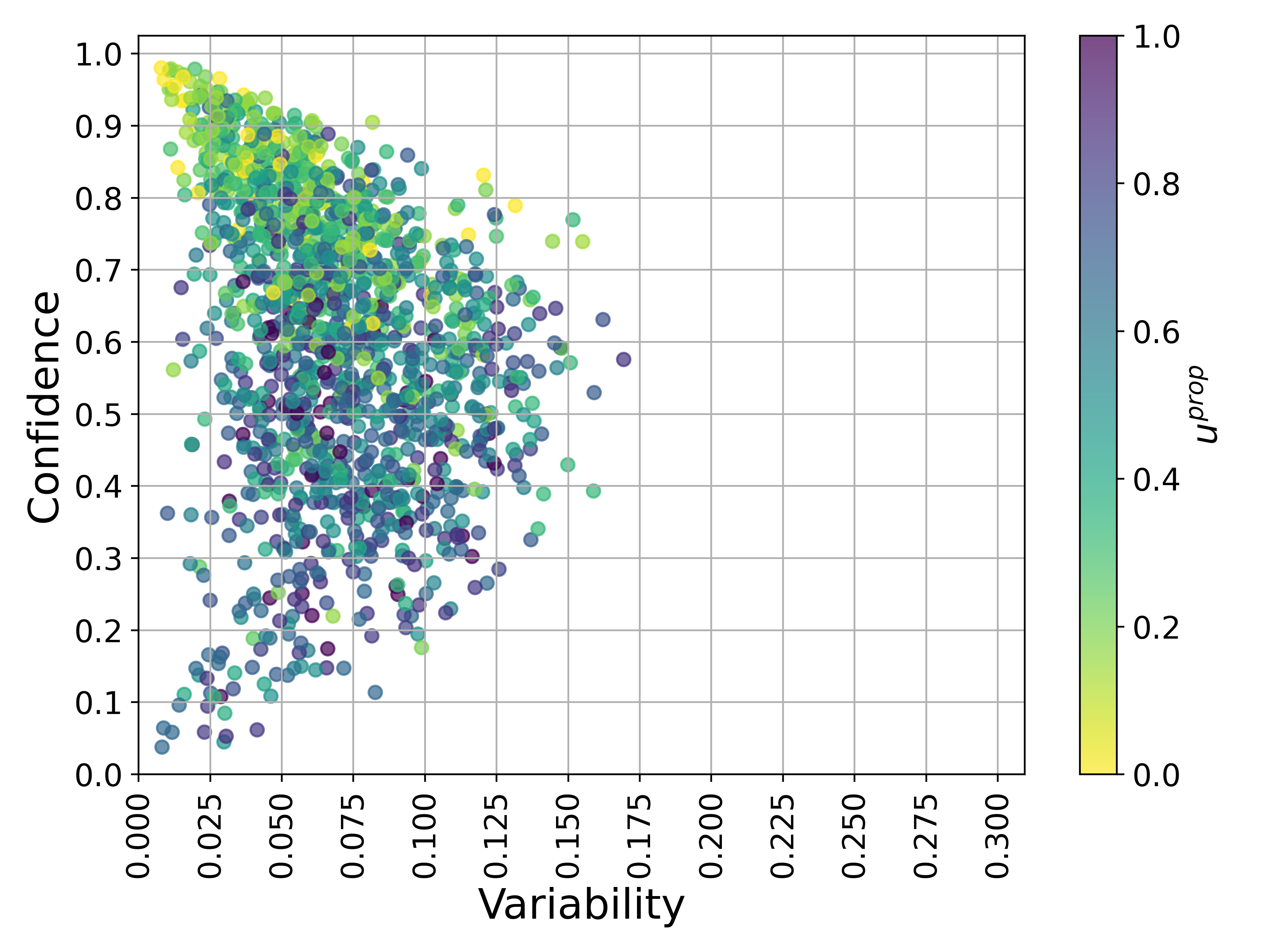}
        \vspace{-1.7em}
        \caption{Mukhoti - \textit{HLV} - $soft_{w}$ }
        \label{fig:row2_right}
    \end{subfigure}

    \vspace{.5ex}

    \begin{subfigure}[b]{0.49\textwidth}
        \centering
        \includegraphics[width=\textwidth]{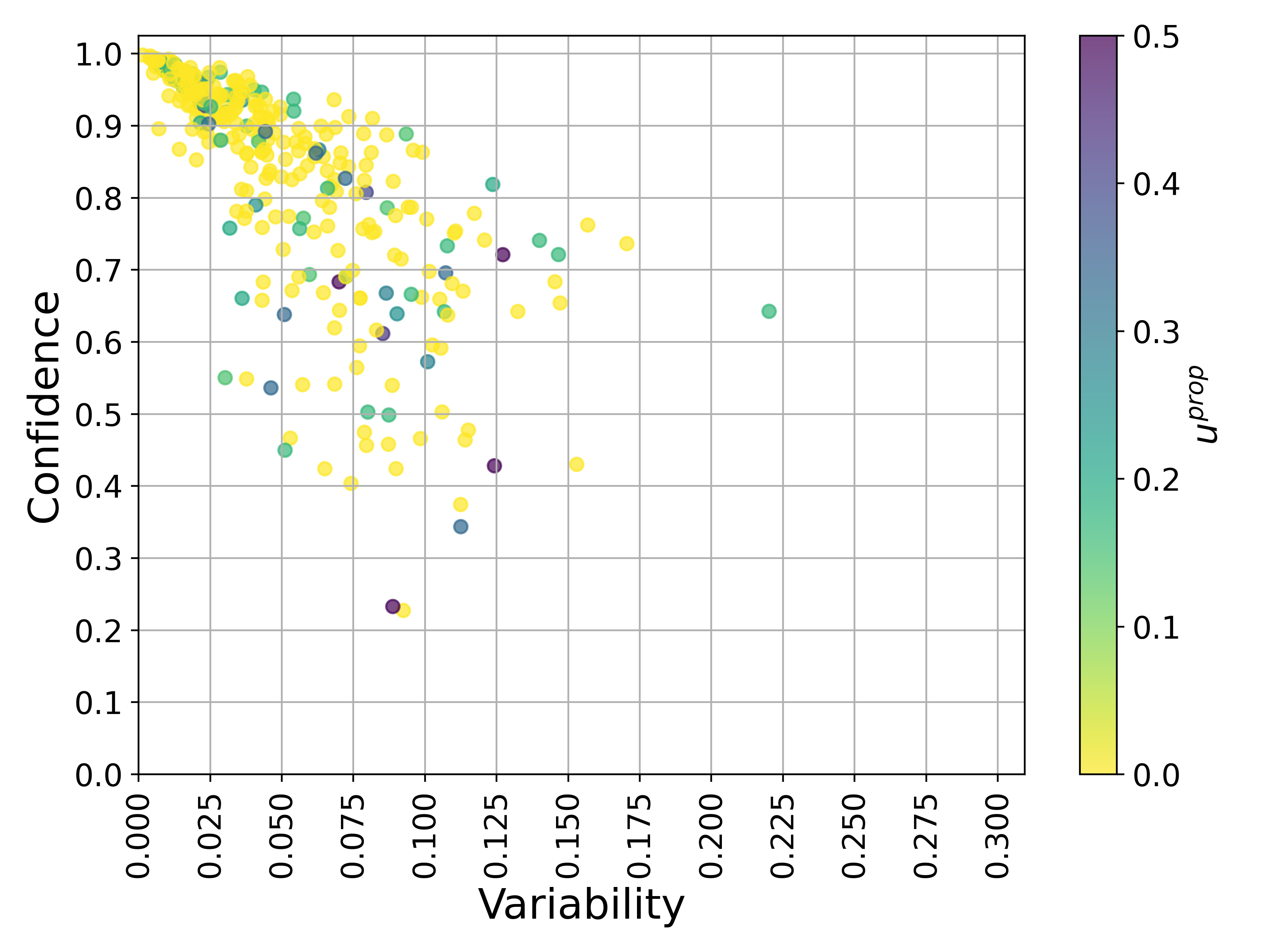}
        \vspace{-1.7em}
        \caption{Mukhoti - \textit{NoHLV} - $synth.$ }
        \label{fig:row3_left}
    \end{subfigure}
    \begin{subfigure}[b]{0.49\textwidth}
        \centering
        \includegraphics[width=\textwidth]{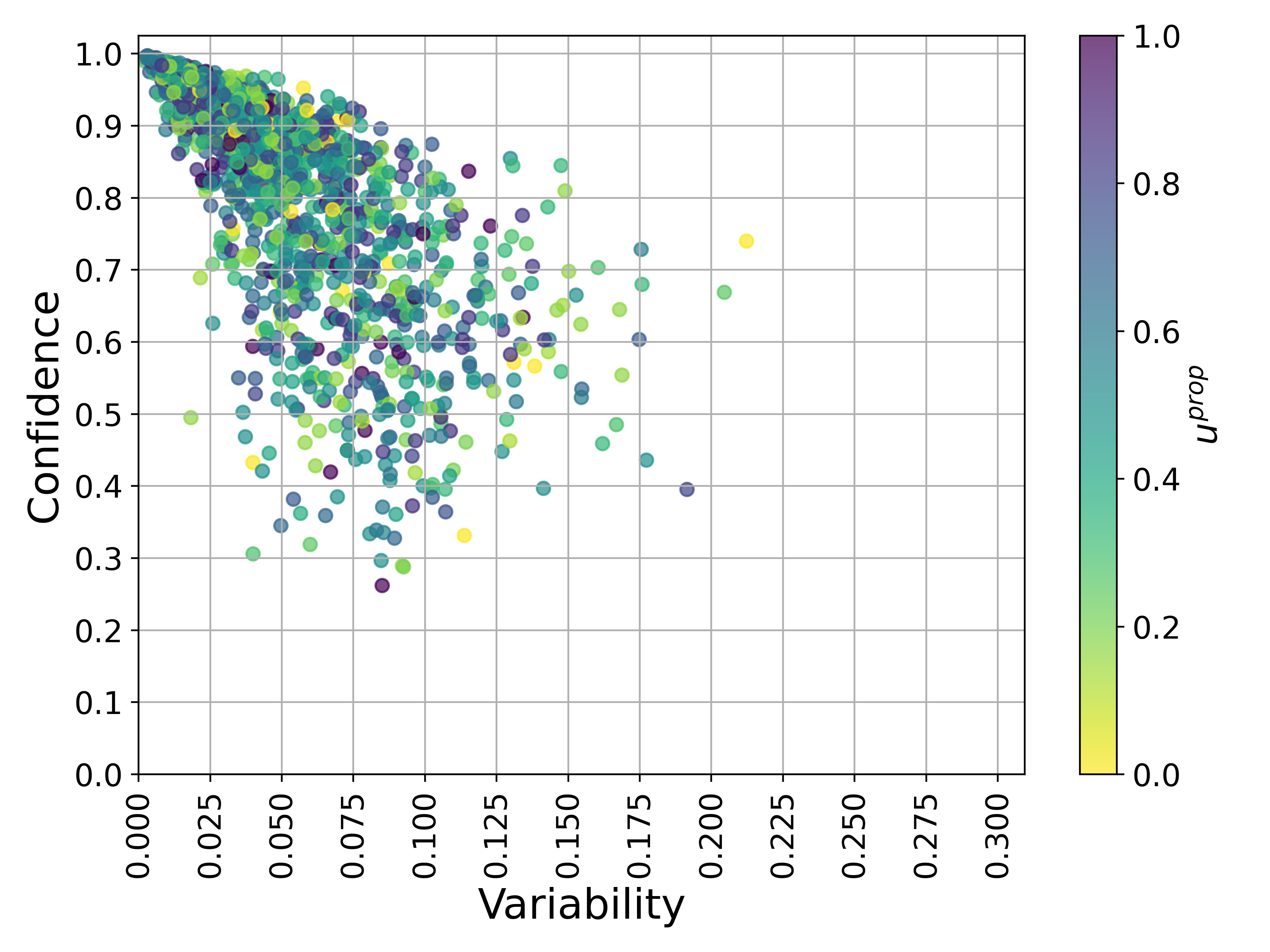}
        \vspace{-1.7em}
        \caption{Mukhoti - \textit{HLV} - $synth.$ }
        \label{fig:row3_right}
    \end{subfigure}

    \caption{
    Late stage training dynamics of LeNet {\small (last 5 epochs)} averaged across 6 random seeds on Mukhoti data, with $soft_w$, $maj._n$ and $orig.$ {\small (synthetic)} contrasted; training on soft-labels indicates lower variability in HLV strata compared to training on majority label ($maj_n$). }
    \label{fig:mukhoti_data_maps_human_uncertainty}

\end{figure}

\begin{figure}[htbp]
  \centering
  \begin{minipage}[b]{0.32\textwidth}
    \centering
    \begin{subfigure}[t]{0.45\textwidth}
        \centering
        \includegraphics[width=0.55\textwidth]{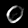}
        \caption*{(a) {\small 
        6 Ann. total, 0/6 are "unsure"}}
    \end{subfigure}\hfill
    \begin{subfigure}[t]{0.45\textwidth}
        \centering
        \includegraphics[width=0.55\textwidth]{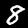}
        \caption*{(b) {\small 
        7 Ann. total,  0/7 are "unsure"}}
    \end{subfigure}
    \vspace{-0.2em}
    \caption*{\makebox[1.1\textwidth][c]{(1) Model $\uparrow$conf. \& $\downarrow$var.}}
  \end{minipage}%
  \hfill
  \begin{minipage}[b]{0.32\textwidth}
    \centering
    \begin{subfigure}[t]{0.45\textwidth}
        \centering
        \includegraphics[width=0.55\textwidth]{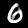}
        \caption*{(c)  {\small 
        6 Ann. total,  2/6 are "unsure"}}
    \end{subfigure}\hfill
    \begin{subfigure}[t]{0.45\textwidth}
        \centering
        \includegraphics[width=0.55\textwidth]{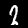}
        \caption*{(d)  {\small 
        7 Ann. total 2/7 are "unsure"}}
    \end{subfigure}
    \vspace{-0.1em}
    \caption*{\makebox[1.1\textwidth][c]{(2) Model $\leftrightarrow$conf. \& $\uparrow$var.}}
  \end{minipage}%
  \hfill
  \begin{minipage}[b]{0.32\textwidth}
    \centering
    \begin{subfigure}[t]{0.45\textwidth}
        \centering
        \includegraphics[width=0.55\textwidth]{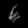}
        \caption*{(e) {\small
        6 Ann. total, 4/6 are "unsure"}}
    \end{subfigure}\hfill
    \begin{subfigure}[t]{0.45\textwidth}
        \centering
        \includegraphics[width=0.55\textwidth]{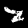}
        \caption*{(f) {\small
        {\footnotesize10} Ann. total, {\footnotesize6/10} "unsure"}}
    \end{subfigure}
    \vspace{-0.2em}
    \caption*{\makebox[1.1\textwidth][c]{(3) Model $\downarrow$conf. \& $\downarrow$var.}}
  \end{minipage}
  \caption{Dataset Cartography examples with human label variation (HLV) from both Mukhoti (a,c,e) and MNIST (b,d,f) - samples extracted from last 5 epochs of training a LeNet model; (1) indicating annotator noise with no "unsure" selections, yet entropy in the final targets, (3.e) shows 4 out of 6 annotators are unsure, while for (3.f) 6 out of 10 are unsure. }
  \label{fig:cartography_sample_images}
  \vspace{-0.3cm}
\end{figure}

\textbf{Model confidence \& variability across strata and targets: }
As hypothesized, the \textit{NoHLV} instances predominantly cluster in the "easy-to-learn" region of the Data Maps, characterized by high confidence and low variability.
Training with majority targets ($maj_n$) on the Mukhoti dataset produces an average confidence of $0.92\pm0.01$. Conversely, training with soft-labels ($soft_w$) produces a slightly lower average confidence of $0.89\pm0.005$, a smoothing effect consistent with 
\cite{peterson_2019_iccv}. Models trained on synthetic data show the lowest average confidence overall at $0.84\pm0.003$.

The \textit{HLV} subset spans a much broader spectrum of the Data Maps. Average confidences drop with the exception of training on synthetic targets, where no samples fall into the "hard-to-learn" region. Within HLV, $maj_n$ training spans all confidences, resulting in an average confidence of $0.73\pm0.03$, while $soft_w$ also spans the entire confidence range its average confidence drops to $0.61\pm0.005$. The synthetic targets ($orig.$) maintain a relatively high confidence of $0.82\pm0.002$, comparatively and no samples fall into the "hard-to-learn" region.
Notably, $maj_n$ training shows a markedly wider spread along the variability axis compared to $soft_w$. To isolate the source of this variance, we use the Jensen-Shannon Divergence between successive epochs, $JSD(P_e || P_{e-1})$ to measure late-stage updates. We find that $maj_n$ and $soft_w$ have effectively converged (unlike $synth.$), see Appendix \ref{app:training_dynamics} for details. This indicates that the variability observed in $maj._n$ and $soft_w$ maps stems from cross-seed variance rather than intra-training confidence updates. MNIST shows a similar confidence and variability pattern, which generalizes across SimpleFFN and DeeperFFN (Appendix \ref{app:training_dynamics}). 

This confidence and variability suppression of $soft_w$ mirrors the improved calibration results observed on the test set (Table \ref{tab:klddiv}), enabling the model to acknowledge uncertainty rather than overfitting to the false certainty of one-hot targets or the human-misaligned synthetic targets. 
Training under $soft_w$ shows the model is learning to be uncertain when humans are uncertain. This alignment is a desired evaluative outcome for tasks where human-like judgment is the target. 

\textbf{Samples of human uncertainty: } Observing the individual sample uncertainties reveals further details: The \textit{NoHLV} subset contains some points where a few 
annotators expressed uncertainty (i.e. purple $u_{prop}$ points). However, these selections align with the other annotators, thus resulting in an aggregated soft-label without human label variation. In the \textit{HLV} subset, overall uncertainty is higher. 
For synthetic data, uncertainties appear randomly distributed across the maps. However, under $soft_w$ training, instances with low $u^{prop}$ (i.e yellow points) densely populate the high-confidence/low-variability region. Again, MNIST shows a similar pattern for both $maj_n$ and $soft_w$ for LeNet and the same pattern holds across the two remaining models (Appendix \ref{app:training_dynamics}).  Inspection of some of these specific samples with \textit{HLV}  indicates that models struggle less with instances containing obvious annotator noise (e.g. Figure \ref{fig:cartography_sample_images} (1)); or with samples which are highly noisy (3). Whereas their cross-seed variance is highest on samples that have valid interpretive ambiguities (2).
While recent studies \cite{xia2024towards} demonstrate that label smoothing degrades selective classification by failing to suppress confidence on poorly fit samples, our preliminary observation (c, d \& f) suggests human-centric soft-labels behave somewhat differently, forcing appropriate confidence suppression on highly ambiguous or hard samples.

\begin{table}[ht]
\caption{Spearman correlation between late stage model training dynamics and human uncertainty ($u_{prop}$)  across 6 seeds.}
\label{tab:spearman_all_nets}
\vspace{0.5em}
\centering
\small
\renewcommand{\arraystretch}{1.05} 
\setlength{\tabcolsep}{2.2pt} 
\begin{tabular}{crcccccc}
\toprule
 & & \multicolumn{2}{c}{\textbf{SimpleFFN}} & \multicolumn{2}{c}{\textbf{DeeperFFN}} & \multicolumn{2}{c}{\textbf{LeNet}} \\
\cmidrule(lr){3-4} \cmidrule(lr){5-6} \cmidrule(lr){7-8}
\textbf{eval.} & \textbf{target} & {Conf.(${u^{prop}}$)} & {Var.(${u^{prop}}$)} & {Conf.(${u^{prop}}$)} & {Var.(${u^{prop}}$)}  & {Conf.(${u^{prop}}$)} & {Var.(${u^{prop}}$)} \\ 
\midrule
\multirow{2}{*}{\rotatebox[origin=c]{90}{\scriptsize MNIST}} & $soft_{w}$ & $-0.62_{(p < 0.001)}$ & $0.41_{(p < 0.001)}$ & $-0.71_{(p < 0.001)}$ & $0.58_{(p < 0.001)}$ & $-0.70_{(p < 0.001)}$ & $0.61_{(p < 0.001)}$ \\
& $maj._n$ & $-0.34_{(p < 0.001)}$ & $0.32_{(p < 0.001)}$ & $-0.36_{(p < 0.001)}$ & $0.35_{(p < 0.001)}$ & $-0.47_{(p < 0.001)}$ & $0.46_{(p < 0.001)}$ \\
\midrule
\multirow{3}{*}{\rotatebox[origin=c]{90}{Mukh.}} & $soft_{w}$ & $-0.66_{(p < 0.001)}$ & $0.25_{(p < 0.001)}$ & $-0.72_{(p < 0.001)}$ & $0.38_{(p < 0.001)}$ & $-0.68_{(p < 0.001)}$ & $0.38_{(p < 0.001)}$ \\
& $maj._n$ & $-0.37_{(p < 0.001)}$ & $0.25_{(p < 0.001)}$ & $-0.38_{(p < 0.001)}$ & $0.27_{(p < 0.001)}$ & $-0.44_{(p < 0.001)}$ & $0.36_{(p < 0.001)}$ \\
& $synth.$ & $-0.08_{(p < 0.001)}$ & $0.06_{(p < 0.05)}$ & $-0.10_{(p < 0.001)}$ & $0.07_{(p < 0.01)}$ & $-0.09_{(p < 0.001)}$ & $0.08_{(p < 0.001)}$ \\
\bottomrule
\end{tabular}
\end{table}

\textbf{Quantifying the alignment between human and model uncertainty: }
To move beyond qualitative Data Map visualizations, we measure the Spearman correlation between human uncertainty ($u^{prop}$) and model training dynamics (Table \ref{tab:spearman_all_nets}). 
Overall, model confidence (conf.) shows a strong negative correlation with human uncertainty, while model variability  (var.) shows a positive correlation.
Training with soft-labels ($soft_w$) produces the strongest correlations. It shows meaningful negative correlations with model confidence and moderate positive correlations with training variability. Converting soft-labels into discrete majority targets $maj._n$ uniformly weakens this alignment. While correlations remain statistically significant, their magnitudes drop notably, indicating that argmax discretization discards valuable uncertainty information \cite{peterson_2019_iccv, uma2021learning}. 
Regarding the synthetic targets from Mukhoti we observe negligible confidence and variability correlation with human uncertainty. 
This reinforces our preceding findings that the synthetic labels generated in this specific setup do not encode human-level uncertainty, see Appendix \ref{app:spearman_umean} for $u_{mean}$ results.

\section{Discussion}
\label{sec:discussion}
Although large-scale benchmarks drive performance, their high dimensionality often complicates the study of underlying model behaviors. This work aligns with the perspective of evaluation as a scientific object by providing a tractable, controlled environment to observe how models internalize human label variance. Rather than serving as a replacement for large-scale benchmarks, our datasets and evaluation function as a diagnostic testbed for investigating the interplay of human and model uncertainty under varying target representations.

\textbf{Label misalignment and leakage: } 
Ground truth labels can diverge from collective human perception: a digit created under the instruction to write a '7' is a '7' by the creators intent, other observers may mostly perceive it as a '1'. We show that  $\sim$3\% of accuracy gains on our MNIST subset stem from fixing label misalignment rather than architectural improvements. Similarly, re-annotating the Mukhoti dataset leads to a majority label change for $\sim$33\% of the samples. These findings suggest that we should prioritize auditing difficult samples to track genuine progress; otherwise, improvements may merely reflect "fitting the noise" in original mislabelings. Furthermore, the identification of train-test leakage in the Mukhoti and additional digit datasets underscores the persistent need for data auditing; without which we risk overestimating the generalization capabilities of our models.

\textbf{Soft-labels as a regularizer: }
On subsets with human label variation, human soft-labels enhance calibration and stabilize training convergence across runs. This alignment ensures model uncertainty tracks human label variation on visually ambiguous/difficult samples. Although human soft-labels are more costly than standard labeling, we show that the 50-annotation standard \cite{peterson_2019_iccv} is not a strict requirement; an average of 6 high-quality labels, consistent with \cite{collins_eliciting_2022}, suffices to drive significant calibration gains.

\textbf{Synthetic proxies: } 
As the field shifts towards automated data labeling, our audit of a widely-adopted synthetic digit dataset provides a counter-example to the assumption that synthetic labels serve as a drop-in replacement for human perception. In our testbed experiments, we observe that while model-generated labels capture latent statistical regularities, they fail to mirror human judgment. Specifically, the lack of correlation between synthetic signals and human uncertainty during training suggests that these synthetic proxies fail to encode human uncertainty. This divergence underscores a continued, specialized role for human-grounded evaluation.

\textbf{Limitations:} We deliberately restricted our testbed to simple digits and smaller models to isolate the effects of differing target encodings eliminating visual confounders and preventing memorization of large-scale architectures. As such we study pure perceptual ambiguity, excluding  external noise like occlusion found in real-world data. Thus, these results should not be interpreted as evidence that human soft-labels will lead to similar results in more complex natural images, specialised domains like medical imaging, or natural language tasks, which contain different ambiguity profiles. Some initial results on data beyond digits suggests a similar pattern but needs further validation, which can be found in Appendix \ref{app:cifar-s}. Furthermore, our synthetic target findings are bound to the datasets specific generation method; whether advanced methods better approximate human uncertainty remains open. Our aggregation methodology abstracts away individual differences in annotators' `unsure' thresholds.

\textbf{Broader impact and future work:} 
Our datasets serve as a controlled environment for studying human vs. model uncertainty and optimizing model behavior under ambiguous conditions. By providing a low-compute entry point for exploring ambiguity and noise handling, this work enables researchers to investigate how models mirror (or diverge from) human perceptual limits. 
Our fine-grained annotations and observations contribute to the development of AI systems that can signal well-calibrated caution in ambiguous cases, which is important for better trust in AI. However, results on these datasets should be treated strictly as a research testbed; high performance is not a direct proxy for real-world reliability. Future work should validate whether these uncertainty dynamics remain fundamental in non-visual modalities and billion-parameter models, while also exploring advanced aggregation and label noise detection methods within this detailed labeling regime.

\textbf{Acknowledgments and disclosures: }
The creation of the dataset was supported with funds from Queen Mary University; the authors have no other funding or competing interests to disclose.

\bibliographystyle{plainnat}
\bibliography{neurips}

\newpage
\appendix

\section{Data Collection Process} \label{appendix_data_collection} 

\subsection{Corpus size}
To establish a lower bound for the number of training instances per class required to achieve reliable performance, we conducted a preliminary sensitivity analysis using the MNIST benchmark \cite{lecun2002gradient}. This analysis serves to justify the scale of our data collection and re-annotation efforts. We used a LeNet architecture \cite{lecun2002gradient} optimized via Adam ($\eta = 10^{-3}$) 
with a batch size of 64 over 25 epochs. To ensure statistical significance, we performed 5-fold repetitions for each subset size. Subsets were constructed via stratified random sampling without replacement, evaluating class-specific counts of $n \in \{50, 100, 150, 200, 300, 500\}$. 
Our empirical findings indicate that 150 instances per class typically sufficed to reach approximately 95\% test accuracy (detailed Table \ref{tab:test_acc}). Consequently, we adopt $n \geq 150$ as the target threshold for our corpus construction to ensure sufficient model capacity and performance stability.  

\begin{table}[h]
    \centering
    \small
    \renewcommand{\arraystretch}{1} 
    \caption{Test Accuracy Results by Sample Size per Digit}
    \label{tab:test_acc}
    \vspace{0.5em}
    \begin{tabular}{rccccccc}
        \toprule
        Samples per digit (class): & 50 & 100 & \textbf{150} & {300} & 500 & 1000 & 2000 \\
        \midrule
        Avg. Acc. & 89.584 & 93.982 & \textbf{95.272} & {96.590} & 97.324 & 98.036 & 98.548 \\
        Avg. Std. & 1.2023 & 0.1955 & \textbf{0.2537} & {0.3102} & 0.3754 & 0.2592 & 0.1506 \\
        \bottomrule
    \end{tabular}
\end{table}

\subsection{Dataset candidates}
\label{appendix-dataset-candidates}
To investigate model behavior across the full spectrum of sample difficulty, we evaluate four digit-based datasets: the standard MNIST \citep{lecun2002gradient}, the Swedish historical handwritten dataset ARDIS \citep{kusetogullari2020ardis}, and two synthetic variants of Ambiguous-MNIST \citep{mukhoti_2023_ddu, weiss2023generating}. We de-duplicate all datasets to ensure evaluation integrity. While MNIST contains no duplicates (70k images), we identified and removed 126 duplicates from ARDIS (7,474 remaining) and 1,392 from the Ambiguous-MNIST \textit{(distributional targets loaded without noise)} variant from \cite{mukhoti_2023_ddu} (10,608 remaining). For the variant by \cite{weiss2023generating}, we restrict our scope to the 10k training set to avoid the overlap found in its test split. 
In Mukhoti's Ambiguous-MNIST images are generated by decoding a linear combination of latent representations from two different digits using a pre-trained Variational Autoencoder \cite{mukhoti_2023_ddu}. Each image is assigned a label by sampling from the softmax probabilities of a pre-existing MNIST model, then filtered by mutual information and stratified by entropy to ensure a diverse and high-quality range of ambiguity \cite{mukhoti_2023_ddu}. Weiss, on the other hand, train a Regularized Adversarial Autoencoder; then use the rAAE's decoder to construct the ambiguous image and its discriminator to obtain the probabilistic labels \cite{weiss2023generating}.

Standard datasets such as MNIST are commonly dominated by 'easy' instances that lead to rapid loss convergence, however, work on annotator uncertainty \citep{peterson_2019_iccv} and dataset cartography \citep{swayamdipta2020dataset} suggests that a model’s ability to generalize is more closely tied to its performance on 'ambiguous' and 'hard' instances. Rather than relying on the skewed distribution of the original sources, we construct datasets balanced across the difficulty spectrum. Specifically, we employ Dataset Cartography \citep{swayamdipta2020dataset} as a filtering mechanism to identify high-information candidates for human labeling across easy, ambiguous, and hard regions. We adopt this over standard confidence-based sampling, as Data Maps have been shown to outperform simple uncertainty metrics in active learning and generalization contexts \citep{zhang_plank2021_cart_AL}.

\begin{figure}[ht!]
    \centering %
    \begin{minipage}{0.57\textwidth}
        \begin{subfigure}{\linewidth}
            \includegraphics[width=\linewidth]{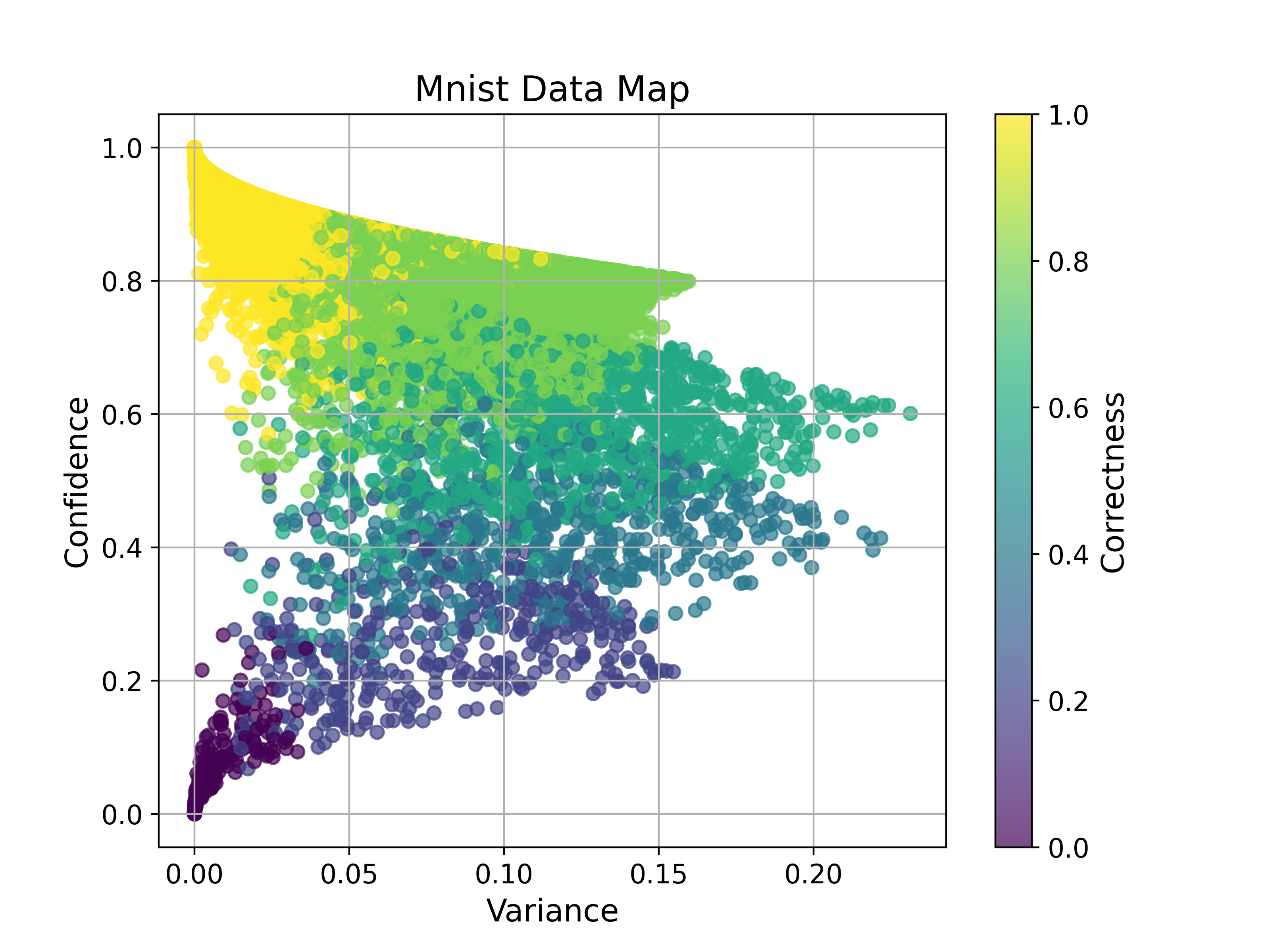}
            \label{fig:sub1}
        \end{subfigure}
    \end{minipage}
    \hfill 
    \begin{minipage}{0.42\textwidth}
        \begin{subfigure}{\linewidth}
            \includegraphics[width=\linewidth]{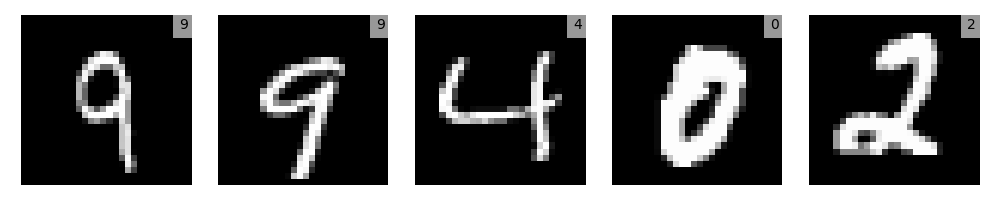}
            \caption{\small Considered 'easy' by training dynamics}
            \label{fig:sub2}
        \end{subfigure}
        
        \begin{subfigure}{\linewidth}
            \includegraphics[width=\linewidth]{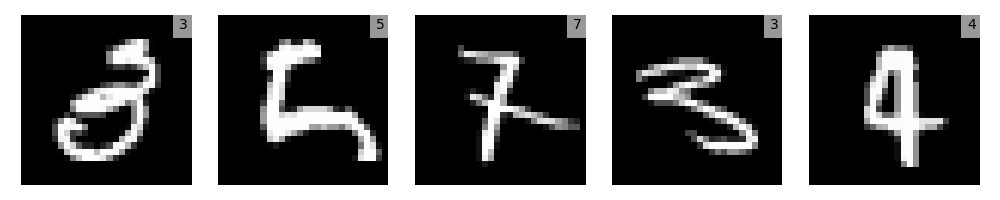}
            \caption{\small Considered 'hard' by training dynamics}            
            \label{fig:sub3}
        \end{subfigure}
        
        \begin{subfigure}{\linewidth}
            \includegraphics[width=\linewidth]{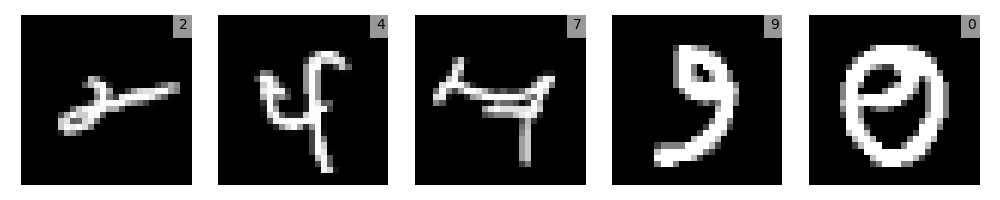}
            \caption{\small Considered 'ambig.' by training dynamics}
            \label{fig:sub4}
        \end{subfigure}
    \end{minipage}
    \vspace{-0.8em}
    \caption{Cartography Map for MNIST using a simple feed-forward neural network for 5 epochs}
    \label{mnist_datacollection_cart}
\end{figure}

\begin{figure}[ht!]
    \centering 
    \begin{minipage}{0.57\textwidth}
        \begin{subfigure}{\linewidth}
            \includegraphics[width=\linewidth]{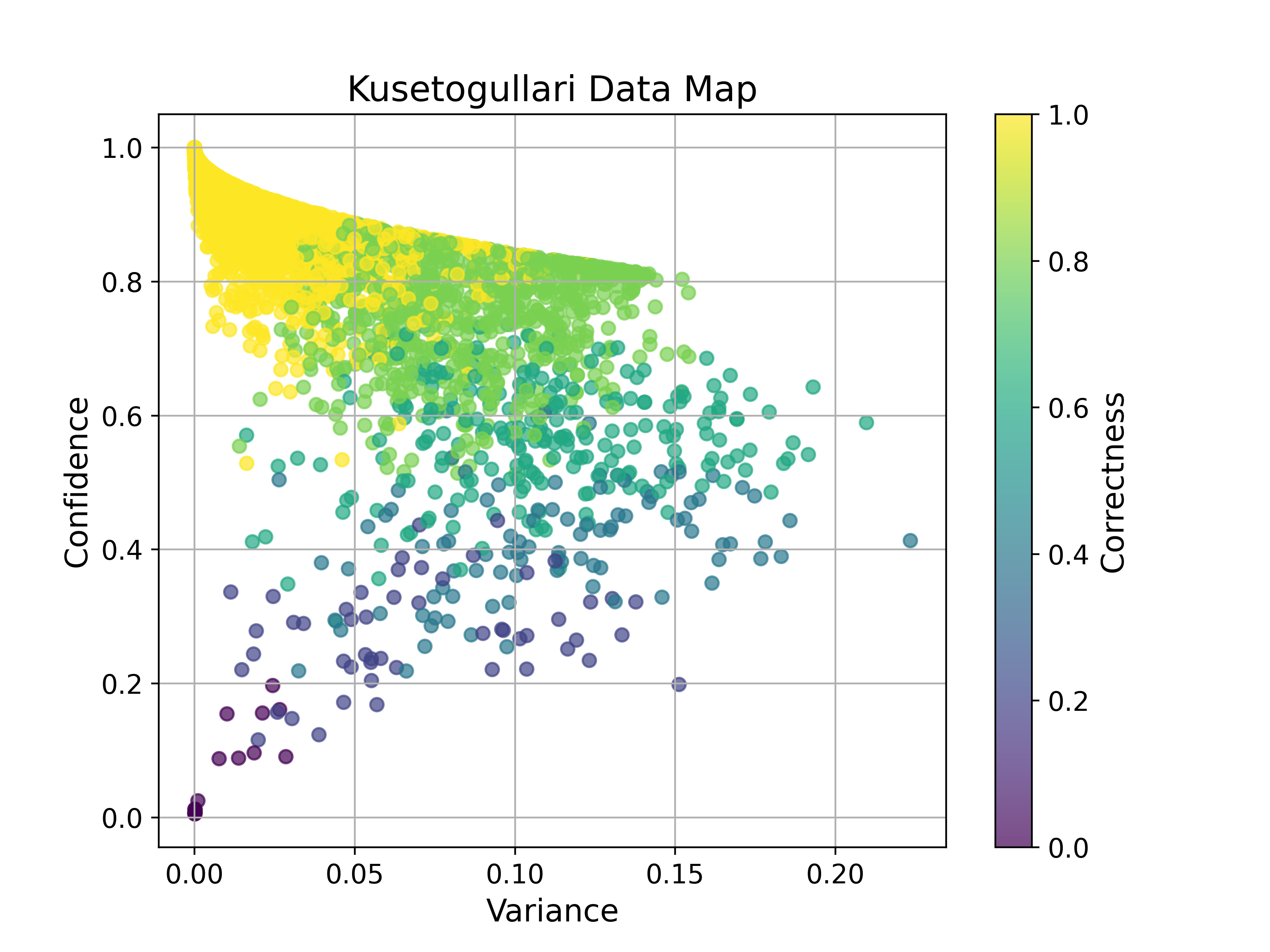}
            \label{fig:sub1}
        \end{subfigure}
    \end{minipage}
    \hfill 
    \begin{minipage}{0.42\textwidth}
        \begin{subfigure}{\linewidth}
            \includegraphics[width=\linewidth]{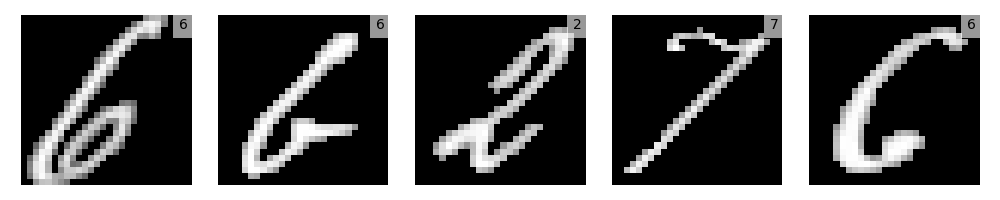}
            \caption{\small Considered 'easy' by training dynamics}
            \label{fig:sub2}
        \end{subfigure}
        
        \begin{subfigure}{\linewidth}
            \includegraphics[width=\linewidth]{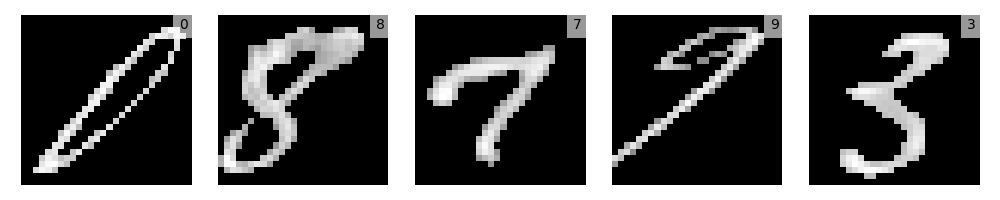}
            \caption{\small Considered 'hard' by training dynamics}            
            \label{fig:sub3}
        \end{subfigure}
        
        \begin{subfigure}{\linewidth}
            \includegraphics[width=\linewidth]{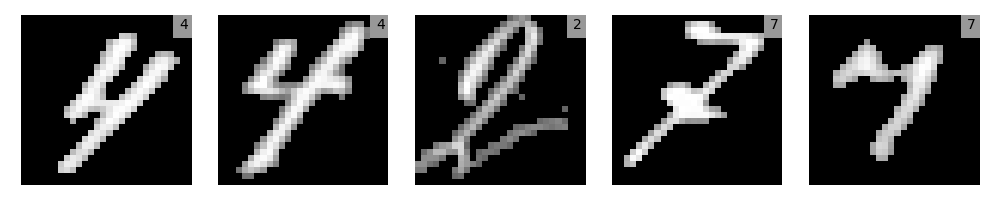}
            \caption{\small Considered 'ambig.' by training dynamics}
            \label{fig:sub4}
        \end{subfigure}
    \end{minipage}
    \vspace{-0.8em}
    \caption{Cartography Map for ARDIS using a simple  feed-forward neural network  for 5 epochs%
    }
\end{figure}

\begin{figure}[ht!]
    \centering %
    \begin{minipage}{0.57\textwidth}
        \begin{subfigure}{\linewidth}
            \includegraphics[width=\linewidth]{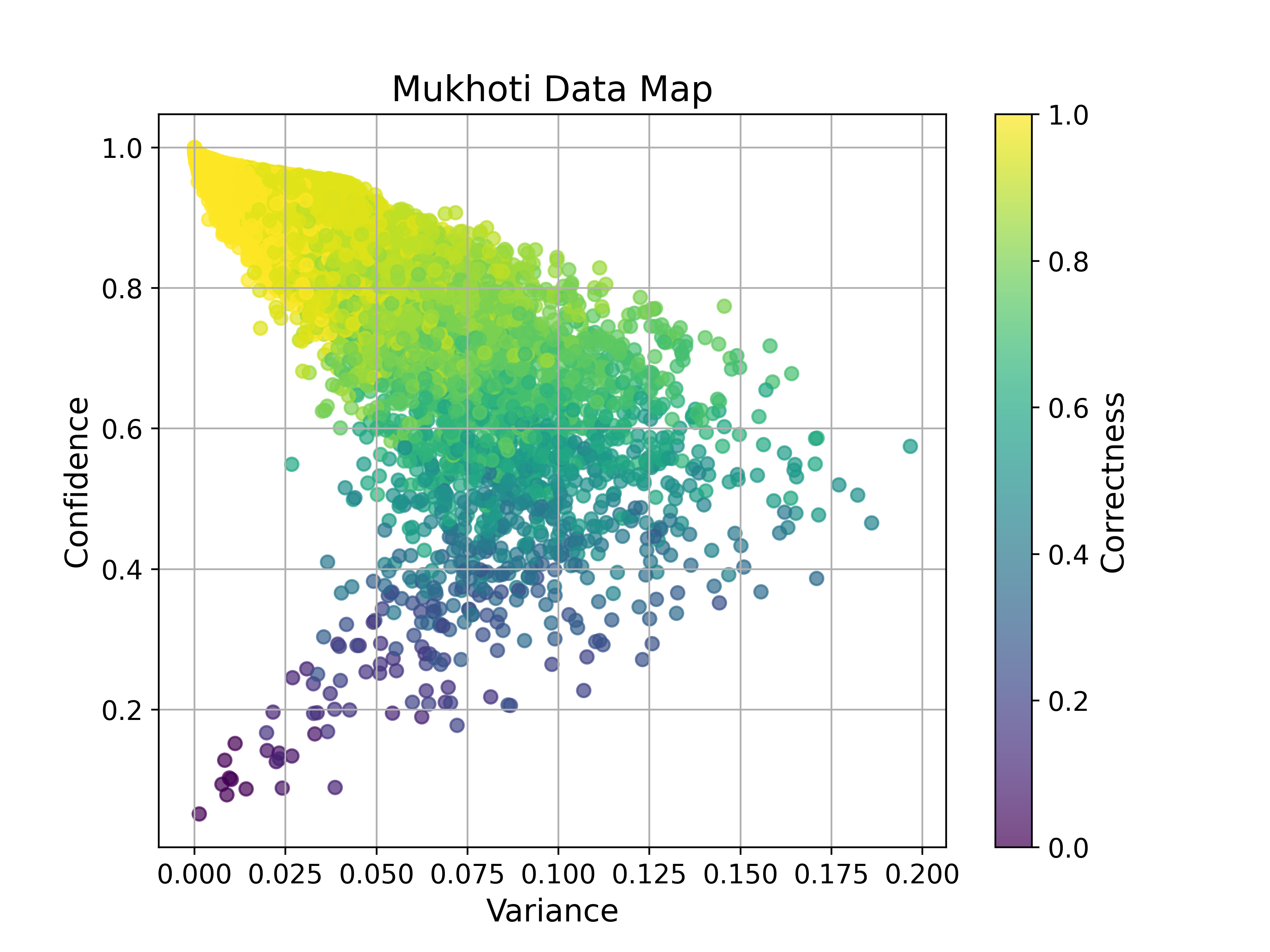}
            \label{fig:sub1}
        \end{subfigure}
    \end{minipage}
    \hfill %
    \begin{minipage}{0.42\textwidth}
        \begin{subfigure}{\linewidth}
            \includegraphics[width=\linewidth]{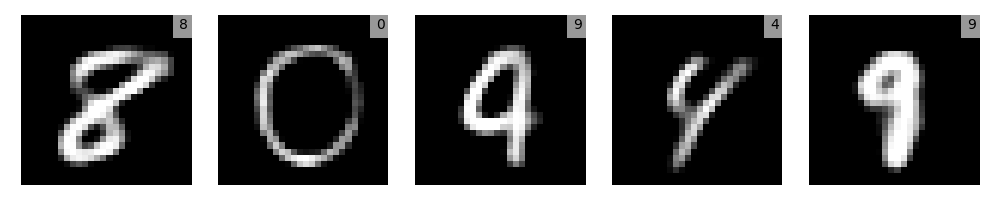}
            \caption{\small Considered 'easy' by training dynamics}
            \label{fig:sub2}
        \end{subfigure}
        
        \begin{subfigure}{\linewidth}
            \includegraphics[width=\linewidth]{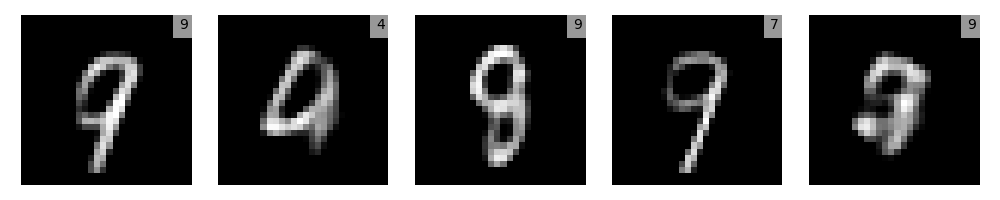}
            \caption{\small Considered 'hard' by training dynamics}            
            \label{fig:sub3}
        \end{subfigure}
        
        \begin{subfigure}{\linewidth}
            \includegraphics[width=\linewidth]{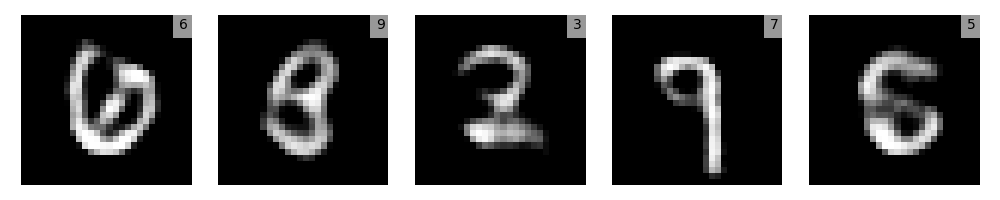}
            \caption{\small Considered 'ambig.' by training dynamics}
            \label{fig:sub4}
        \end{subfigure}
    \end{minipage}
    \vspace{-0.8em}
    \caption{Cartography Map for  Mukhoti using a simple  feed-forward neural network  for 20 epochs}
\end{figure}

\begin{figure}[ht!]
    \centering %
    \begin{minipage}{0.57\textwidth}
        \begin{subfigure}{\linewidth}
            \includegraphics[width=\linewidth]{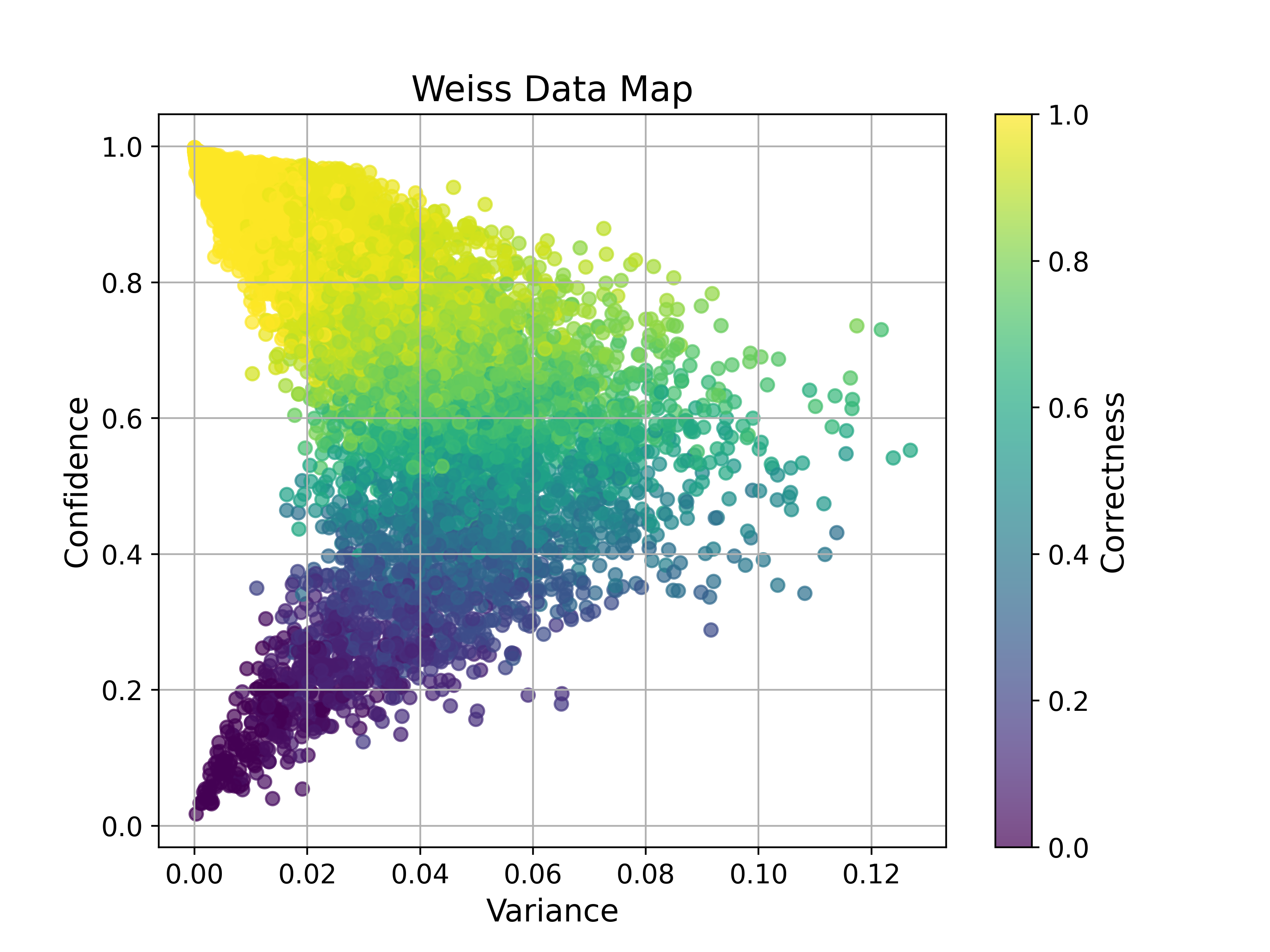}
            \label{fig:sub1}
        \end{subfigure}
    \end{minipage}
    \hfill 
    \begin{minipage}{0.42\textwidth}
        \begin{subfigure}{\linewidth}
            \includegraphics[width=\linewidth]{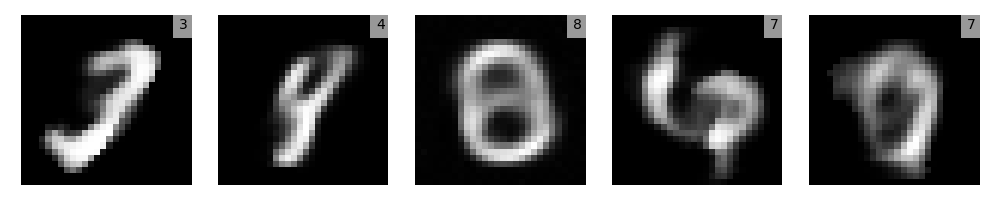}
            \caption{\small Considered 'easy' by training dynamics}
            \label{fig:sub2}
        \end{subfigure}
        
        \begin{subfigure}{\linewidth}
            \includegraphics[width=\linewidth]{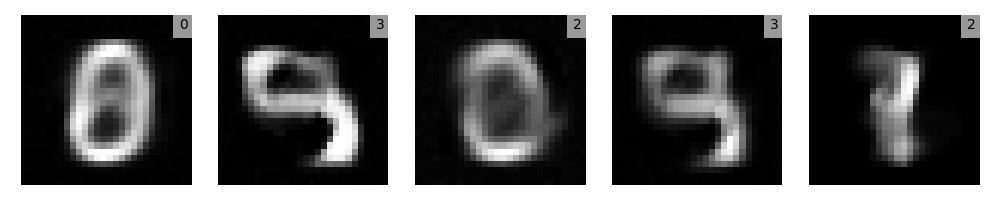}
            \caption{\small Considered 'hard' by training dynamics}            
            \label{fig:sub3}
        \end{subfigure}
        
        \begin{subfigure}{\linewidth}
            \includegraphics[width=\linewidth]{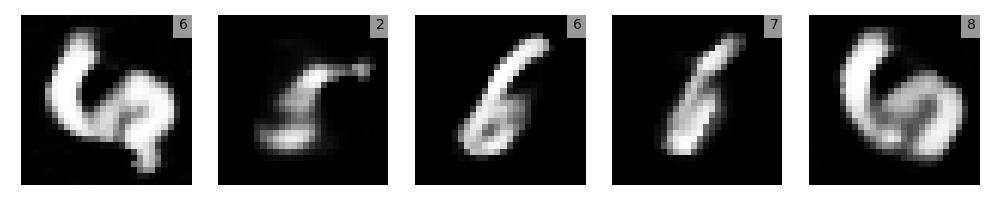}
            \caption{\small Considered 'ambig.' by training dynamics}
            \label{fig:sub4}
        \end{subfigure}
    \end{minipage}
    \vspace{-0.8em}
    \caption{Cartography Map for  Weiss using a simple  feed-forward neural network  for 30 epochs}
\end{figure}

\newpage


\begin{figure}[ht!]
    \centering
    \begin{minipage}{0.4\textwidth}
        \centering
        \includegraphics[width=0.72\linewidth]{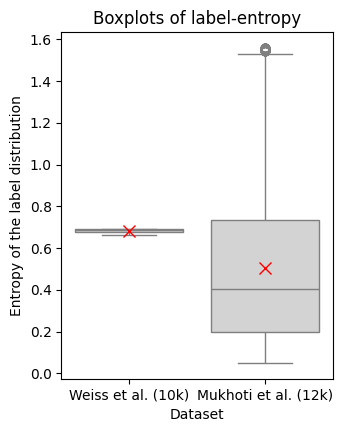}
        \caption{Entropy comparison: Weiss shows higher label entropy, while Mukhoti has a broader, more varied distribution.}
        \label{fig:mukhoti_vs_weiss_entropy}
    \end{minipage}
    \hfill
    \begin{minipage}{0.58\textwidth} 
    \centering
    \includegraphics[width=0.52\linewidth]{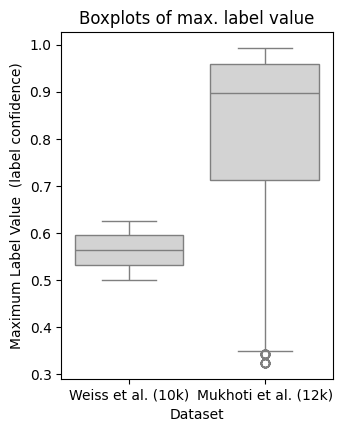}
    \caption{Distribution of maximum confidence scores for Mukhoti and Weiss. Mukhoti samples show higher overall confidence, while Weiss samples center around a 0.55 probability.
    } 
    \label{fig:mukhoti_vs_weiss_maxlabel}
\end{minipage}
\end{figure}

\subsection{Dataset construction} 
To construct the corpus, we first process all four datasets by removing duplicate images across the corpora. We strictly filtered for exact image duplicates, not near-duplicates. This de-duplication mitigates the train-test leakage we observed in three of the four candidate datasets \cite{kusetogullari2020ardis,weiss2023generating, mukhoti_2023_ddu}, ensuring a cleaner signal for evaluation.
Following de-duplication, we perform cartography mapping using a simple one-hidden-layer feed-forward network (see \ref{appendix-dataset-candidates}). 
ARDIS has only 7,474 images, and a simple LeNet learns them all easily, so the difficulty distribution is too narrow for our analysis of hard/ambiguous cases. The Weiss dataset has more samples, but on inspection doesn’t reflect naturally occurring ambiguous digits: the majority are synthetic combinations of $\sim$50\% one digit's pixels with 50\% of another's, thus not naturally ambiguous written digits (see example images in \ref{appendix-dataset-candidates}). We select MNIST and Mukhoti's \cite{mukhoti_2023_ddu} distributional Ambiguous-MNIST as our primary sources. 
MNIST is chosen over ARDIS because its larger scale (70k vs 7,474 images) provides a richer tail of naturally occurring hard and ambiguous cases.
Between the synthetic datasets, \cite{mukhoti_2023_ddu} offers a more comprehensive distribution across the difficulty spectrum and better adheres to the natural data manifold compared to \cite{weiss2023generating}.

We sample subsets of MNIST and Mukhoti for re-labeling from the cartography regions. We adopt a training horizon of 5 epochs for MNIST and 20 epochs for Ambiguous-MNIST. By limiting MNIST epochs, we prevent the dominant easy samples from reaching total convergence, thereby preserving the signal required to identify the "long tail" of harder cases. 
The difficulty regions are defined by thresholds on confidence ($\mu$) and variability ($\sigma$).
MNIST is filtered according to the following setup: 
Easy: high-confidence, stable instances ($\mu > 0.7, \sigma < 0.125$).
Hard: consistently misclassified instances ($\mu < 0.3, \sigma < 0.125$).
Ambiguous: high-variability instances ($0.3 \leq \mu \leq 0.7, \sigma > 0.125$). Ambiguous-MNIST is filtered according to the following setup:
Easy: high-confidence, stable instances ($\mu > 0.7, \sigma < 0.1$).
Hard: consistently misclassified instances ($\mu < 0.3, \sigma < 0.1$).
Ambiguous: high-variability instances ($0.3 \leq \mu \leq 0.7, \sigma > 0.1$). The use of different variability thresholds was an empirical adjustment due to somewhat different learning dynamics of the two datasets. Because Mukhoti is inherently more difficult and contains more complex ambiguity, it required more epochs to converge into stable cartography clusters compared to MNIST. The absolute scale of confidence variability across epochs differs slightly between the two. We calibrated the thresholds empirically for each dataset's specific cartography distribution to ensure a sufficient, representative sample size across all three categories (easy, hard, and ambiguous).

The resulting datasets are partitioned into training, validation, and test sets using stratified sampling to ensure the difficulty distribution remains consistent across splits. 
To maintain sufficient class representation and to ensure sufficient model capacity and stability, we enforce a constraint of at least 150 `easy’ instances per digit in the training set, a threshold empirically derived from a LeNet sensitivity analysis on MNIST (see section "A.1 Corpus size above" for full experimental details). The final MNIST subset comprises 2,131 training, 457 validation, and 457 test samples. The Ambiguous-MNIST (Mukhoti) subset consists of 1,738 training, 373 validation, and 373 test sample.

\section{Annotation}
\label{appendix_annotation}

We recruited 160 individual annotators from Prolific from three regions (France, Germany and the USA). We selected by nationality and the annotators first language being the same as nationality but also selected for fluency in English, with task instructions being in English. To make sure we only include high quality annotators we filtered for a previous approval rate of 99-100\% and integrated attention checks into each task as an additional quality measure. We included three unambiguous, gold-standard clean digits into the task and only retained labels from annotators who correctly identified all three of these control images.
We hypothesized there might be regional variations in handwriting styles (e.g., how a "1" vs. "7" or "4" vs. "6" is written and perceived between the USA vs. France or Germany). However, our analysis revealed no notable regional differences. We observed regional variance in completion rates; however, the average median completion time was 14.59 minutes (approx. £15.65/hr). Regarding data integrity, we retain all collected annotations, excluding only a small subset of samples from a single annotator who flagged a display failure during those specific images.

\begin{figure}[h!]
    \centering
    \begin{minipage}{0.62\textwidth}
        \centering
        \includegraphics[width=\linewidth]{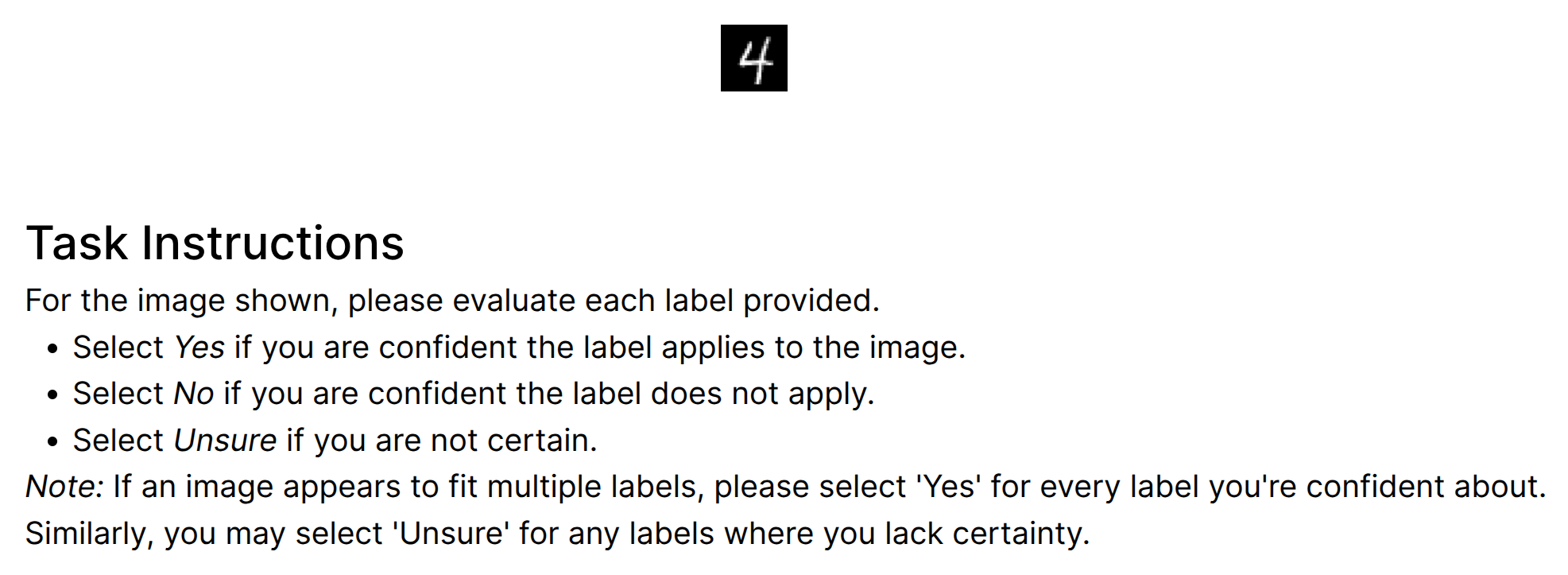}
        \caption{Task instruction screen}
    \end{minipage}
    \hfill
    \begin{minipage}{0.37\textwidth}
        \centering
        \includegraphics[width=\linewidth]{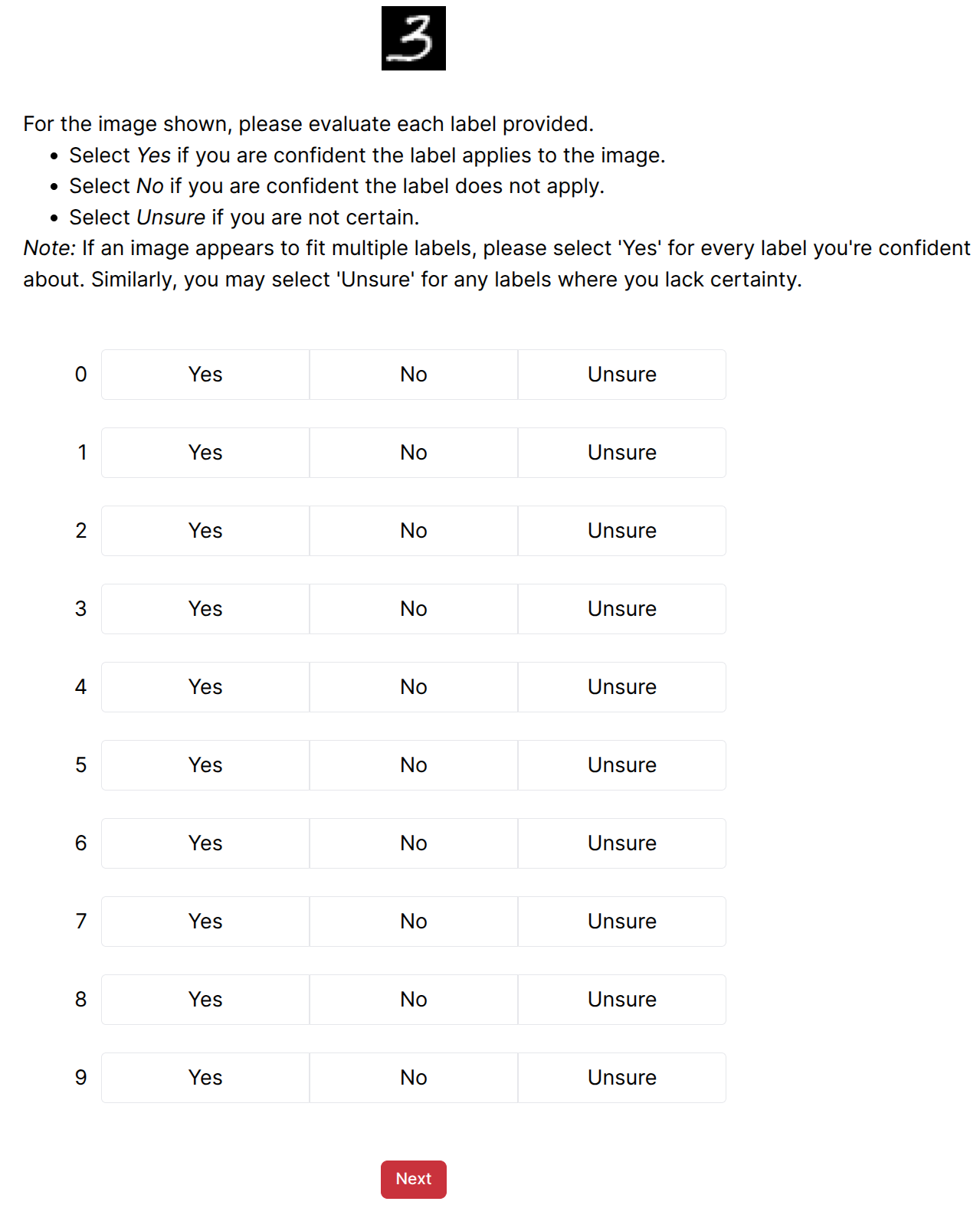}
        \caption{Example screen of digit labeling}
    \end{minipage}
\end{figure}

The aim was to distribute images evenly, targeting at least two annotators from each country per image. However, due to platform routing and the filtering of annotators who failed the attention checks, the final dataset contains an average of 6 annotations per image (ranging overall from 3 to 10).

\begin{table}[h]
\centering
\caption{Dataset annotation distribution: number of samples by annotator count; on average samples have around 6 annotators}
\label{tab:annotation_dist}
\small
\vspace{0.4em}
\renewcommand{\arraystretch}{.8}
\begin{tabular}{lcccccccc}
\toprule
\textbf{Number of Annotations per Image} & 3 & 4 & 5 & 6 & 7 & 8 & 9 & 10 \\ \midrule
\textbf{Number of Images}      & 26 & 341 & 849 & 2620 & 1399 & 271 & 14 & 10 \\ \bottomrule
\end{tabular}
\end{table}

\section{Aggregation}
\label{appendix_aggregation}
\textbf{Annotator-level aggregation:}
Let the dataset be $\mathcal{D}=\left\{\left(x_n, Z_n\right)\right\}_{n=1}^N$, where $x_n$ is an image and $Z_n$ is its corresponding set of uncertainty-aware annotations. This structure is adapted from the concept of Basic Uncertain Information (BUI) \cite{mesiar2017bui}, which pairs a judgment with its reliability. We have a total pool of $M$ annotators, $\mathcal{A}=\left\{a_m\right\}_{m=1}^M$. For each image $x_n$, its annotation set $Z_n=\left\{\mathbf{z}_{n, m} \mid a_m \in \mathcal{A}_n\right\}$ is provided by a subset of annotators $\mathcal{A}_n \subseteq \mathcal{A}$, where $3 \leq\left|\mathcal{A}_n\right| \leq 10$.

Each individual certainty-aware annotation $\mathbf{z}_{n, m}$ is a pair $\left(\mathbf{y}_{n, m}, u_{n, m}\right)$ $\in \Delta^{10} \times[0,1]$. The first element, $\mathbf{y}_{n, m}$, is a probability distribution over $K=11$ possible labels, such that $\mathbf{y}_{n, m} \in \Delta^{K-1}$, where $\Delta^{K-1}$ is the ($K-1$)-simplex. The second element,
$u_{n, m} \in[0,1]$, is an uncertainty score representing the annotator's uncertainty in their provided annotation for that image.

\textbf{Annotator-level soft-label ($\mathbf{y}_{n, m}$): }
We transform raw multi-label selections (Yes, No, Unsure) into a probability distribution $\mathbf{y}_{n, m}$ over $K$ classes. We compare two weighting schemes: 
\begin{itemize}[nosep]
    \item Equally-weighted ($soft_e$): both "Yes" and "Unsure" selections are assigned a weight of 1.
    \item Uncertainty-weighted ($soft_w$): "Unsure" selections are down-weighted to 0.5 to reflect lower class-likelihood, while "Yes" remains 1. 
\end{itemize}
In case an annotator confidently selects "No" for all 10 classes, then we assign 1 to a Nan class, signifying that the sample in question doesn't resemble any of the digits; hence 11 possible labels. The resulting weights are normalized to the $(K-1)$-simplex.

\textbf{Annotator-level uncertainty ($u_{n, m}$):}
We derive an implicit uncertainty score $u_{n, m} \in [0, 1]$ based on annotator behavior. 
For each annotator $a_m$ and sample $x_n$, $u_{n, m}$ is defined as the ratio of "Unsure" selections to the total number of label options $K$. Thus, $u_{n, m} = 0$ denotes maximum confidence, while $u_{n, m} = 0.33$ indicates the annotator was unsure about one-third of the possible classes. 

\textbf{Image-level soft-label ($\mathbf{y}_{n}$):} To consolidate individual judgments into an image-level representation, we compute an aggregated annotation pair $\mathbf{z}_n = ( \overline{\mathbf{y}}_n, u_n)$ for each image $x_n$.  Given the set of individual annotations $Z_n = \{(\mathbf{y}_{n, m}, u_{n, m}) \mid a_m \in \mathcal{A}_n\}$, the aggregated soft-label is calculated via arithmetic mean of individual distributions: $\overline{\mathbf{y}}_n = \frac{1}{|\mathcal{A}_n|} \sum_{a_m \in \mathcal{A}_n} \mathbf{y}_{n, m}$. 

\textbf{Image-level annotator uncertainty ($u_{n}$):} To estimate the collective uncertainty $u_n$ for a sample $x_n$, we propose two aggregation proxies:
\begin{itemize}[nosep]
    \item  Average uncertainty ($u_{n}^{\text{mean}}$): The average uncertainty across the annotators, calculated as: $u_{n}^{\text{mean}} = \frac{1}{|\mathcal{A}_n|} \sum_{a_m \in \mathcal{A}_n} u_{n, m} \in [0, 1]$
    \item  Unsure proportion ($u_{n}^{\text{prop}}$): The fraction of annotators who expressed uncertainty at least once for a given sample: $u_{n}^{\text{prop}} = \frac{1}{|\mathcal{A}_n|} \sum_{a_m \in \mathcal{A}_n} \mathbb{I}(u_{n,m} > 0)   \in [0, 1] $, where $\mathbb{I}(\cdot)$ is the indicator function.
\end{itemize}
 While $u_{n}^{\text{mean}}$ measures the  magnitude of doubt, $u_{n}^{\text{prop}}$ captures the consensus of doubt across the annotator pool for a given sample n.

\newpage
\section{Annotated dataset}
\label{appendix_dataset_stats}

\begin{figure}[h!]
    \centering
    \begin{minipage}{0.48\textwidth}
        \centering
        \includegraphics[width=\linewidth]{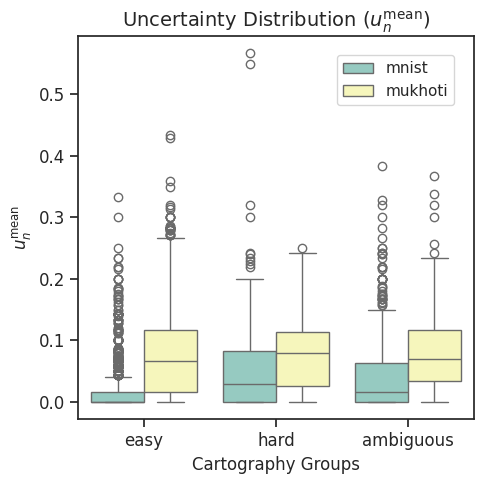}
    \end{minipage}
    \hfill
    \begin{minipage}{0.48\textwidth}
        \centering
        \includegraphics[width=\linewidth]{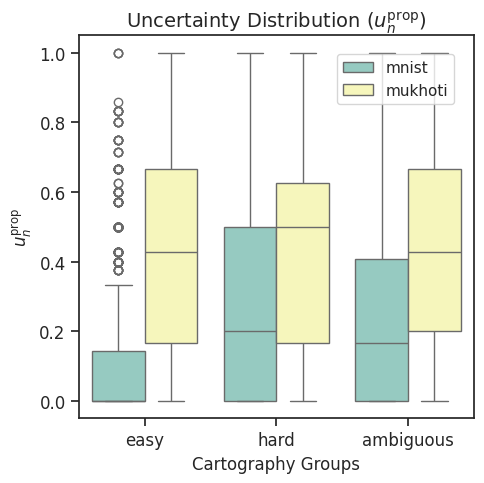}
    \end{minipage}
    \caption{Distribution of image-level uncertainty. Comparison of uncertainty metrics $u_n^{\text{mean}}$ (left) and $u_n^{\text{prop}}$ (right) across MNIST and Mukhoti datasets. Samples are stratified into "easy," "hard," and "ambiguous" groups based on the initial Dataset Cartography \cite{swayamdipta2020dataset} used for filtering. Mukhoti consistently exhibits higher median uncertainty and greater variance across all categories compared to MNIST.}
\end{figure}

\begin{figure}[h!]
    \centering
    \begin{minipage}{0.48\textwidth}
        \centering
        \includegraphics[width=\linewidth]{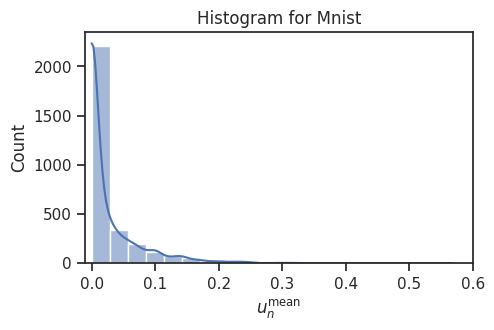}
    \end{minipage}
    \hfill
    \begin{minipage}{0.48\textwidth}
        \centering
        \includegraphics[width=\linewidth]{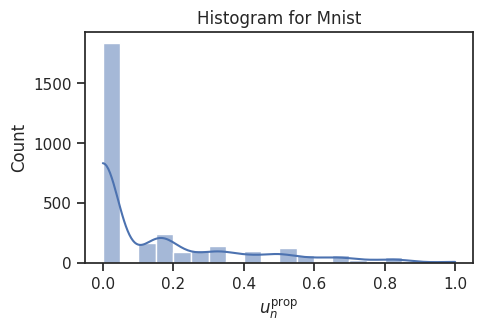}
    \end{minipage}
    \begin{minipage}{0.48\textwidth}
        \centering
        \includegraphics[width=\linewidth]{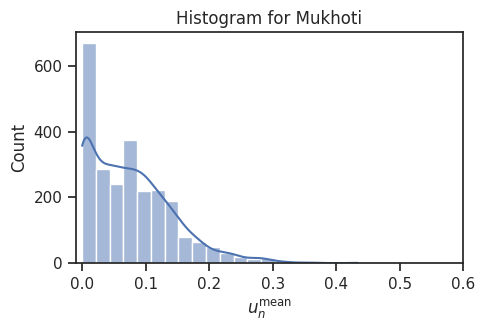}
        \caption{Distribution of $u_n^{\text{mean}}$ image-level uncertainty. Histogram  for MNIST (top) and Mukhoti (bottom). MNIST shows a sharp concentration near zero, reflecting high annotator consensus, whereas Mukhoti shows a significantly heavier tail and broader uncertainty distribution.}
    \end{minipage}
    \hfill
    \begin{minipage}{0.48\textwidth}
        \centering
        \includegraphics[width=\linewidth]{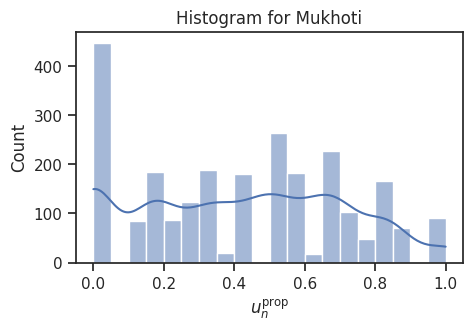}
        \caption{Distribution of $u_n^{\text{prop}}$ image-level uncertainty. Histogram comparing the $u_n^{\text{prop}}$ metric for MNIST (top) and Mukhoti (bottom). The Mukhoti dataset shows more high-uncertainty samples ($u_n^{\text{prop}} > 0.5$) compared to MNIST. }
    \end{minipage}
\end{figure}

\paragraph{Inter-annotator agreement:} we report Krippendorff's alpha (nominal) given that each item receives a variable number of ratings and not every annotator rates every item. Computed over each annotator's primary single-digit label (the digit marked "yes" or ”unsure” when exactly one was chosen; ratings with multiple digits are excluded, as there is no rank in our data per annotator rating).

\begin{table}[ht]
\centering
\caption{Inter-annotator and label agreement statistics across datasets.}
\label{tab:inter_annotator_agreement}
\begin{tabular}{lcc}
\toprule
\textbf{Dataset} & \textbf{Krippendorff's $\alpha$} \small (raw annotators) & \textbf{Cohen's $\kappa$} (original vs.\ new label) \\
\midrule
MNIST            & 0.973                            & 0.957 \\
Mukhoti          & 0.835                            & 0.630 \\
\bottomrule
\end{tabular}
\end{table}

A caveat applies to both statistics, for related but distinct reasons. For $\kappa$ on Mukhoti, the original label is a softmax distribution collapsed to its argmax, and the new label is itself an aggregate (the most-voted digit across annotators) collapsed from a vote distribution that may be far from unanimous. In both cases the statistic is computed only over the annotators (and, for Mukhoti's original label, the generation process) that chooses to label an image as one digit; so the true level of disagreement/uncertainty on the most contested items is larger than either number suggests.

\begin{table}[h]
\centering
\small
\caption{Spearman correlation for MNIST (Left) and Mukhoti (Right); correlation between different uncertainty representations. $ms$ is the normalised average miliseconds, $soft_wH$ is the shannon-entropy of the $soft_w$-label, $soft_eH$ is the shannon-entropy of the $soft_e$-label.} 
\vspace{0.6em}
\renewcommand{\arraystretch}{.95} %
\setlength{\tabcolsep}{2pt} %
\begin{tabular}{r|cccccc}
\multicolumn{6}{c}{\textbf{MNIST}} \\
& $ms$ & $u_n^{mean}$ & $u_n^{prop}$ & $soft_wH$ & $soft_eH$ \\ \hline
$ms$ & 1.000 &       &       &       &       \\
$u_n^{mean}$ & 0.388 & 1.000 &       &       &       \\
$u_n^{prop}$ & 0.401 & 0.984 & 1.000 &       &       \\
$soft_wH$ & 0.343 & 0.788 & 0.760 & 1.000 &       \\
$soft_eH$ & 0.345 & 0.792 & 0.767 & 0.999 & 1.000 \\ 
\end{tabular}
\quad 
\begin{tabular}{r|ccccc}
\multicolumn{6}{c}{\textbf{Mukhoti}} \\
& $ms$ & $u_n^{mean}$ & $u_n^{prop}$ & $soft_wH$ & $soft_eH$ \\ \hline
$ms$ & 1.000 &       &       &       &       \\
$u_n^{mean}$ & 0.475 & 1.000 &       &       &       \\
$u_n^{prop}$ & 0.492 & 0.917 & 1.000 &       &       \\
$soft_wH$ & 0.446 & 0.828 & 0.756 & 1.000 &       \\
$soft_eH$ & 0.449 & 0.825 & 0.760 & 0.996 & 1.000 \\ 
\end{tabular}

\vspace{0.5em}
\end{table}

\section{Experimental details}
\label{app:experimental_details}
The optimization parameters were held constant across architectures to ensure that performance variations are attributable to label regimes rather than hyperparameter tuning. Images were normalized using the global mean and standard deviation of their respective training subsets.
\begin{table}[h]
    \centering
    \small
    \caption{Hyperparameters, Optimization Settings, and Normalization Constants}
    \label{tab:hyperparameters}
    \vspace{0.5em}
    \begin{tabular}{ll}
        \toprule
        \textbf{Optimization Parameter} & \textbf{Value} \\
        \midrule
        Optimizer & Adam \\
        Initial Learning Rate ($\eta$) & $10^{-3}$ \\
        Batch Size ($B$) & 64 \\
        Max Epochs & 60 \\
        Loss Function & Cross-Entropy \\
        Training Seeds & $\{32, 12, 86, 10, 34, 99\}$ \\
        \midrule
        \textbf{Dataset Normalization} & \textbf{Mean ($\mu$), Std. Dev. ($\sigma$)} \\
        \midrule
        MNIST & $0.1325, 0.3101$ \\
        Mukhoti & $0.0976, 0.2427$ \\
        \bottomrule
    \end{tabular}
\end{table}
To eliminate non-deterministic variance and ensure seed-wise temporal alignment, we controlled the random state of the environment. The following implementation was used to initialize all training runs:

\begin{lstlisting}[language=Python, caption=seed function]
import torch
import numpy as np
import random

def set_seed(seed=42):
    random.seed(seed)
    np.random.seed(seed)
    torch.manual_seed(seed)
    torch.cuda.manual_seed(seed)
    torch.cuda.manual_seed_all(seed)
    torch.backends.cudnn.deterministic = True
    torch.backends.cudnn.benchmark = False
\end{lstlisting}

All experiments were conducted on a single consumer-grade laptop (12th Gen Intel i7-12700H × 20). A full run of all 6 seeds for the main results completes in approximately 8-10 hours.

\begin{table}[ht]
\caption{$\uparrow$ Accuracy - models trained on MNIST and Mukhoti with different targets and evaluated on original ($orig.$) labels and our human soft-labels ($soft_{w}$) - $orig.$ in the case of Mukhoti means the original synthetic labels, while for MNIST it is the original one-hot labels
}
\label{tab:acc_overall}
\vspace{0.5em}
\centering
\small
{
\renewcommand{\arraystretch}{.8} 
\begin{tabular}{lrcccccc}
\toprule
 & & \multicolumn{2}{c}{\textbf{SimpleFFN}} & \multicolumn{2}{c}{\textbf{DeeperFFN}} & \multicolumn{2}{c}{\textbf{LeNet}} \\
\cmidrule(lr){3-4} \cmidrule(lr){5-6} \cmidrule(lr){7-8}
\textbf{} & \textbf{target} & \textbf{$orig.$} & \textbf{$soft_{w}$} & \textbf{$orig.$} & \textbf{$soft_{w}$} & \textbf{$orig.$} & \textbf{$soft_{w}$} \\ 
\midrule
\multirow{4}{*}{\rotatebox[origin=c]{90}{MNIST}}
    & $orig.$ & 73.16$_{\pm 0.73}$ & 74.47$_{\pm 0.81}$ & 73.05$_{\pm 1.55}$ & 74.00$_{\pm 1.55}$ & 84.54$_{\pm 0.71}$ & 85.01$_{\pm 1.18}$ \\ 
    & $maj._{n}$ & 74.11$_{\pm 0.43}$ & 75.53$_{\pm 0.29}$ & 73.41$_{\pm 0.73}$ & 74.58$_{\pm 0.73}$ & 84.39$_{\pm 0.72}$ & 85.96$_{\pm 0.56}$ \\ 
    & $soft_{w}$ & \textbf{74.40}$_{\pm{0.55}}$ & \textbf{75.97}$_{\pm {0.15}}$ & \textbf{75.13}$_{\pm {0.37}}$ & \textbf{76.44}$_{\pm{0.67}}$ & \textbf{85.41}$_{\pm {0.78}}$ & \textbf{86.69}$_{\pm {0.94}}$ \\ 
    & $soft_{e}$  & 73.52$_{\pm 0.86}$ & 75.16$_{\pm 0.95}$ & 74.87$_{\pm 0.59}$ & 76.40$_{\pm 0.71}$ & 85.41$_{\pm 1.24}$ & 86.69$_{\pm 1.10}$ \\
\midrule
\multirow{4}{*}{\rotatebox[origin=c]{90}{Mukh.}} 
& $orig.$   & \textbf{76.23}$_{\pm 1.17}$ & 57.78$_{\pm 0.89}$ & \textbf{78.91}$_{\pm 1.20}$ & 58.13$_{\pm 0.57}$ & \textbf{87.67}$_{\pm 2.52}$ & 60.46$_{\pm 1.39}$ \\ 
    & $maj._{n}$ & 54.29$_{\pm 1.57}$ & 61.66$_{\pm 1.20}$ & 53.44$_{\pm 2.47}$ & 62.51$_{\pm 2.75}$ & 61.39$_{\pm 1.92}$ & 65.28$_{\pm 1.16}$ \\ 
    & $soft_{w}$ & 56.12$_{\pm 1.51}$ & 65.28$_{\pm 1.90}$ & 55.59$_{\pm 1.71}$ & 67.43$_{\pm 1.51}$ & 63.72$_{\pm 1.82}$ & 71.63$_{\pm 1.69}$ \\ 
    & $soft_{e}$ & 55.32$_{\pm 2.16}$ & \textbf{65.77}$_{\pm 2.02}$ & 55.50$_{\pm 1.82}$ & \textbf{67.56}$_{\pm 2.33}$ & 63.63$_{\pm 2.08}$ & \textbf{72.12}$_{\pm 1.40}$ \\
\bottomrule
\end{tabular}}
\end{table}


\section{Evaluation }
\label{appendix_evaluation}

\subsection{Accuracy - Aggregate Performance Trends}
\label{appendix_evaluation_accuracy}

Table \ref{tab:acc_overall} presents the top-1 accuracy across all model-target combinations. We observe two distinct behaviors: MNIST shows marginal, consistent gains with soft-labels, whereas Mukhoti shows a sharp divergence between synthetic and human evaluation subsets.

\textbf{MNIST:} As detailed in Table~\ref{tab:acc_overall}, global performance on the MNIST dataset remains relatively stable across different target types, scaling predictably as model capacity increases (from SimpleFFN to LeNet). Training on $soft_{w}$ targets leads to consistently better accuracy than training on $orig.$, $maj._{n}$, and $soft_{e}$ targets across all evaluation sets. Models evaluated on the $soft_{w}$ test set consistently achieved higher accuracy (approx. +1\% to +1.5\%) compared to those evaluated on the $orig.$ test set.

\textbf{Mukhoti:} 
The synthetic dataset shows distinct characteristics compared to MNIST. Table \ref{tab:acc_overall} reveals high sensitivity to the target type and a large performance gap between the $orig.$ (synthetic) and $soft_{w}$ (human) evaluation sets. When models are evaluated on the synthetic test partition, synthetic training 
results in  substantially higher accuracy (e.g., 87.67\% for LeNet) compared to soft targets ($\sim$63\%). Conversely, this trend reverses on the human-centric $soft_{w}$ test set: $soft_{e}$ and $soft_{w}$ targets provide better performance ($\sim$72\%), while the synthetic ($orig.$) target degrades to $\sim$60\%. 
The observed performance drop indicates that the inductive biases learned from synthetic supervision do not generalize to human-annotated ground truth. Our stratified analysis in the main papers helps identify where these representations diverge.

\subsection{Calibration}
\label{appendix_evaluation_calibration}
In addition to KLD we capture the brier scores (Table \ref{tab:brierscores}) for all models across the different targets. Looking at the $new$ test set and the HLV subset, similarly to KLD, the best soft-label targets show brier scores that are significantly lower compared to those models trained on $orig./synth.$ or $maj._n$.

\begin{table}[ht]
\caption{$\downarrow$ Brier Score - models trained on MNIST and Mukhoti with different targets and evaluated on original ($orig.$) labels and our $new$ human soft-labels ($soft_{w}$)}
\label{tab:brierscores}
\vspace{0.5em}
\centering
\small
\renewcommand{\arraystretch}{.8} %
\setlength{\tabcolsep}{4pt} %
\begin{tabular}{ccrcccccc}
\toprule
 & & & \multicolumn{2}{c}{\textbf{SimpleFFN}} & \multicolumn{2}{c}{\textbf{DeeperFFN}} & \multicolumn{2}{c}{\textbf{LeNet}} \\
\cmidrule(lr){4-5} \cmidrule(lr){6-7} \cmidrule(lr){8-9}
& \textbf{eval.} & \textbf{target} & \textbf{$No HLV$} & \textbf{$HLV$} & \textbf{$No HLV$} & \textbf{$HLV$} & \textbf{$No HLV$} & \textbf{$HLV$} \\ 
\midrule

\multirow{8}{*}{\rotatebox[origin=c]{90}{MNIST}} & \multirow{4}{*}{\rotatebox[origin=c]{90}{{\scriptsize{$orig.$}}}} & $orig.$ & - & 0.613$_{ \pm 0.022 }$ & - & 0.610$_{ \pm 0.017 }$ & - & 0.425$_{ \pm 0.010 }$ \\ 
    & & $maj._{n}$ & - & 0.611$_{ \pm 0.017 }$ & - & 0.611$_{ \pm 0.008 }$ & - & 0.454$_{ \pm 0.010 }$ \\
    & & $soft_{w}$ & - & \textbf{0.577}$_{ \pm 0.015 }$ & - & 0.554$_{ \pm 0.017 }$ & - & 0.370$_{ \pm 0.024 }$ \\ 
    & & $soft_{e}$ & - & 0.579$_{ \pm 0.013 }$ & - & \textbf{0.552}$_{ \pm 0.017 }$ & - & \textbf{0.366}$_{ \pm {0.024} }$ \\
    \cmidrule{2-9}  
& \multirow{4}{*}{\rotatebox[origin=c]{90}{{\scriptsize$new$}}}
    & $orig.$ & 0.225$_{ \pm 0.006 }$ & 0.399$_{ \pm 0.018 }$ & 0.215$_{ \pm 0.010 }$ & 0.402$_{ \pm 0.013 }$ & 0.094$_{ \pm 0.018 }$ & 0.290$_{ \pm 0.006 }$ \\ 
    & & $maj._{n}$ & 0.217$_{ \pm 0.006 }$ & 0.387$_{ \pm 0.012 }$ & 0.214$_{ \pm 0.015 }$ & 0.382$_{ \pm 0.012 }$ & 0.093$_{ \pm 0.020 }$ & 0.281$_{ \pm 0.008 }$ \\
    & & $soft_{w}$ & \textbf{0.213}$_{ \pm 0.008 }$ & 0.333$_{ \pm 0.009 }$ & \textbf{0.192}$_{ \pm 0.013 }$ & \textbf{0.312}$_{ \pm 0.013 }$ & \textbf{0.073}$_{ \pm 0.011 }$ & 0.189$_{ \pm 0.013 }$ \\ 
    & & $soft_{e}$ & 0.214$_{ \pm 0.008 }$ & \textbf{0.331}$_{ \pm {0.007} }$ & 0.195$_{ \pm 0.010 }$ & \textbf{0.312}$_{ \pm 0.013 }$ & \textbf{0.073}$_{ \pm 0.011 }$ & \textbf{0.183}$_{ \pm {0.012} }$ \\
\midrule
\multirow{8}{*}{\rotatebox[origin=c]{90}{Mukh.}} & \multirow{4}{*}{\rotatebox[origin=c]{90}{{\scriptsize{$orig.$}}}} & $synth.$ & \textbf{0.218}$_{ \pm 0.018 }$ & \textbf{0.187}$_{ \pm 0.009 }$ & \textbf{0.167}$_{ \pm 0.006 }$ & \textbf{0.149}$_{ \pm 0.002 }$ & \textbf{0.079}$_{ \pm 0.007 }$ & \textbf{0.075}$_{ \pm 0.006 }$ \\
    & & $maj._n$ & 0.305$_{ \pm 0.013 }$ & 0.460$_{ \pm 0.012 }$ & 0.303$_{ \pm 0.016 }$ & 0.481$_{ \pm 0.020 }$ & 0.227$_{ \pm 0.014 }$ & 0.411$_{ \pm 0.040 }$ \\
    & & $soft_w$ & 0.243$_{ \pm 0.012 }$ & 0.415$_{ \pm 0.016 }$ & 0.240$_{ \pm 0.007 }$ & 0.419$_{ \pm 0.006 }$ & 0.184$_{ \pm 0.010 }$ & 0.341$_{ \pm 0.013 }$ \\
    & & $soft_e$ & 0.249$_{ \pm 0.010 }$ & 0.423$_{ \pm 0.014 }$ & 0.245$_{ \pm 0.009 }$ & 0.427$_{ \pm 0.011 }$ & 0.186$_{ \pm 0.010 }$ & 0.349$_{ \pm 0.013 }$ \\
    \cmidrule{2-9}
& \multirow{4}{*}{\rotatebox[origin=c]{90}{{\scriptsize$new$}}}
    & $synth.$ & 0.419$_{ \pm 0.026 }$ & 0.337$_{ \pm 0.008 }$ & 0.367$_{ \pm 0.008 }$ & 0.337$_{ \pm 0.007 }$ & 0.297$_{ \pm 0.026 }$ & 0.334$_{ \pm 0.005 }$ \\
    & & $maj._{n}$ & 0.273$_{ \pm 0.016 }$ & 0.263$_{ \pm 0.003 }$ & 0.249$_{ \pm 0.034 }$ & 0.270$_{ \pm 0.016 }$ & 0.134$_{ \pm 0.023 }$ & 0.237$_{ \pm 0.006 }$ \\ 
    & & $soft_{w}$ & \textbf{0.194}$_{ \pm 0.006 }$ & \textbf{0.185}$_{ \pm 0.006 }$ & \textbf{0.160}$_{ \pm 0.013 }$ & \textbf{0.184}$_{ \pm 0.005 }$ & 0.110$_{ \pm 0.015 }$ & \textbf{0.157}$_{ \pm 0.005 }$ \\ 
    & & $soft_{e}$ & 0.205$_{ \pm 0.010 }$ & 0.188$_{ \pm 0.006 }$ & 0.168$_{ \pm 0.008 }$ & 0.188$_{ \pm 0.002 }$ & \textbf{0.104}$_{ \pm 0.012 }$ & \textbf{0.157}$_{ \pm {0.005} }$ \\
\bottomrule
\end{tabular}
\end{table}

\subsection{Training Dynamics}
\label{app:training_dynamics}
To further investigate why $soft_w$ produces better calibration than collapsed representations ($maj_n$, $orig$), we analyze the epoch-wise training dynamics using Jensen-Shannon Divergence (JSD) between the predictive distributions of successive epochs: $\text{JSD}(P_{e_t} \parallel P_{e_{t-1}})$. 
While we use KL Divergence for static ground-truth evaluations, we use JSD for tracking epoch-to-epoch shifts because it is symmetric and bounded within $[0, 1]$.
We observe that models trained on soft targets ($soft_w$) show lower $JSD$ in earlier epochs compared to those trained on majority-voted labels ($maj_n$). This suggests that soft-labels provide a smoother optimization landscape, without the phase transitions visible in the $maj_n$ supervision (esp. on the HLV subset). While both targets eventually converge to a near stable state (near-zero $JSD$), the $soft_w$ models maintain a more consistent trajectory, particularly in the HLV subsets.

\begin{figure}[htbp]
\vspace{-0.5em}
    \centering
    \begin{subfigure}[b]{0.49\textwidth}
        \centering
        \includegraphics[width=\textwidth]{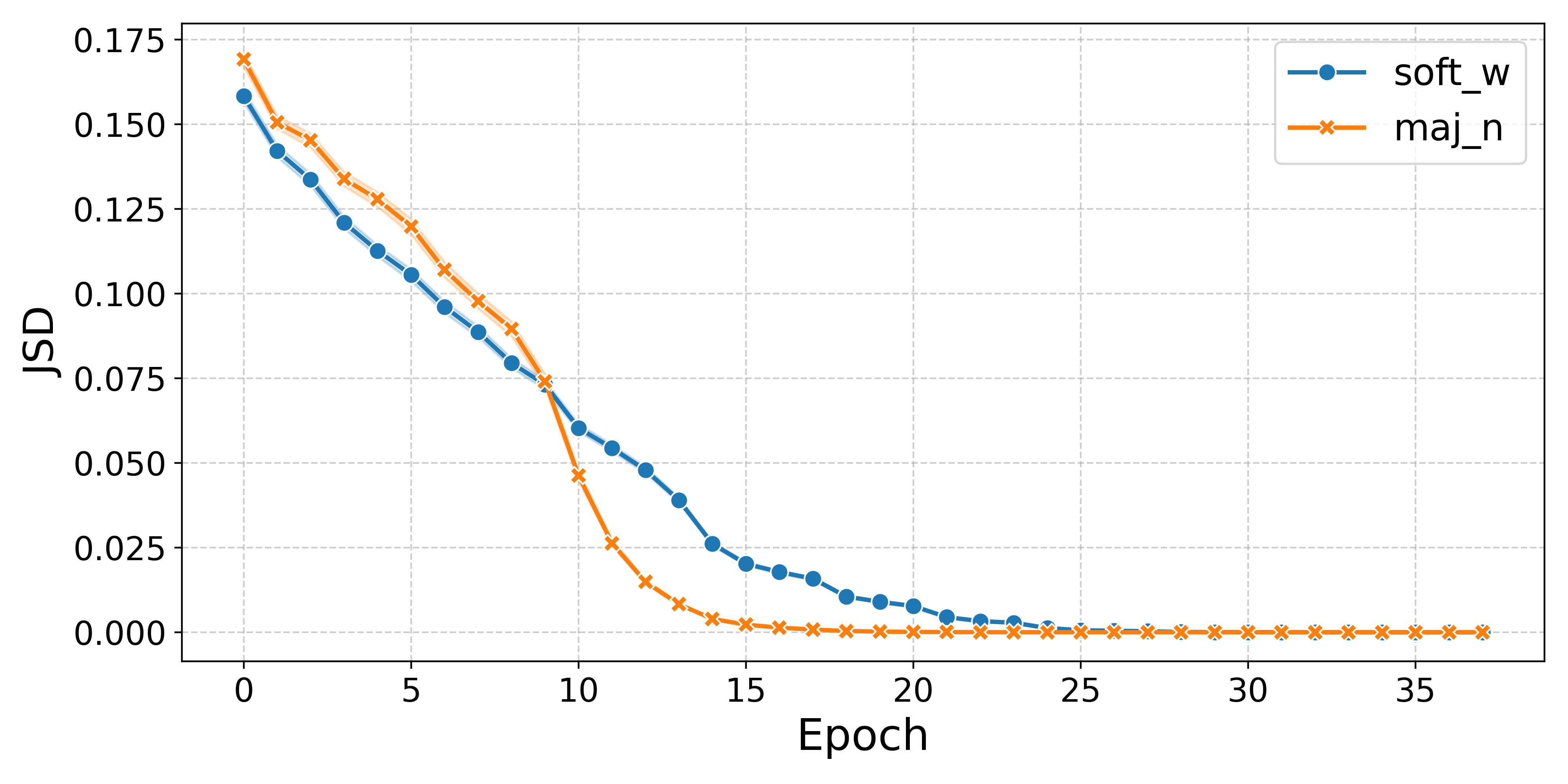}
        \vspace{-1.7em}
        \caption{MNIST - \textit{NoHLV}}
        \label{fig:row1_left}
    \end{subfigure}
    \begin{subfigure}[b]{0.49\textwidth}
        \centering
        \includegraphics[width=\textwidth]{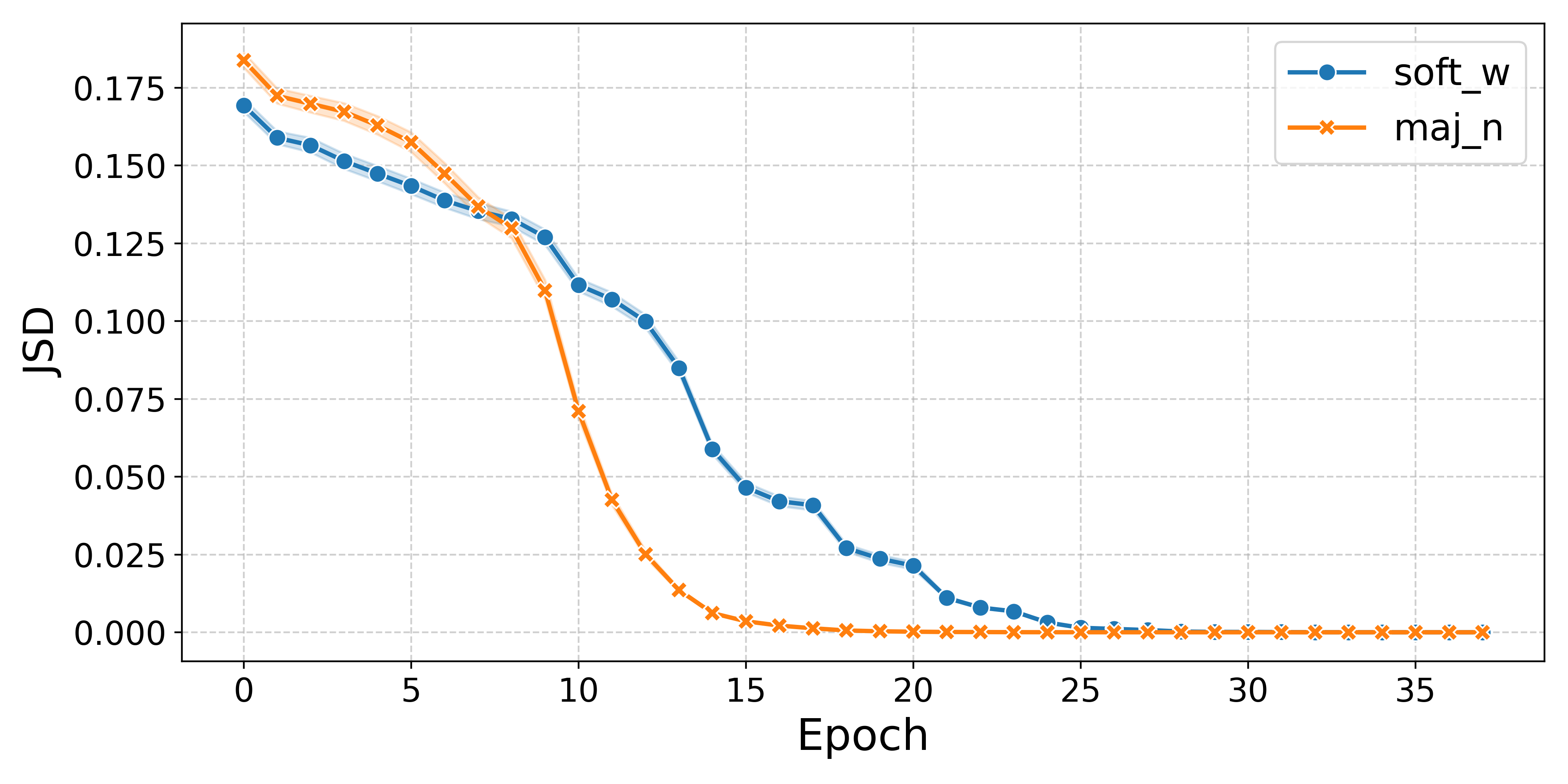}
        \vspace{-1.7em}
        \caption{MNIST - \textit{HLV}}
        \label{fig:row1_right}
    \end{subfigure}
    
    \vspace{.5ex}
    
    \begin{subfigure}[b]{0.49\textwidth}
        \centering
        \includegraphics[width=\textwidth]{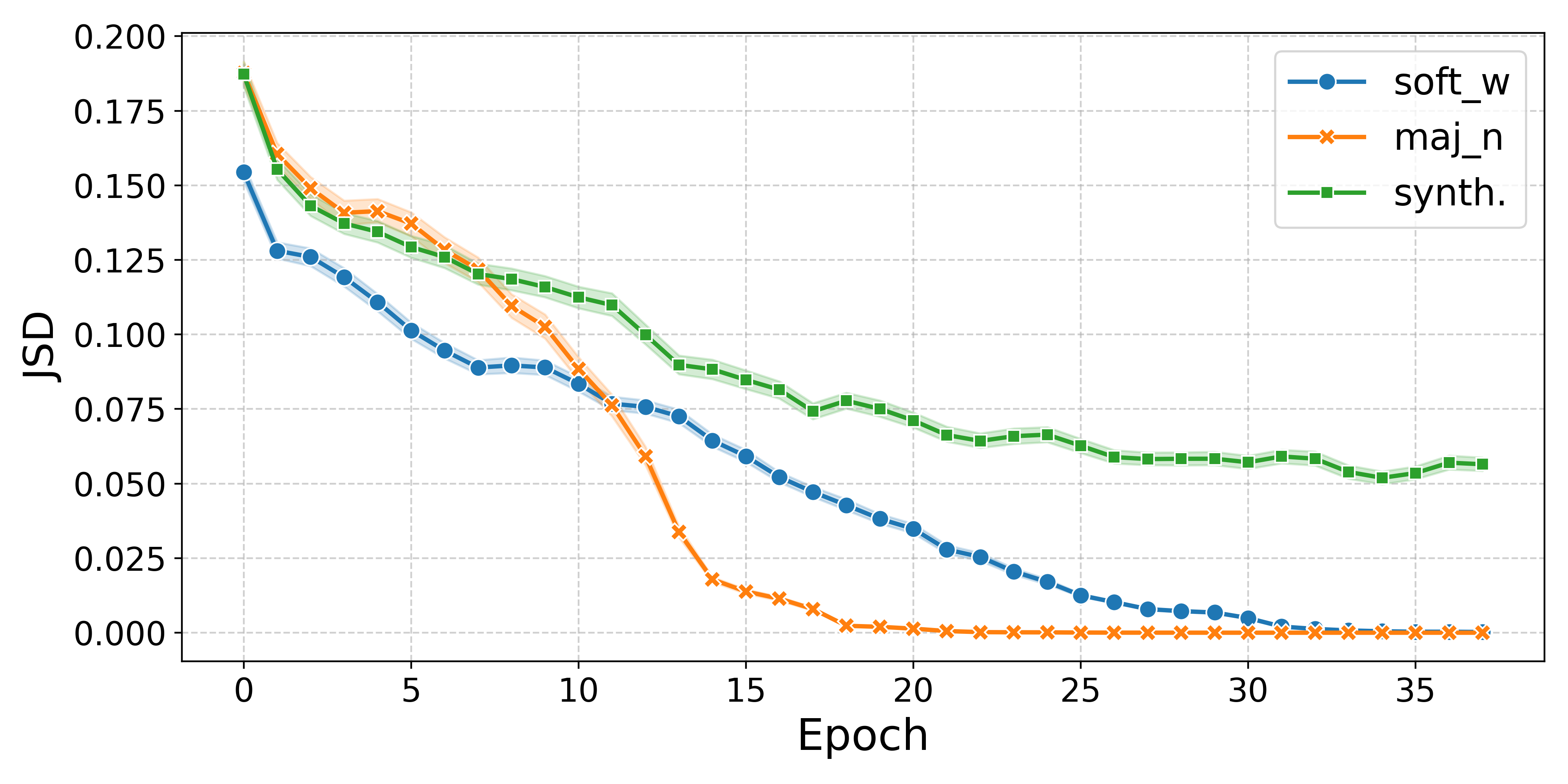}
        \vspace{-1.7em}
        \caption{Mukhoti - \textit{NoHLV}}        
        \label{fig:row3_left}
    \end{subfigure}
    \begin{subfigure}[b]{0.49\textwidth}
        \centering
        \includegraphics[width=\textwidth]{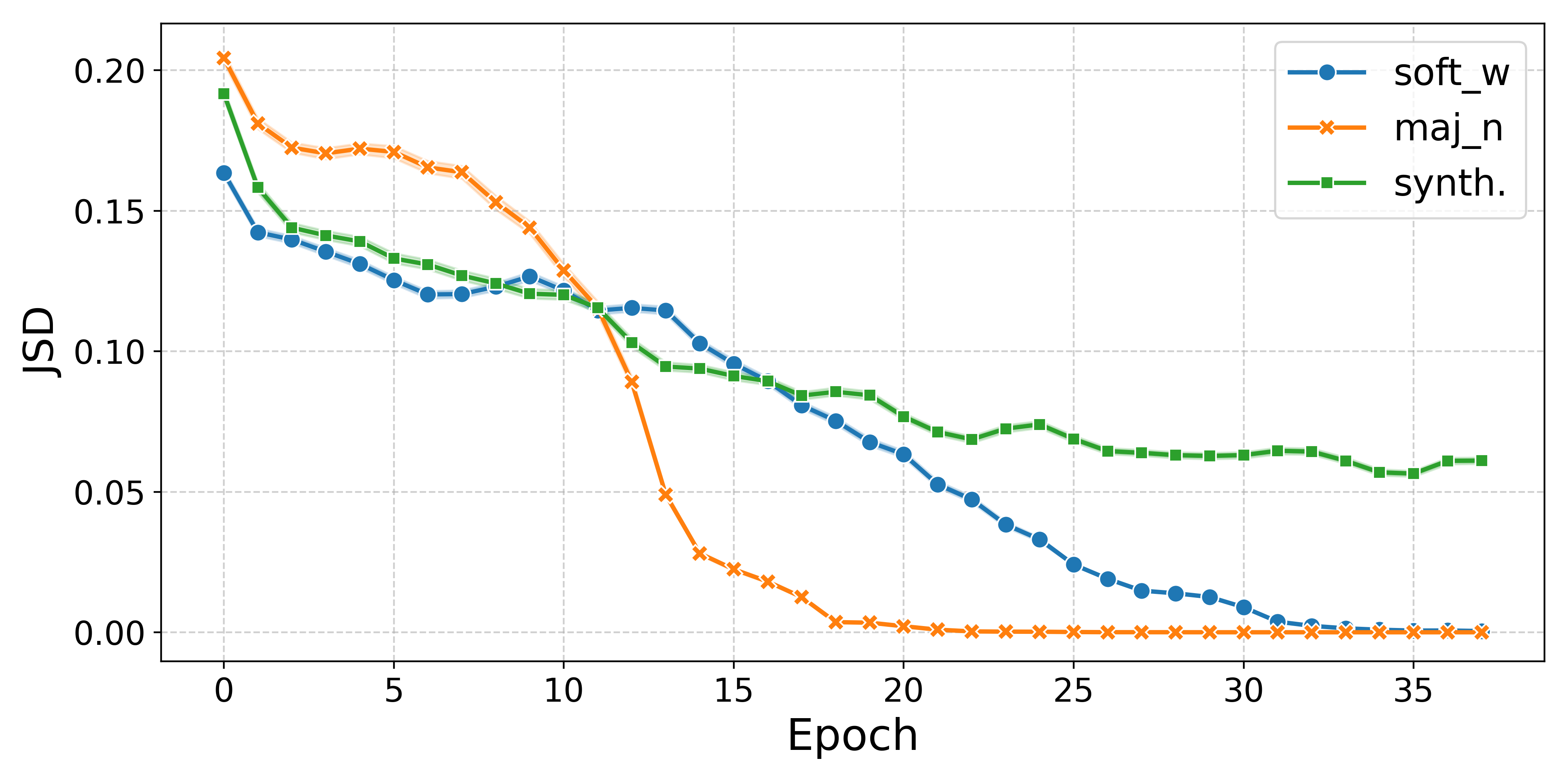}
        \vspace{-1.7em}
        \caption{Mukhoti - \textit{HLV}}        
    \end{subfigure}
    
    \caption{
    The JSD between successive epochs $JSD(P_e || P_{e-1})$ serves as a proxy for how much the model’s \textit{beliefs} are shifting - LeNet over 6 seeds }
    \label{fig:jsd_epochs}
\end{figure}

\begin{figure}[htbp]
    \centering
    \begin{subfigure}[b]{0.49\textwidth}
        \centering
        \includegraphics[width=\textwidth]{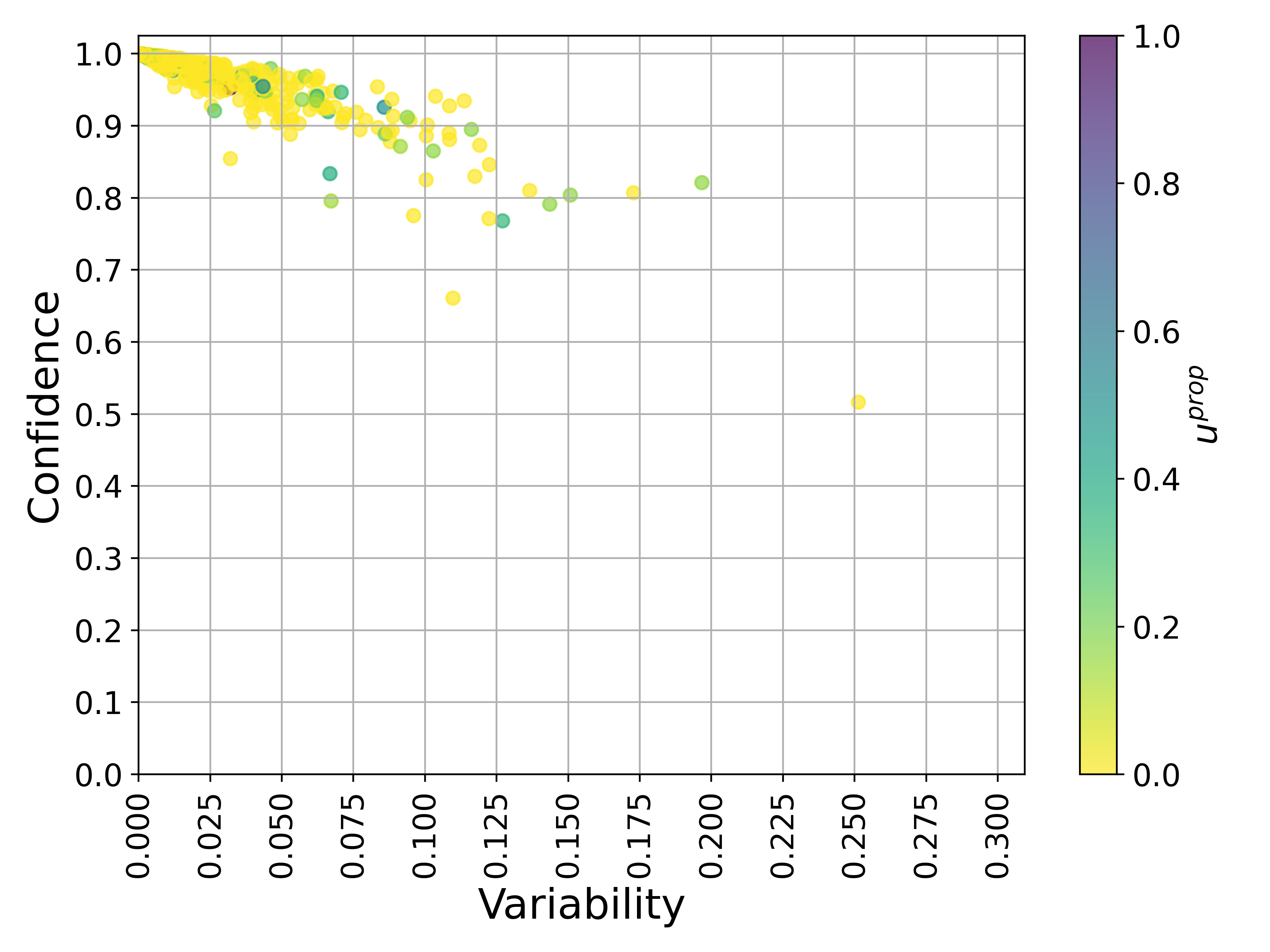}
        \vspace{-1.7em}
        \caption{MNIST - \textit{NoHLV} - $maj._{n}$ }
        \label{fig:row1_right}
    \end{subfigure}
    \begin{subfigure}[b]{0.49\textwidth}
        \centering
        \includegraphics[width=\textwidth]{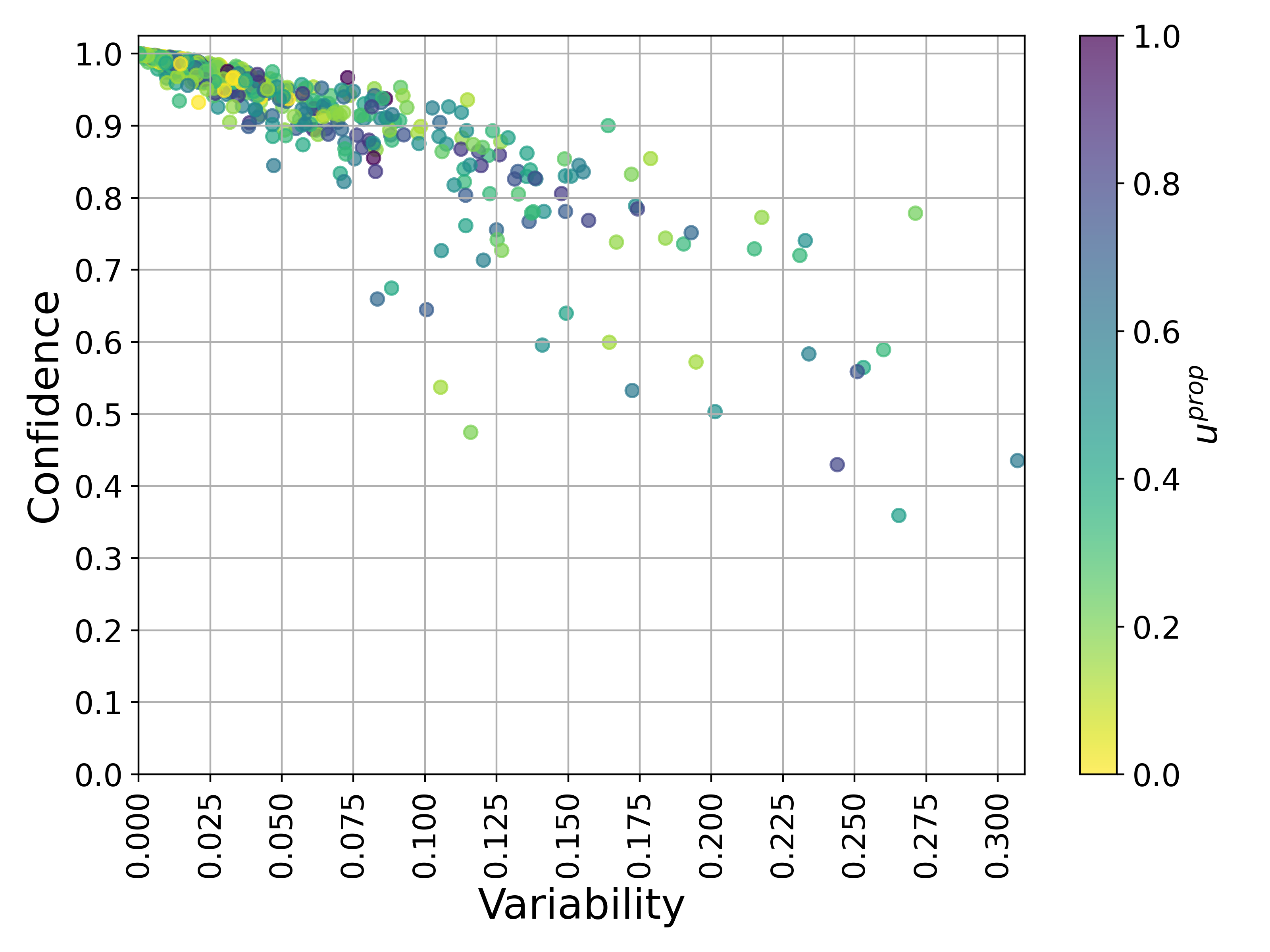}
        \vspace{-1.7em}
        \caption{MNIST - \textit{HLV} - $maj._{n}$}
        \label{fig:row2_right}
    \end{subfigure}

    \vspace{.5ex} 

    \begin{subfigure}[b]{0.49\textwidth}
        \centering
        \includegraphics[width=\textwidth]{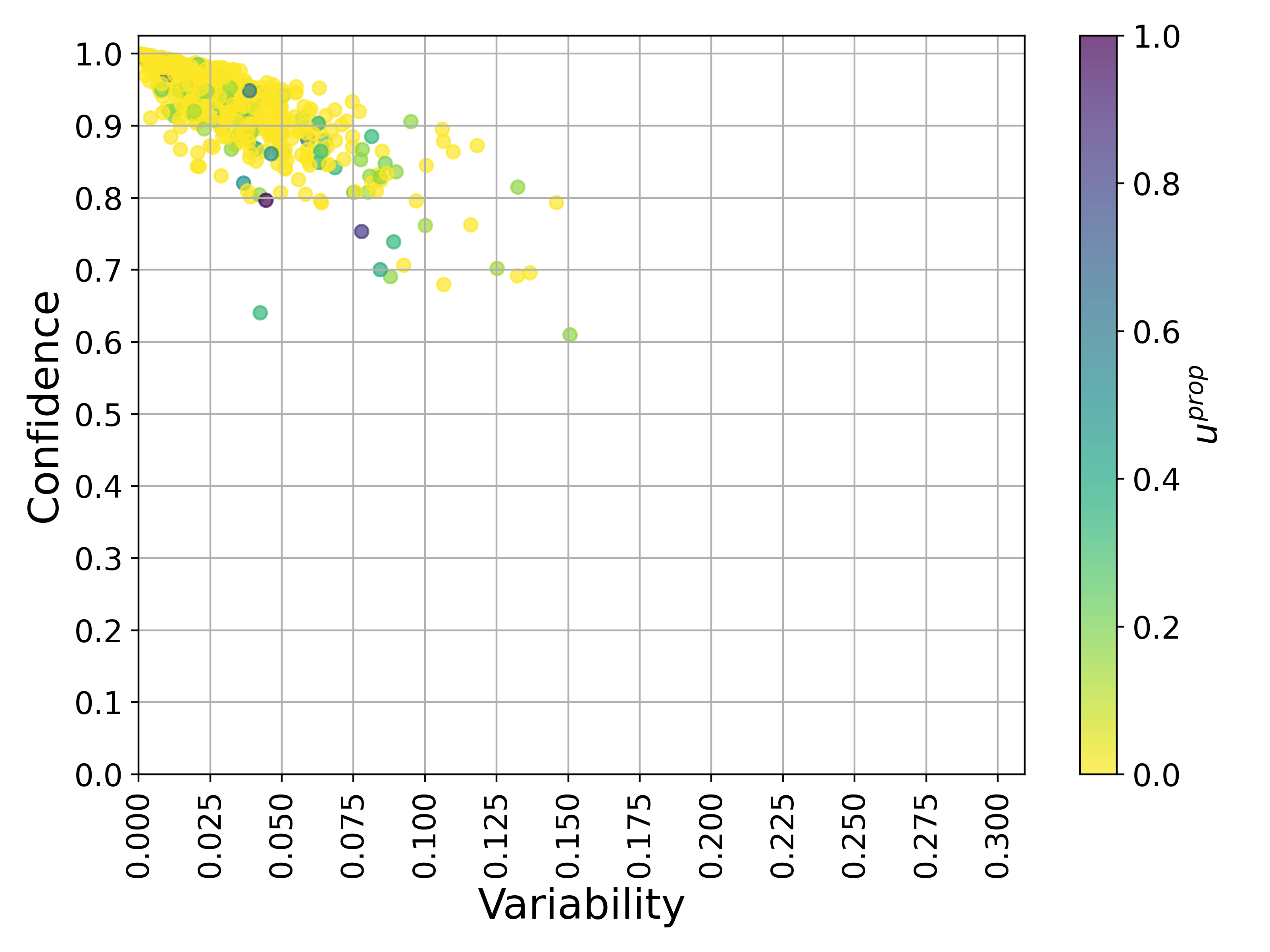}
        \vspace{-1.7em}
        \caption{MNIST - \textit{NoHLV} - $soft_{w}$ }
        \label{fig:row1_left}
    \end{subfigure}
    \begin{subfigure}[b]{0.49\textwidth}
        \centering
        \includegraphics[width=\textwidth]{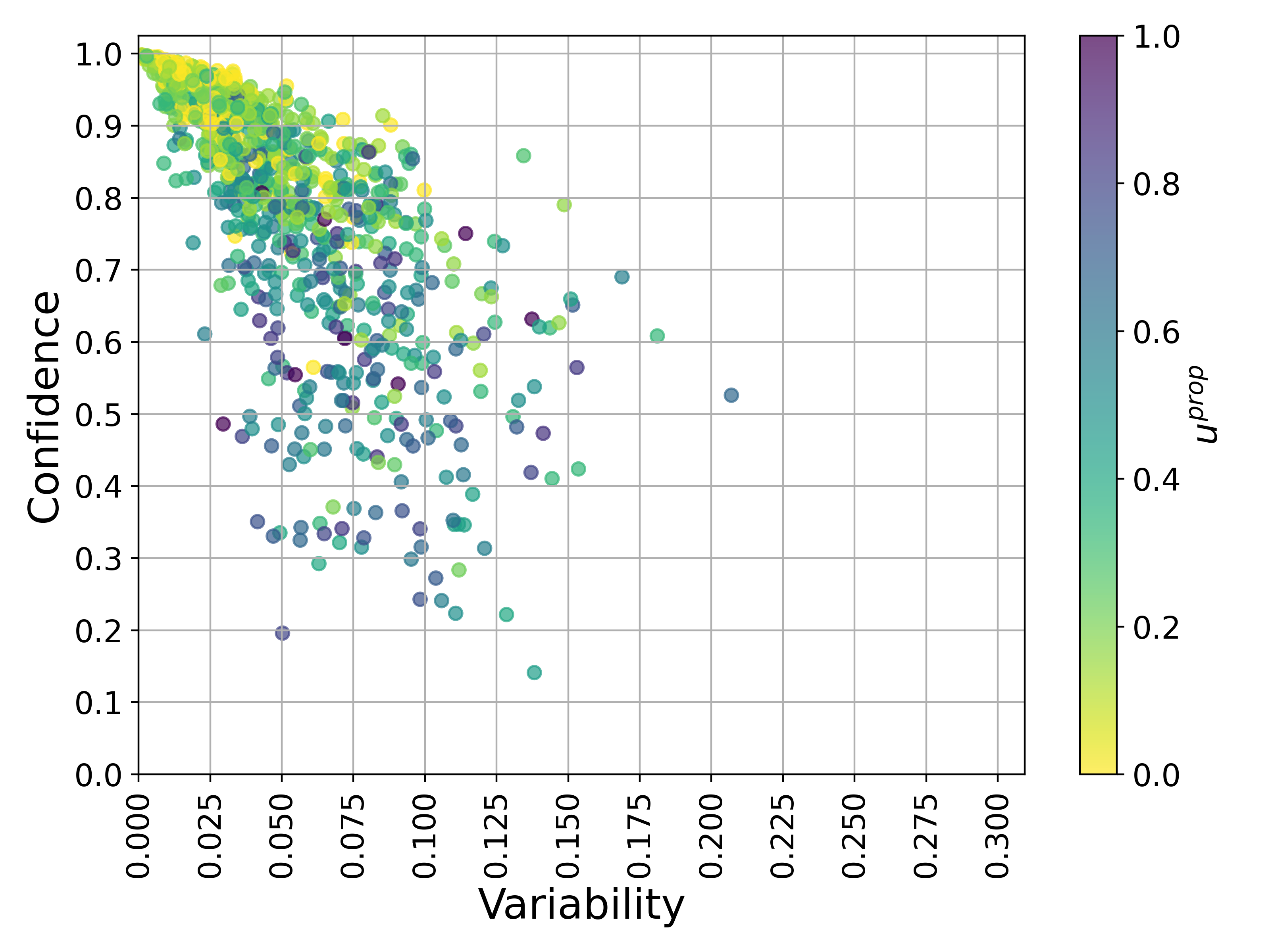}
        \vspace{-1.7em}
        \caption{MNIST - \textit{HLV} - $soft_{w}$}
        \label{fig:row2_left}
    \end{subfigure}

    \caption{
    Late stage training dynamics of LeNet {\small (last 5 epochs)} averaged across 6 random seeds on MNIST data, with $soft_w$ and $maj._n$ contrasted;
    training on soft-labels indicates lower variability in HLV strata compared to training on majority label. The mean confidence values for training under $maj._n$ targets are:
    $\mu_{NoHLV}\approx0.99\pm0.004$ and $\mu_{HLV}\approx0.96\pm0.016$;
    while for the $soft_w$ training regime the confidences in the subsets are on average lower:  $\mu_{NoHLV}\approx0.96\pm0.007$ and $\mu_{HLV}\approx0.79\pm0.009$; same as observed in \cite{peterson_2019_iccv}.}
    \label{fig:mnist_cartography_human_uncert}
\end{figure}

\begin{figure}[htbp]
    \centering
    \begin{subfigure}[b]{0.49\textwidth}
        \centering
        \includegraphics[width=\textwidth]{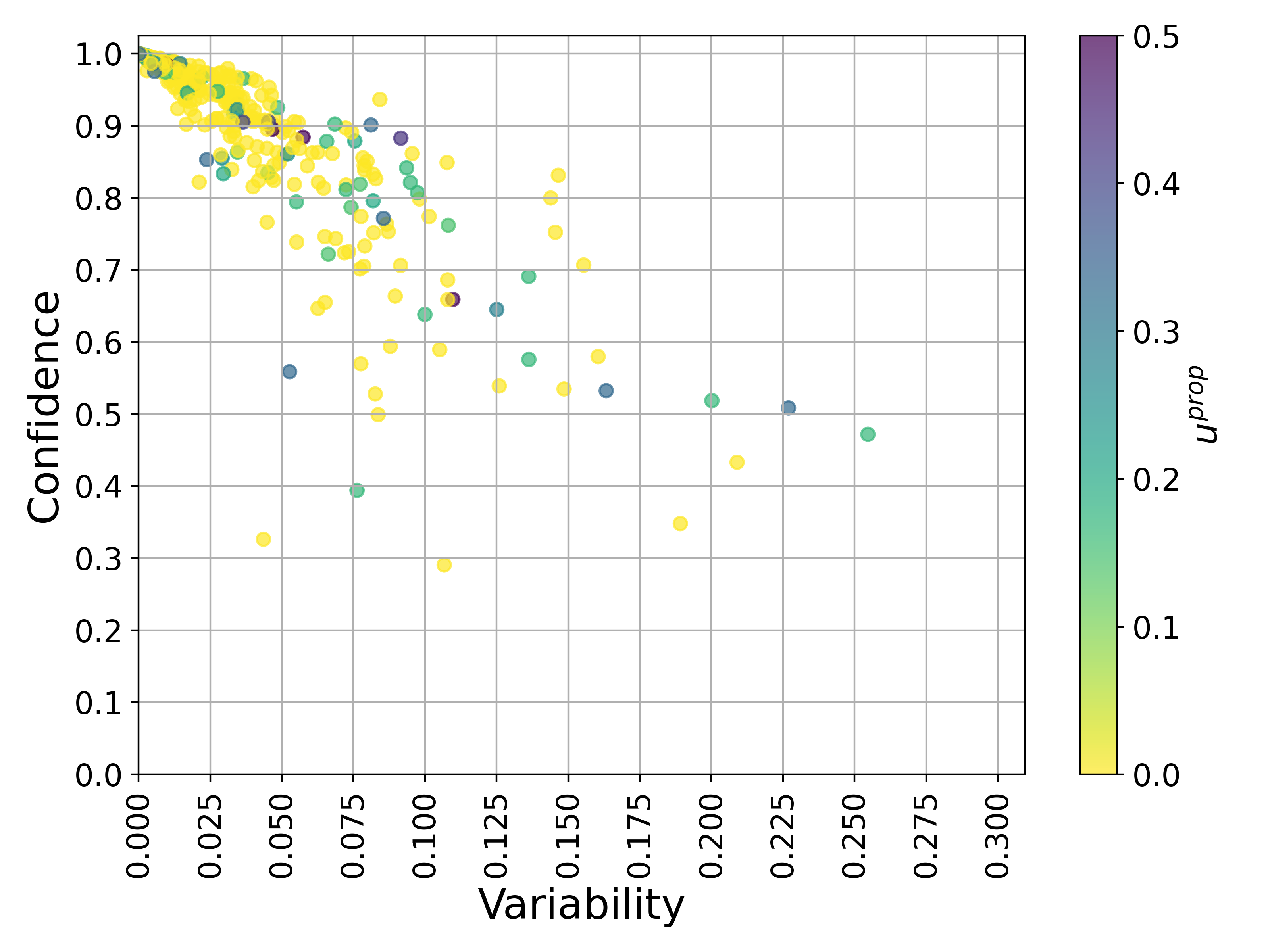}
        \vspace{-1.7em}
        \caption{Mukhoti - \textit{NoHLV} - $maj._{n}$ }
        \label{fig:row1_left}
    \end{subfigure}
    \begin{subfigure}[b]{0.49\textwidth}
        \centering
        \includegraphics[width=\textwidth]{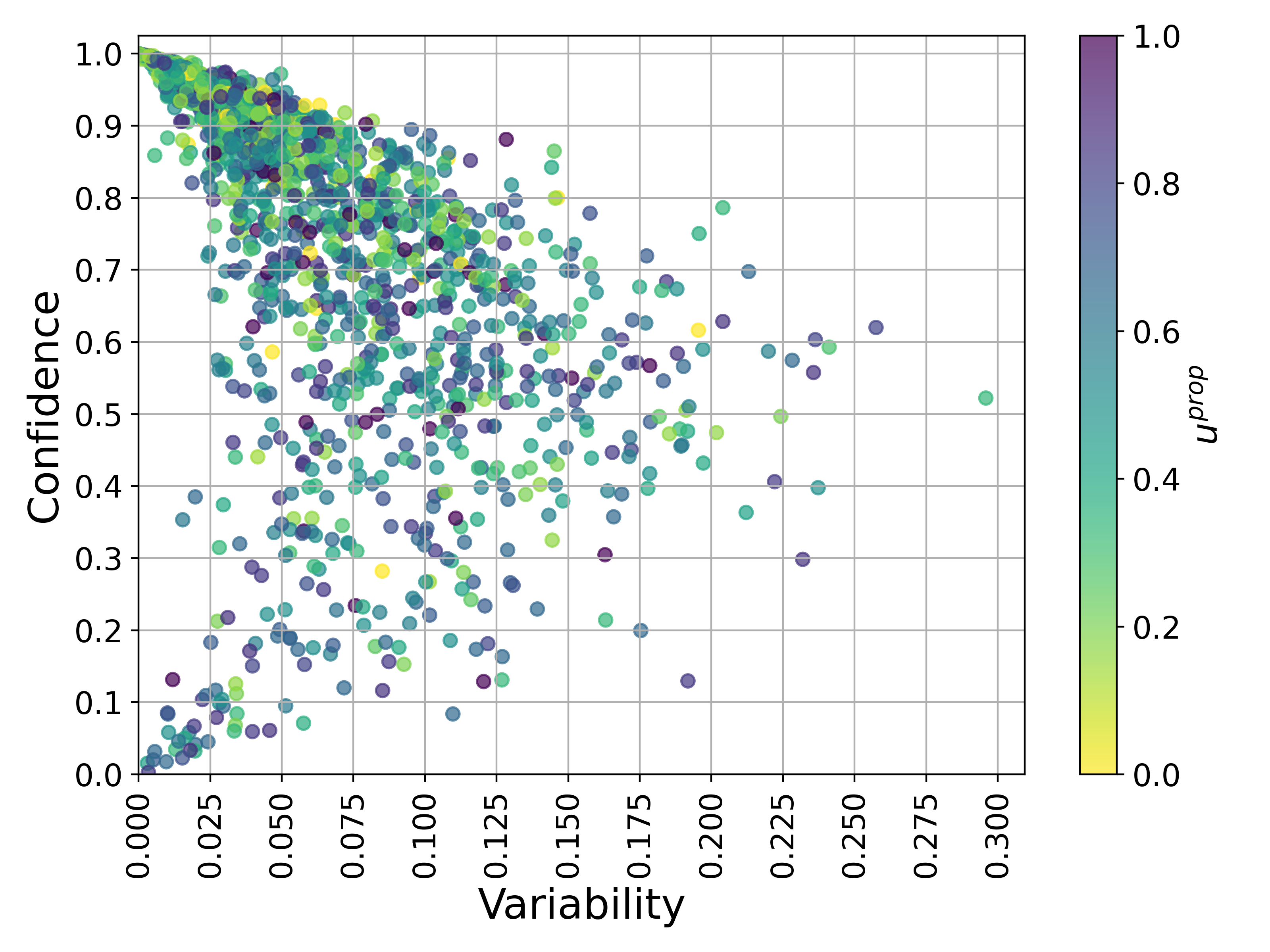}
        \vspace{-1.7em}
        \caption{Mukhoti - \textit{HLV} - $maj._{n}$ }
        \label{fig:row1_right}
    \end{subfigure}

    \vspace{.5ex} 

    \begin{subfigure}[b]{0.49\textwidth}
        \centering
        \includegraphics[width=\textwidth]{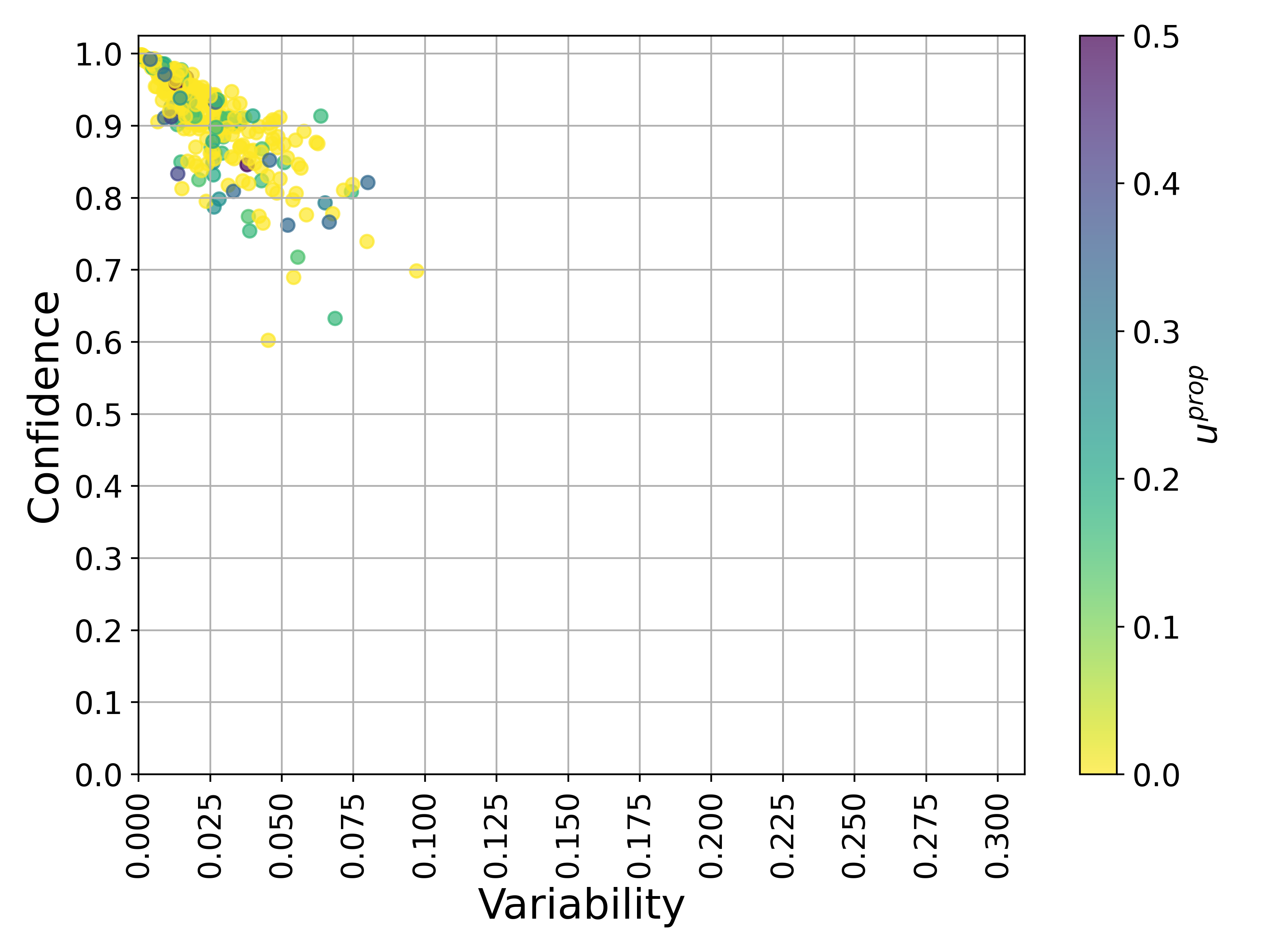}
        \vspace{-1.7em}
        \caption{Mukhoti - \textit{NoHLV} - $soft_{w}$ }
        \label{fig:row2_left}
    \end{subfigure}
    \begin{subfigure}[b]{0.49\textwidth}
        \centering
        \includegraphics[width=\textwidth]
        {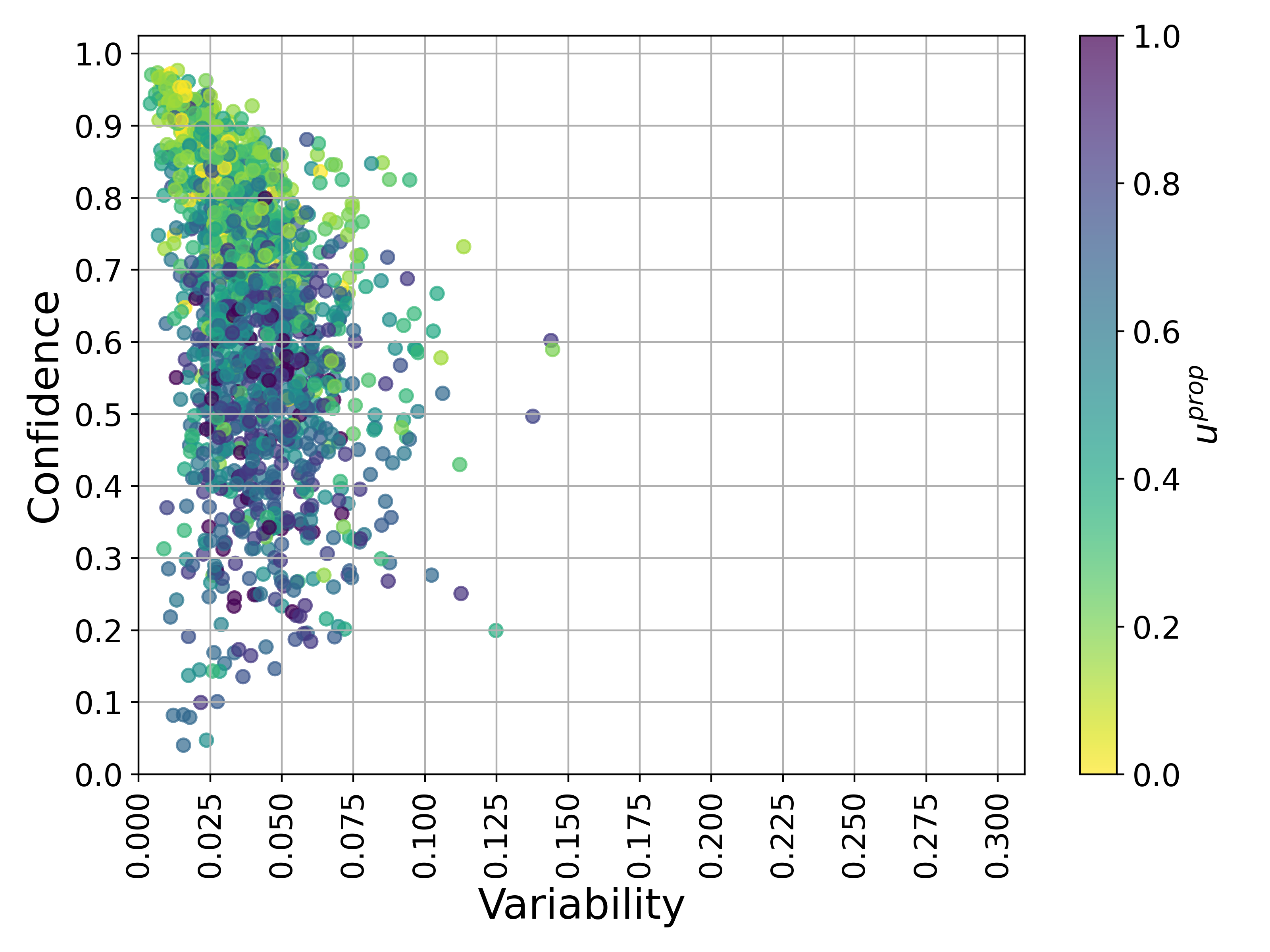}
        \vspace{-1.7em}
        \caption{Mukhoti - \textit{HLV} - $soft_{w}$ }
        \label{fig:row2_right}
    \end{subfigure}

    \vspace{.5ex}

    \begin{subfigure}[b]{0.49\textwidth}
        \centering
        \includegraphics[width=\textwidth]{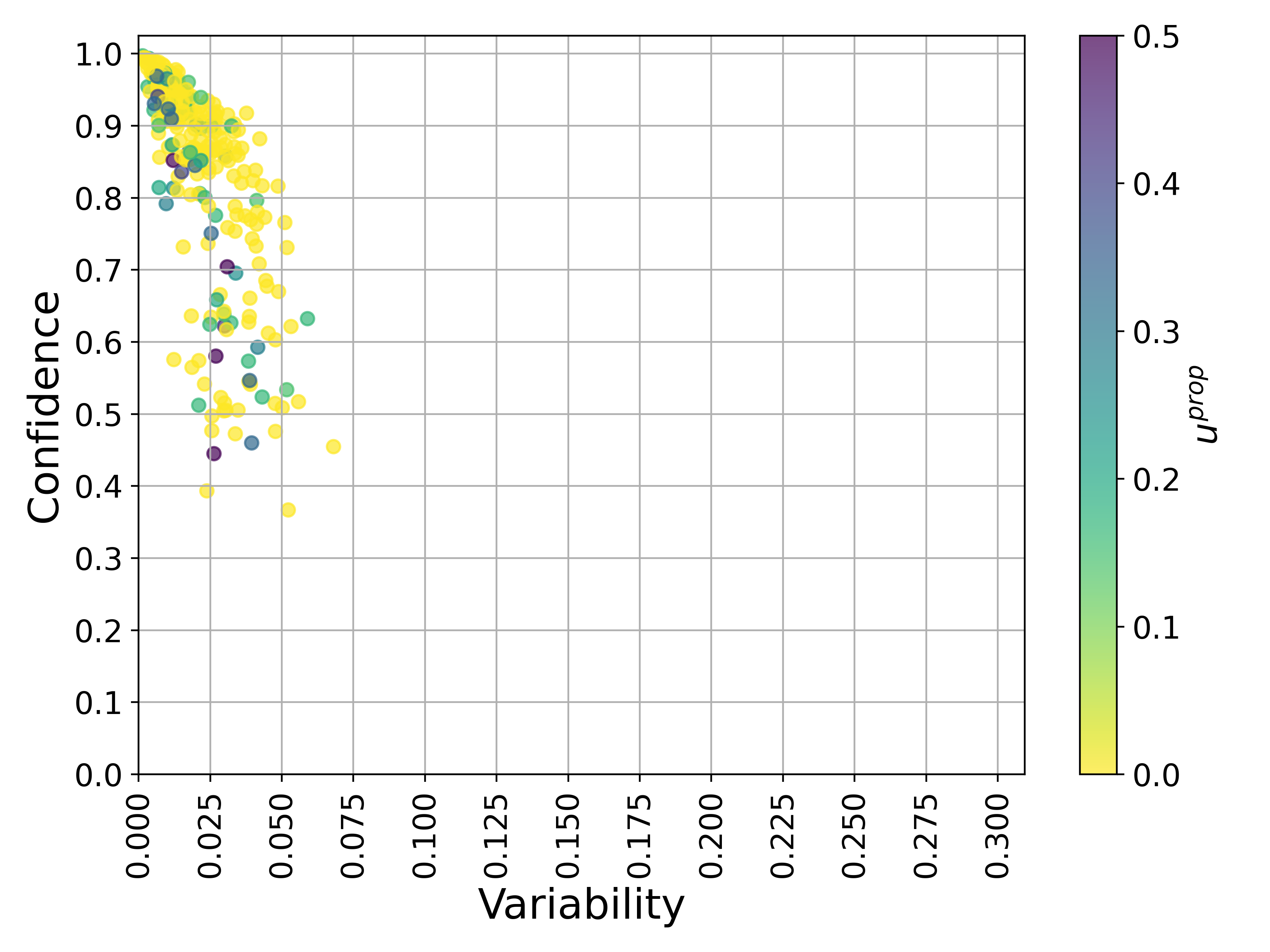}
        \vspace{-1.7em}
        \caption{Mukhoti - \textit{NoHLV} - $synth.$ }
        \label{fig:row3_left}
    \end{subfigure}
    \begin{subfigure}[b]{0.49\textwidth}
        \centering
        \includegraphics[width=\textwidth]{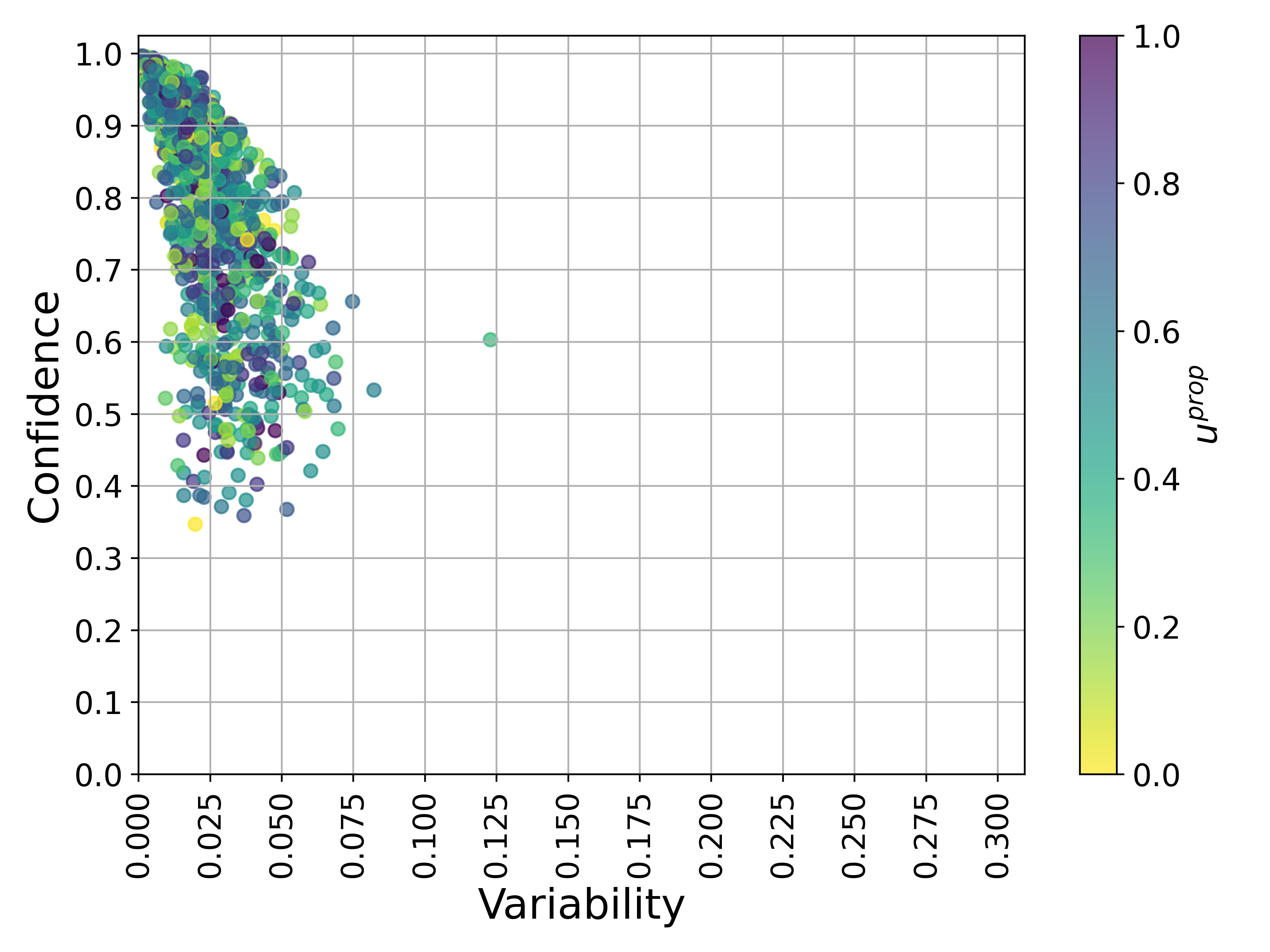}
        \vspace{-1.7em}
        \caption{Mukhoti - \textit{HLV} - $synth.$ }
        \label{fig:row3_right}
    \end{subfigure}

    \caption{
    Late stage training dynamics of DeeperFFN {\small (last 5 epochs)} averaged across 6 random seeds on Muhkoti data, with $soft_w$ and $maj._n$ contrasted;
    training on soft-labels indicates lower variability in HLV strata compared to training on majority label. 
    }
    \label{fig:mukhoti_data_maps_human_uncertainty_deeper}
\end{figure}

\begin{figure}[htbp]
    \centering
    \begin{subfigure}[b]{0.49\textwidth}
        \centering
        \includegraphics[width=\textwidth]{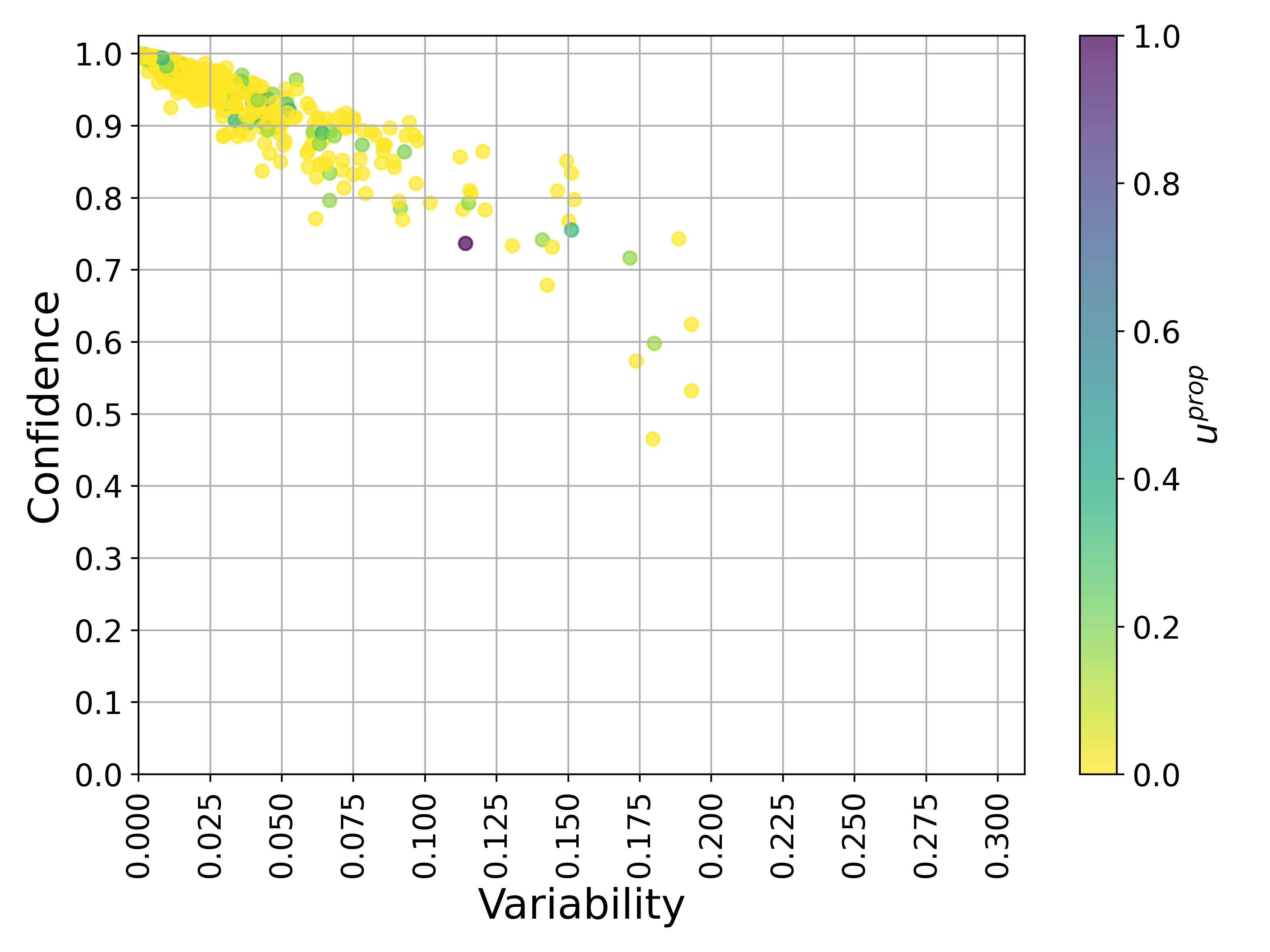}
        \vspace{-1.7em}
        \caption{MNIST - \textit{NoHLV} - $maj._{n}$ }
        \label{fig:row1_right}
    \end{subfigure}
    \begin{subfigure}[b]{0.49\textwidth}
        \centering
        \includegraphics[width=\textwidth]{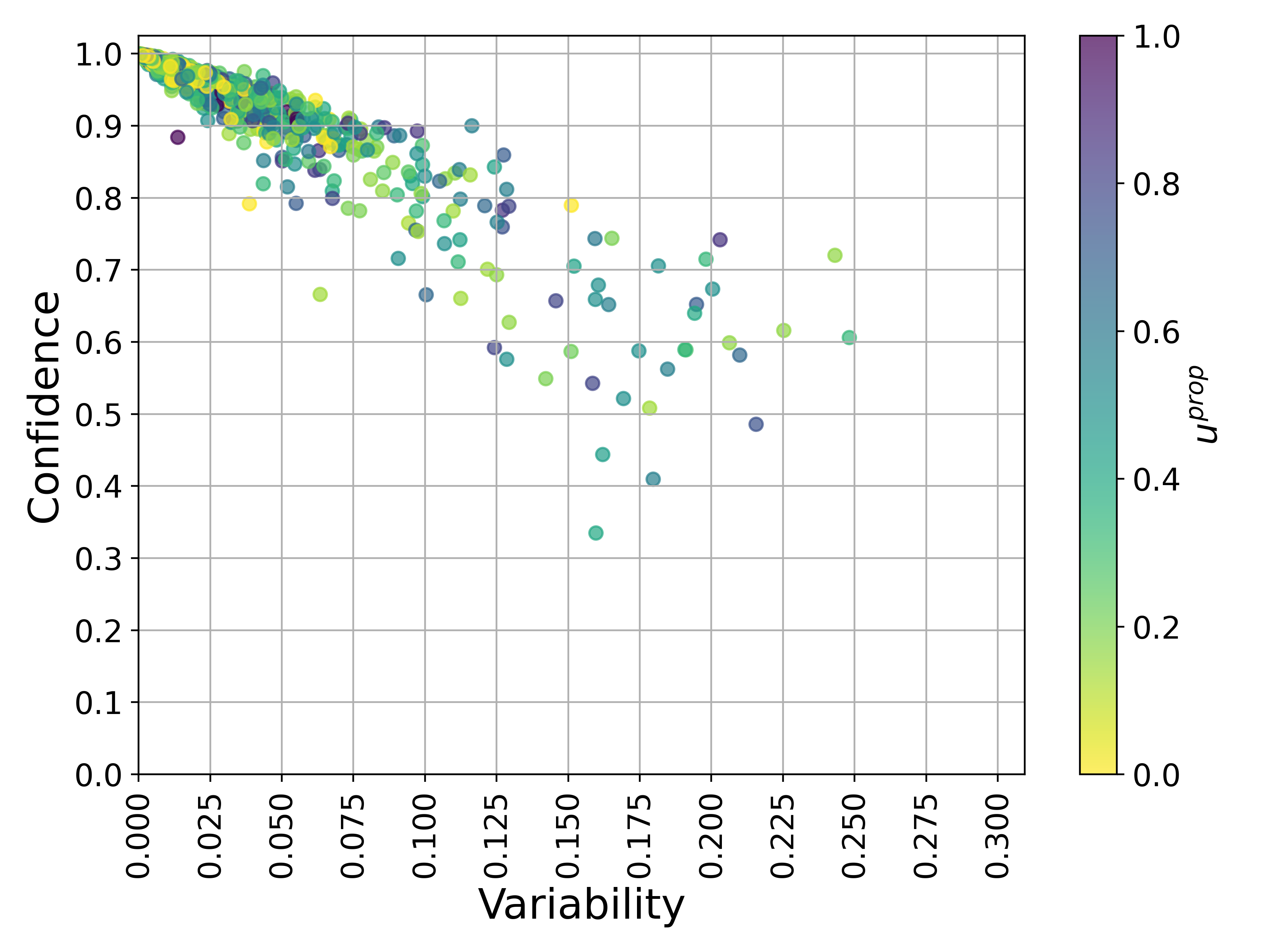}
        \vspace{-1.7em}
        \caption{MNIST - \textit{HLV} - $maj._{n}$}
        \label{fig:row2_right}
    \end{subfigure}

    \vspace{.5ex} 

    \begin{subfigure}[b]{0.49\textwidth}
        \centering
        \includegraphics[width=\textwidth]{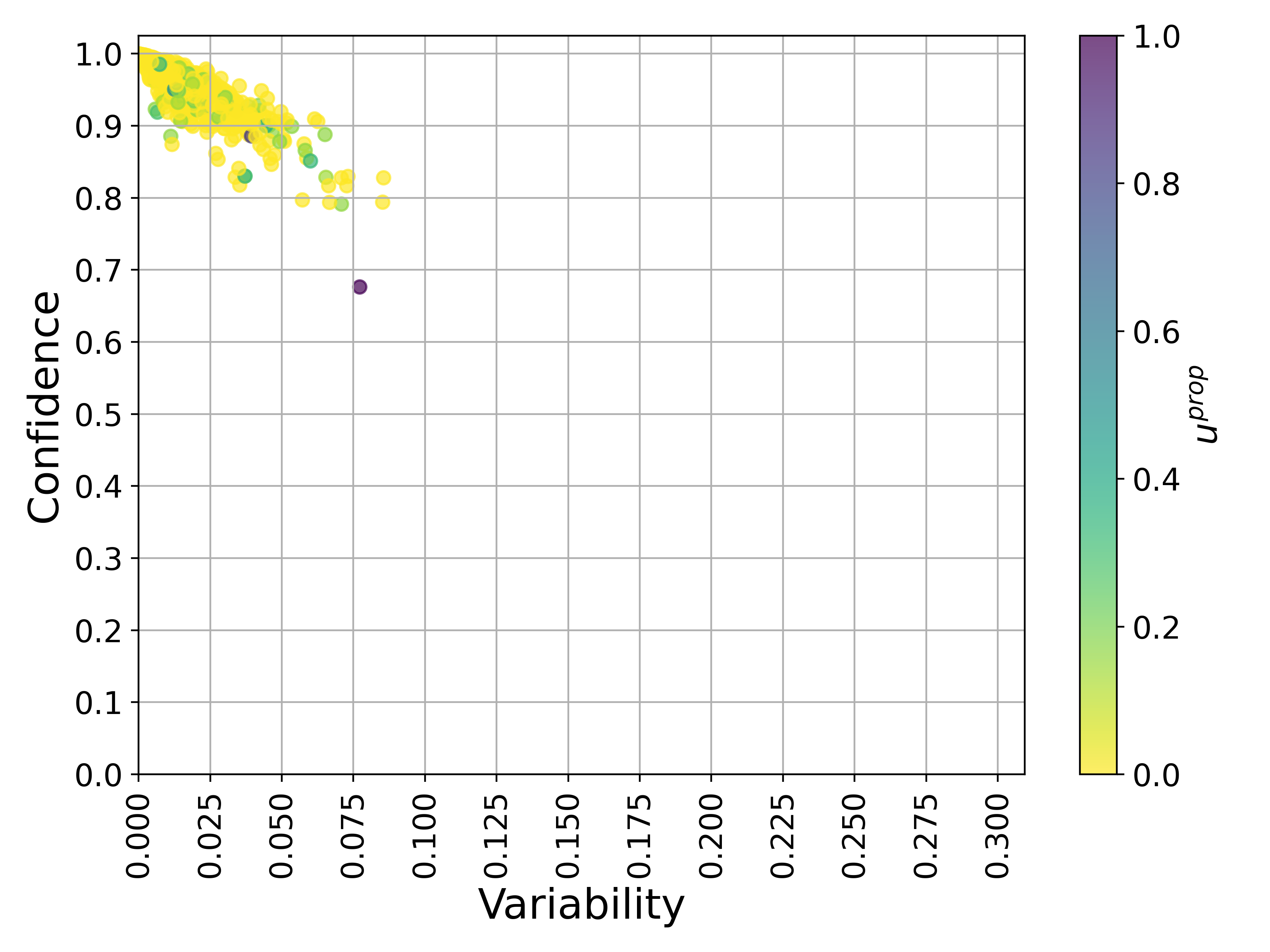}
        \vspace{-1.7em}
        \caption{MNIST - \textit{NoHLV} - $soft_{w}$ }
        \label{fig:row1_left}
    \end{subfigure}
    \begin{subfigure}[b]{0.49\textwidth}
        \centering
        \includegraphics[width=\textwidth]{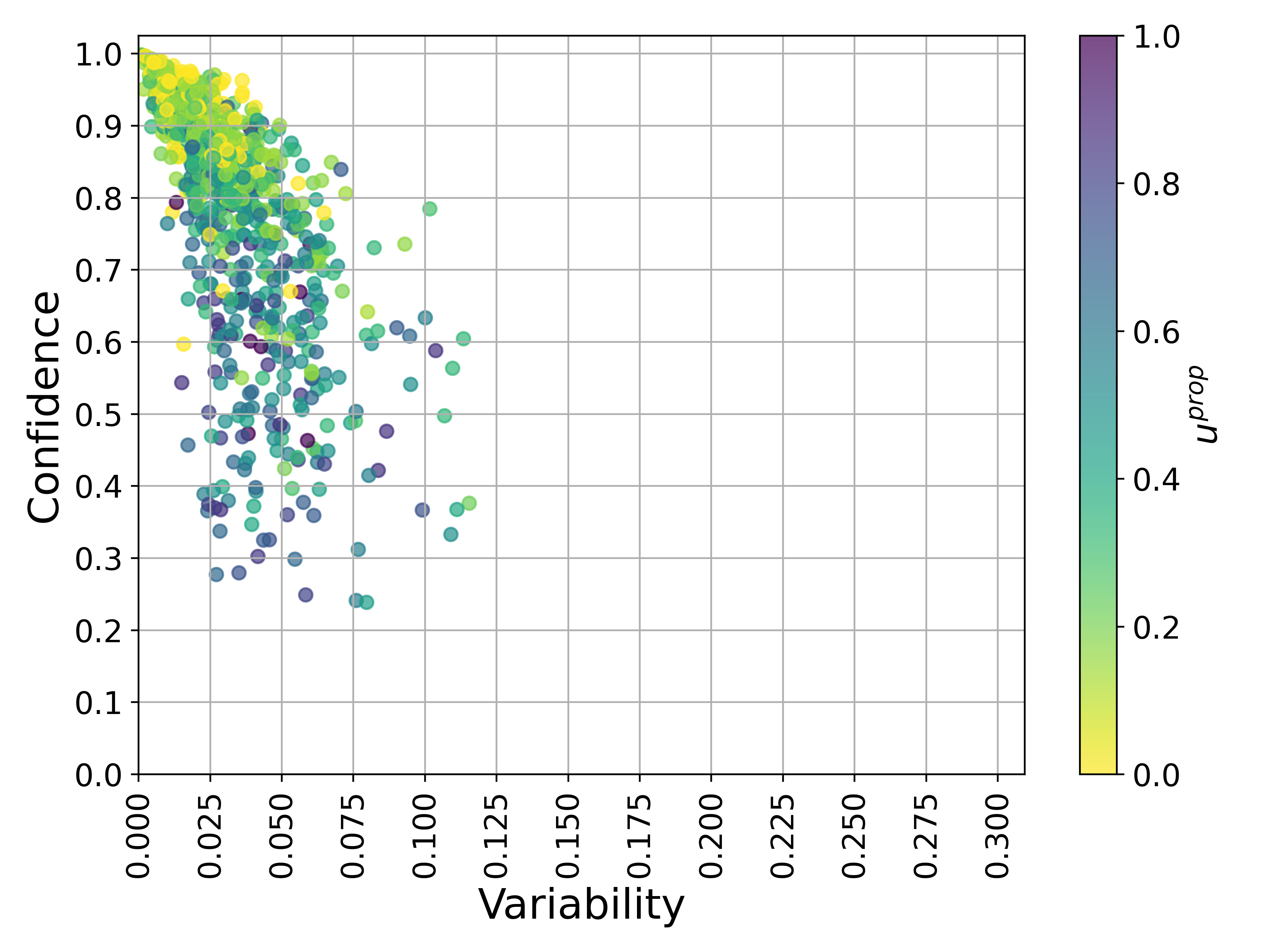}
        \vspace{-1.7em}
        \caption{MNIST - \textit{HLV} - $soft_{w}$}
        \label{fig:row2_left}
    \end{subfigure}

    \caption{
    Late stage training dynamics of DeeperFFN {\small (last 5 epochs)} averaged across 6 random seeds on MNIST data, with $soft_w$ and $maj._n$ contrasted;
    training on soft-labels indicates lower variability in HLV strata compared to training on majority label.}
    \label{fig:mnist_cartography_human_uncert_deeper}
\end{figure}

\begin{figure}[htbp]
    \centering
    \begin{subfigure}[b]{0.49\textwidth}
        \centering
        \includegraphics[width=\textwidth]{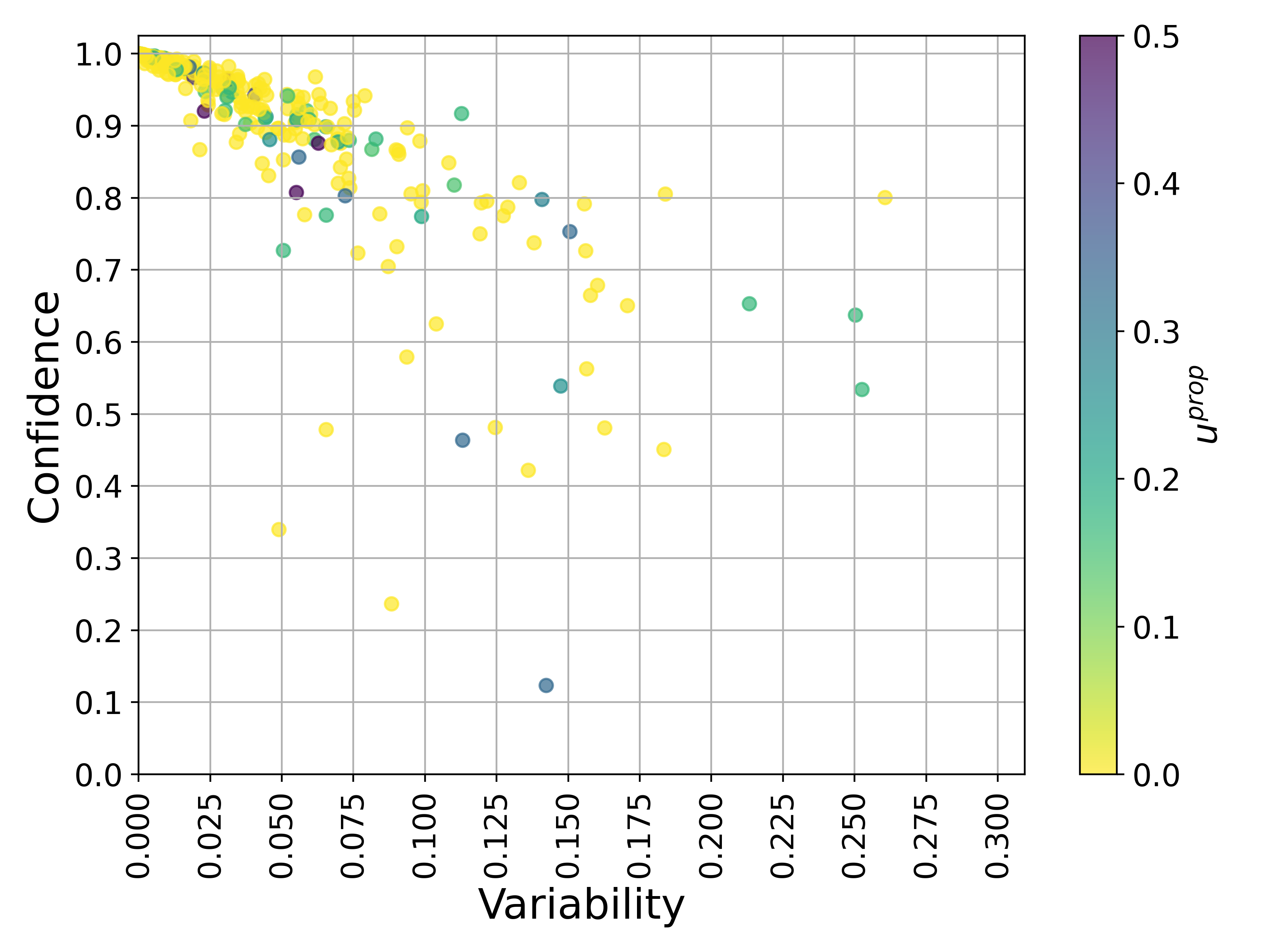}
        \vspace{-1.7em}
        \caption{Mukhoti - \textit{NoHLV} - $maj._{n}$ }
        \label{fig:row1_left}
    \end{subfigure}
    \begin{subfigure}[b]{0.49\textwidth}
        \centering
        \includegraphics[width=\textwidth]{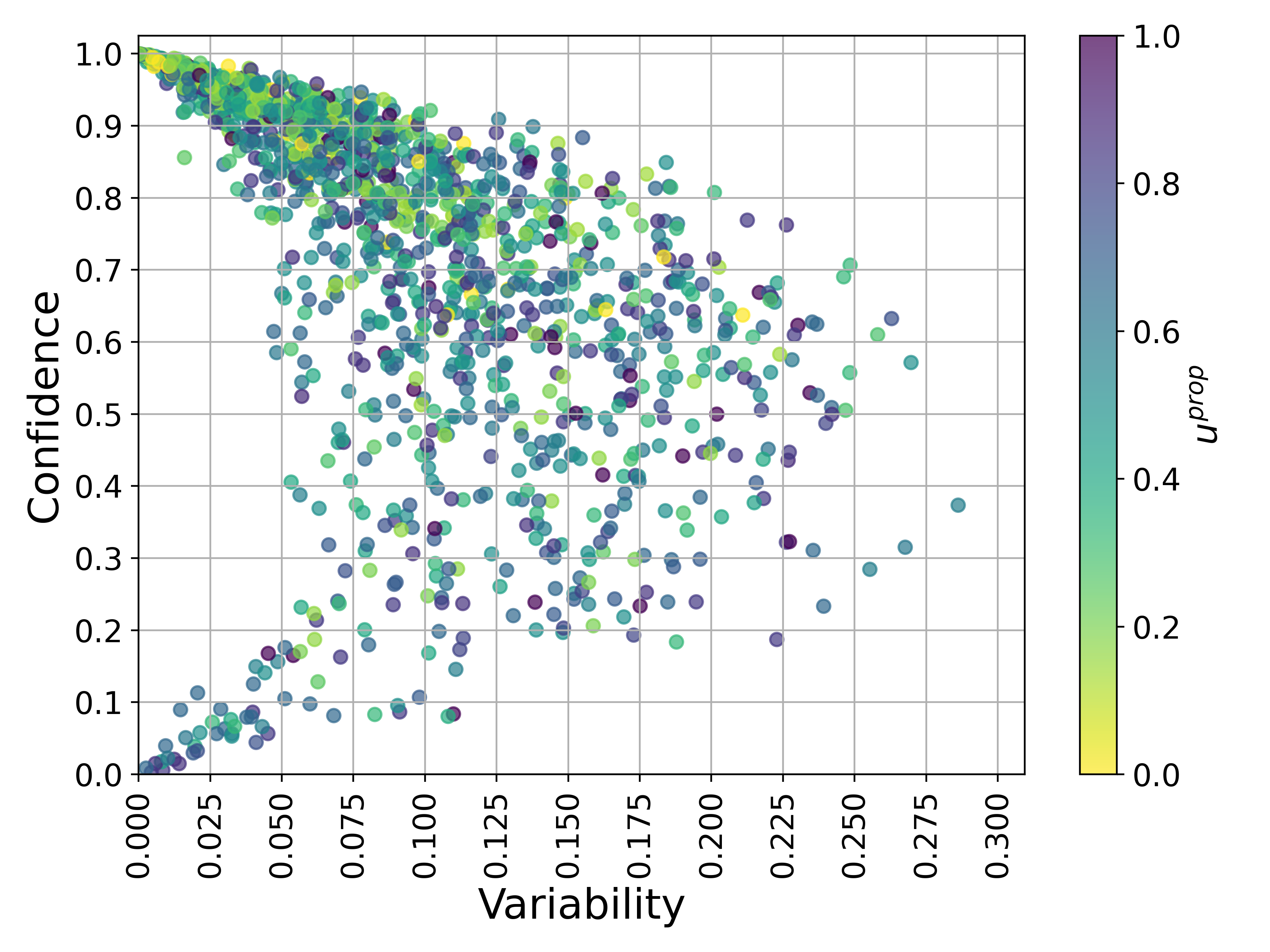}
        \vspace{-1.7em}
        \caption{Mukhoti - \textit{HLV} - $maj._{n}$ }
        \label{fig:row1_right}
    \end{subfigure}

    \vspace{.5ex} 

    \begin{subfigure}[b]{0.49\textwidth}
        \centering
        \includegraphics[width=\textwidth]{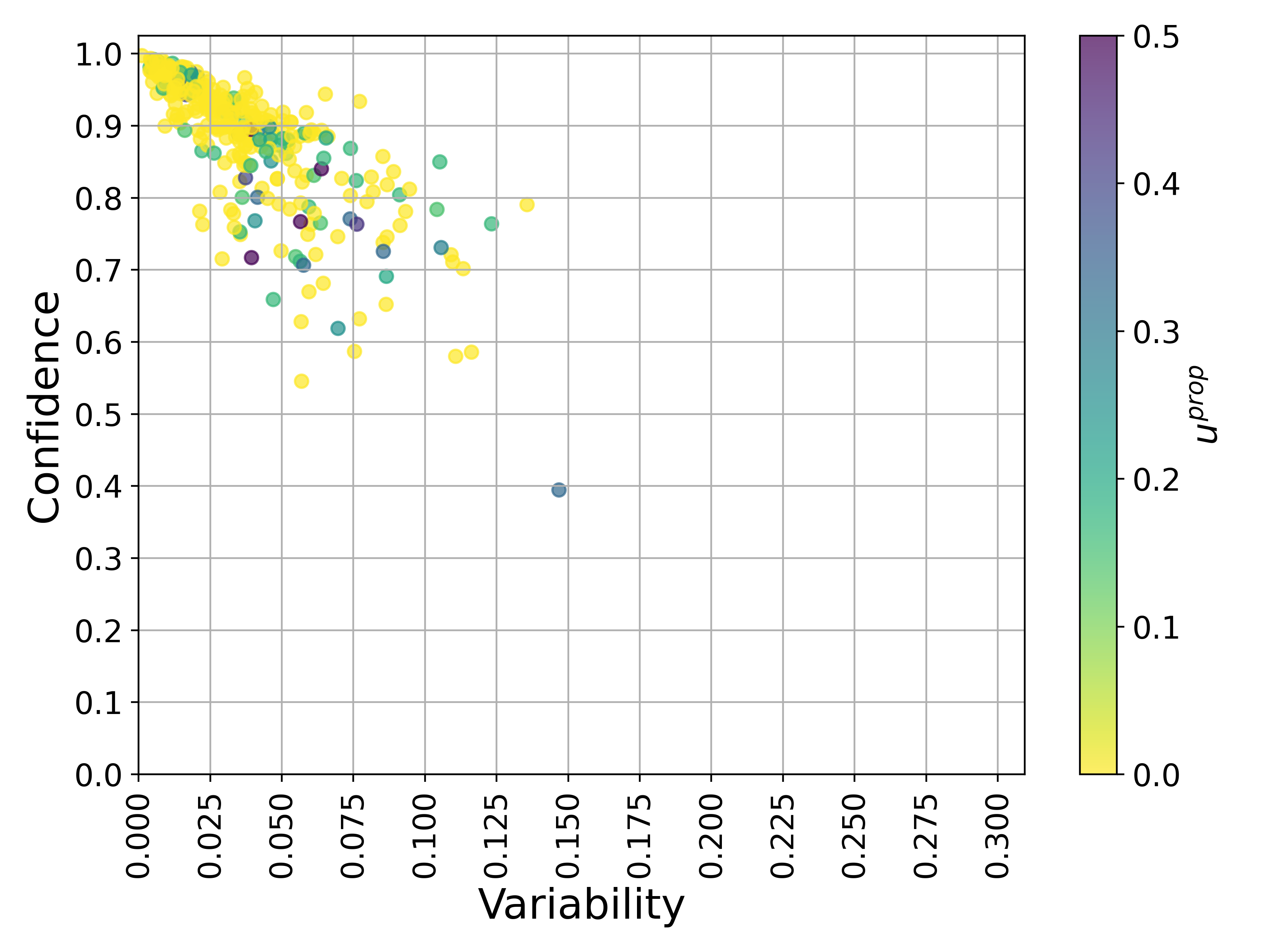}
        \vspace{-1.7em}
        \caption{Mukhoti - \textit{NoHLV} - $soft_{w}$ }
        \label{fig:row2_left}
    \end{subfigure}
    \begin{subfigure}[b]{0.49\textwidth}
        \centering
        \includegraphics[width=\textwidth]
        {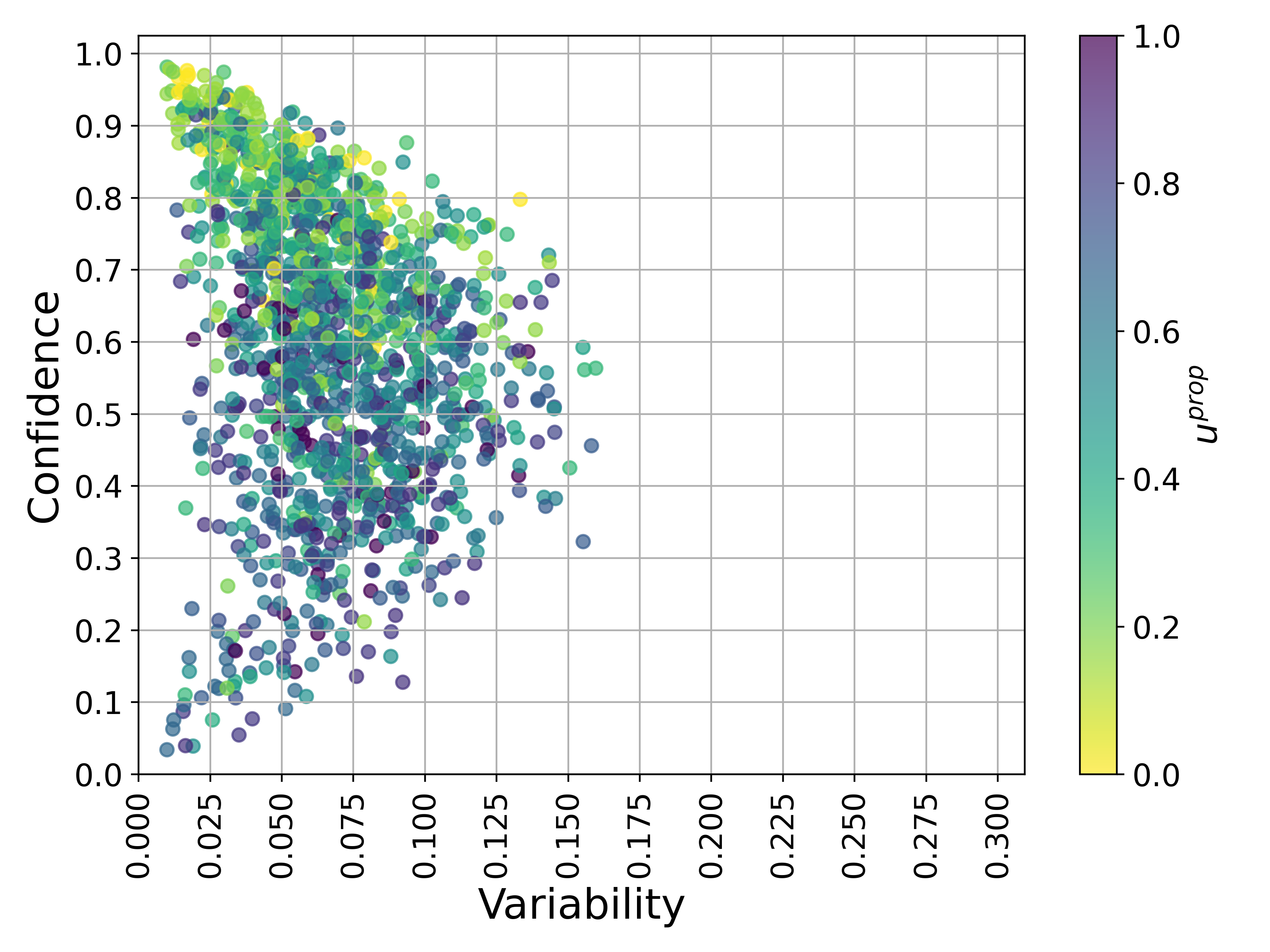}
        \vspace{-1.7em}
        \caption{Mukhoti - \textit{HLV} - $soft_{w}$ }
        \label{fig:row2_right}
    \end{subfigure}

    \vspace{.5ex}

    \begin{subfigure}[b]{0.49\textwidth}
        \centering
        \includegraphics[width=\textwidth]{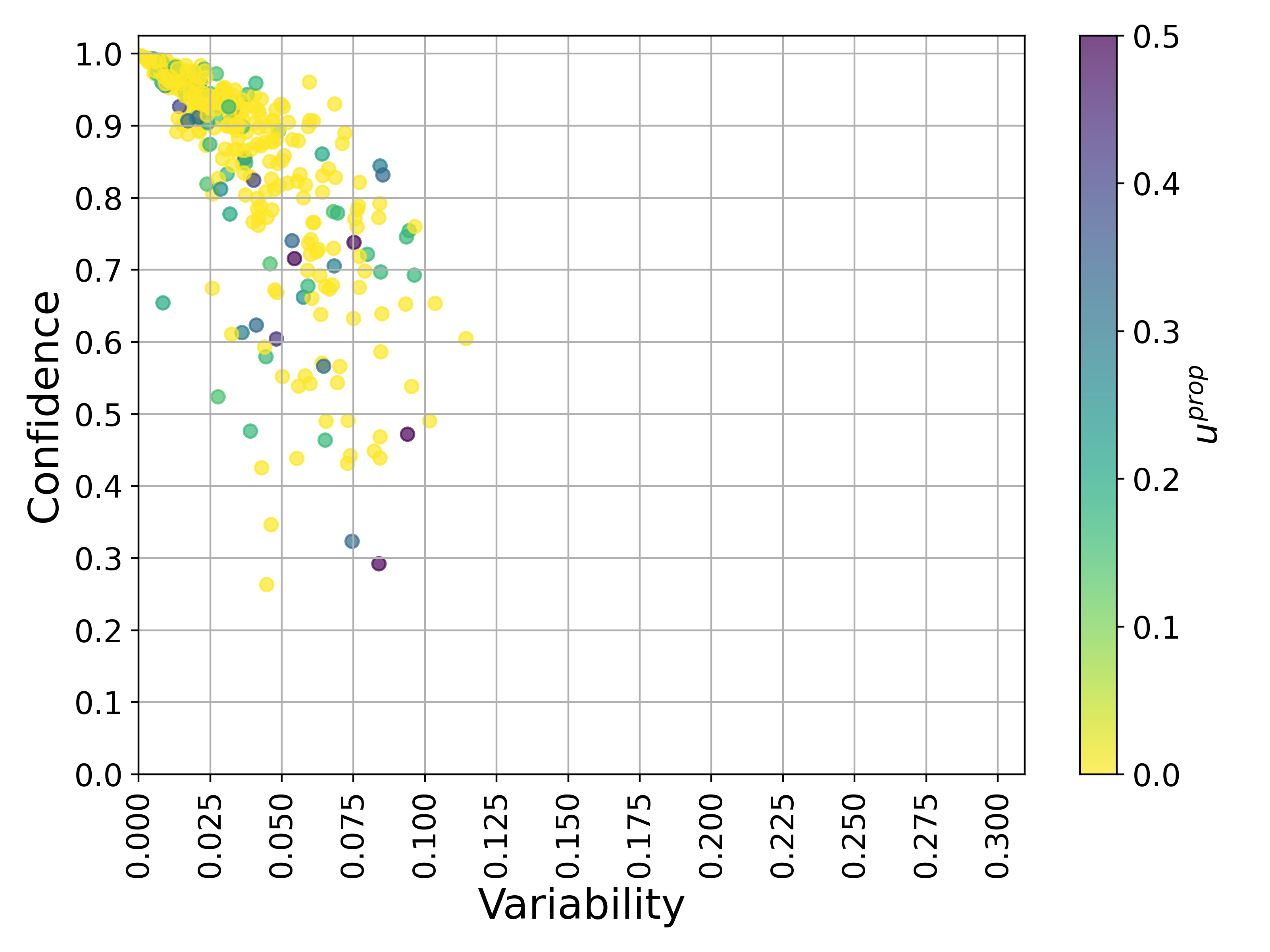}
        \vspace{-1.7em}
        \caption{Mukhoti - \textit{NoHLV} - $synth.$ }
        \label{fig:row3_left}
    \end{subfigure}
    \begin{subfigure}[b]{0.49\textwidth}
        \centering
        \includegraphics[width=\textwidth]{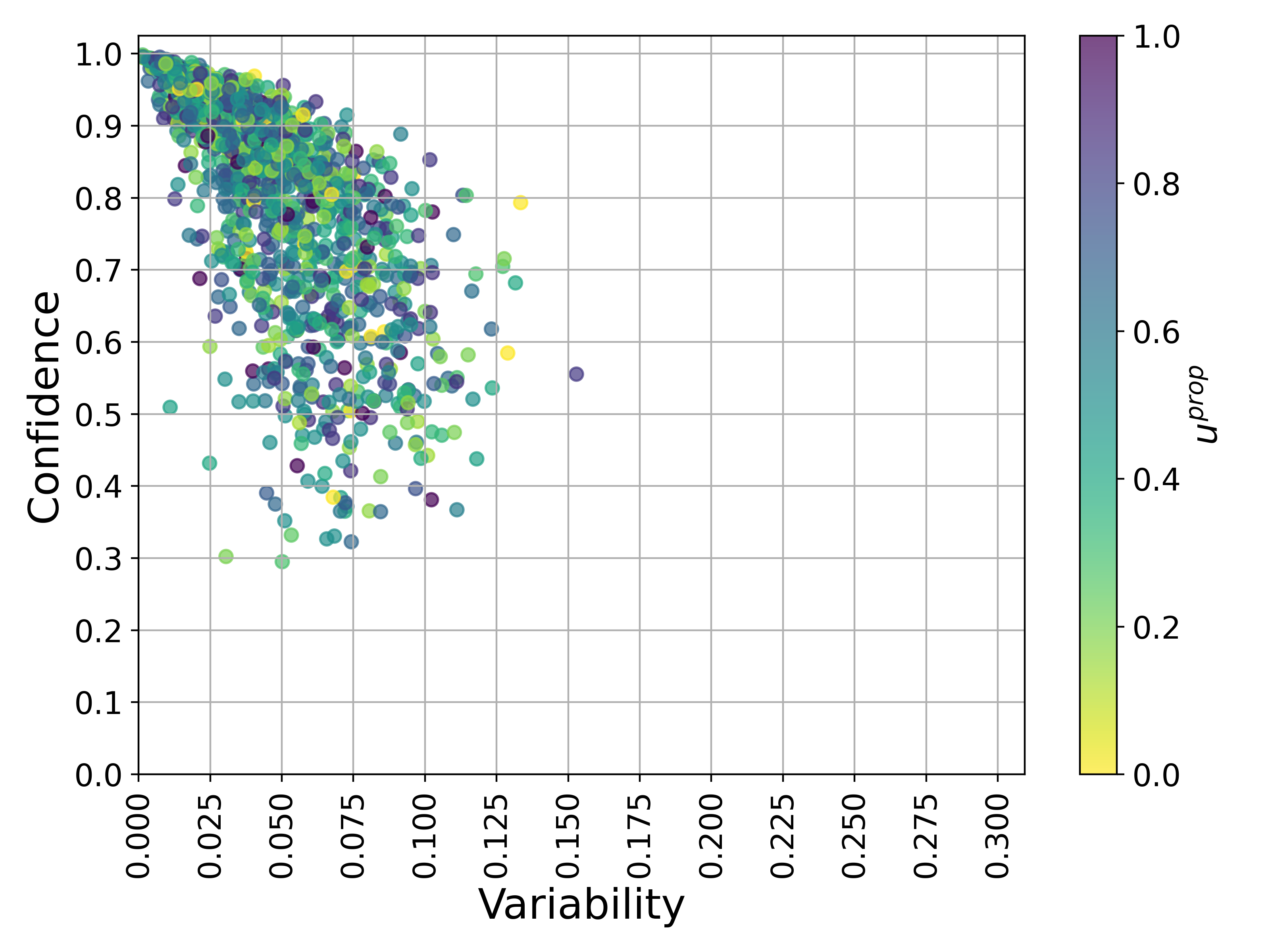}
        \vspace{-1.7em}
        \caption{Mukhoti - \textit{HLV} - $synth.$ }
        \label{fig:row3_right}
    \end{subfigure}

    \caption{
    Late stage training dynamics of SimpleFFN {\small (last 5 epochs)} averaged across 6 random seeds on Muhkoti data, with $soft_w$ and $maj._n$ contrasted;
    training on soft-labels indicates lower variability in HLV strata compared to training on majority label. 
    }
    \label{fig:mukhoti_data_maps_human_uncertainty_simple}
\end{figure}

\begin{figure}[htbp]
    \centering
    \begin{subfigure}[b]{0.49\textwidth}
        \centering
        \includegraphics[width=\textwidth]{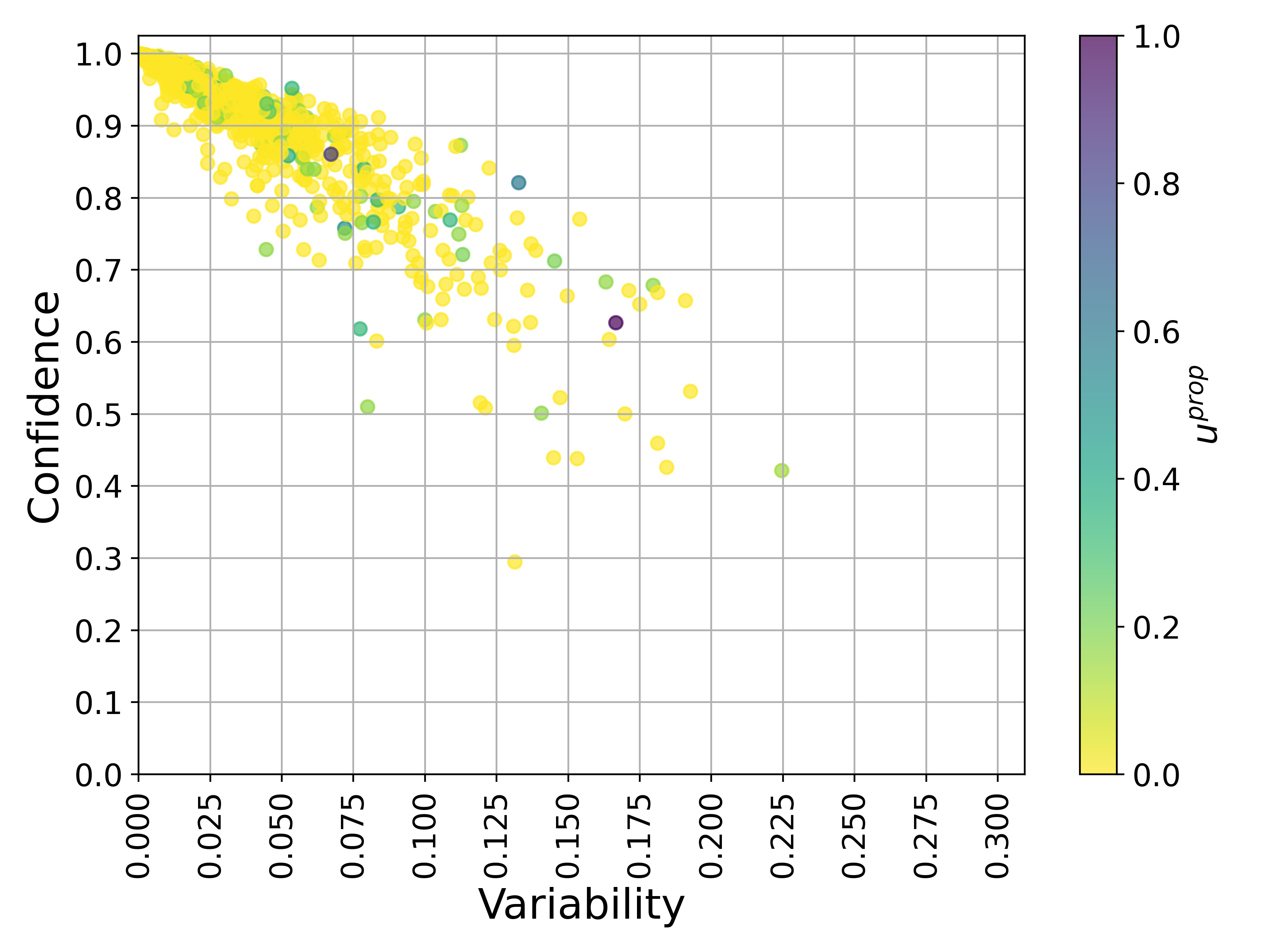}
        \vspace{-1.7em}
        \caption{MNIST - \textit{NoHLV} - $maj._{n}$ }
        \label{fig:row1_right}
    \end{subfigure}
    \begin{subfigure}[b]{0.49\textwidth}
        \centering
        \includegraphics[width=\textwidth]{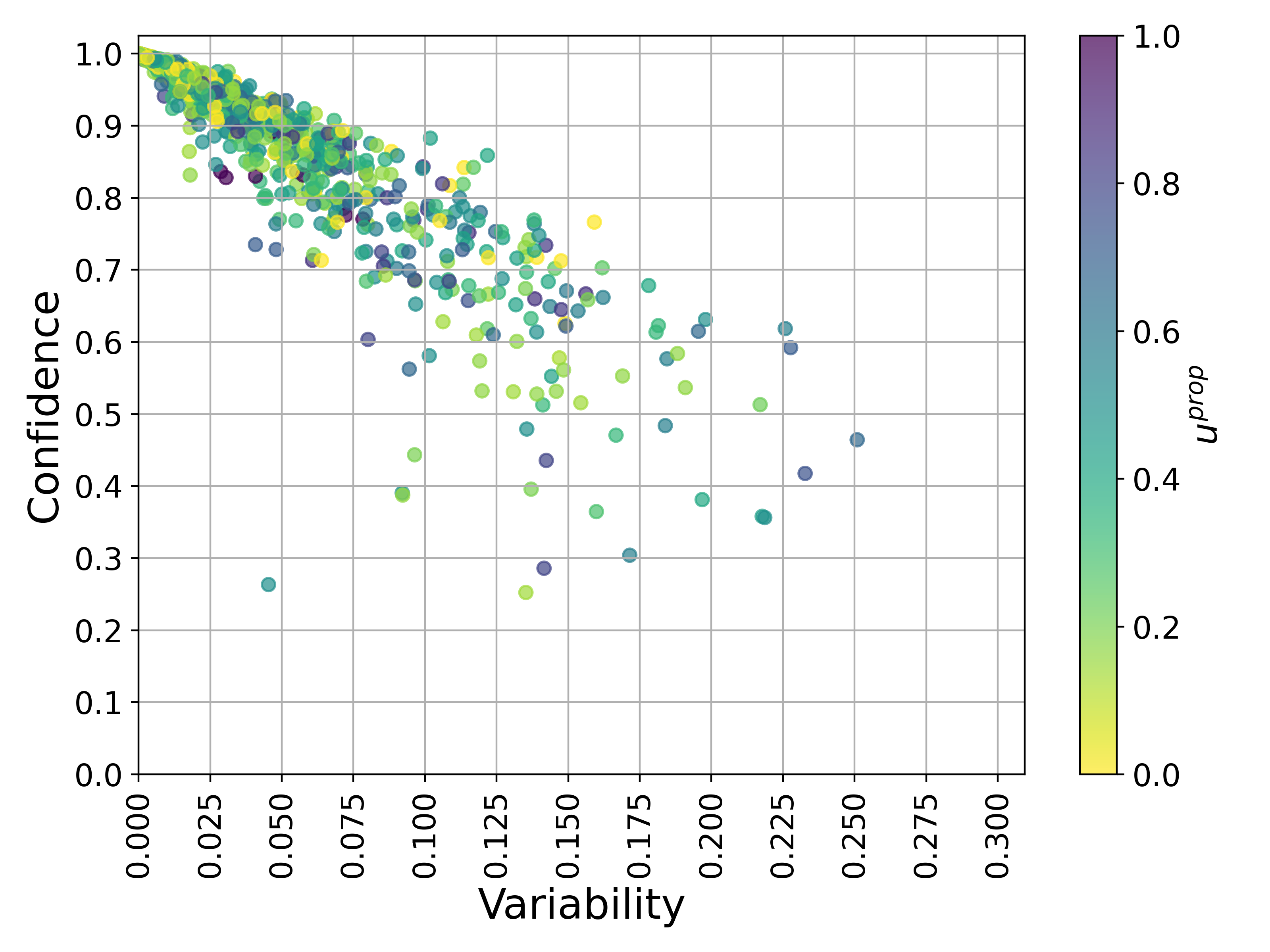}
        \vspace{-1.7em}
        \caption{MNIST - \textit{HLV} - $maj._{n}$}
        \label{fig:row2_right}
    \end{subfigure}

    \vspace{.5ex} 

    \begin{subfigure}[b]{0.49\textwidth}
        \centering
        \includegraphics[width=\textwidth]{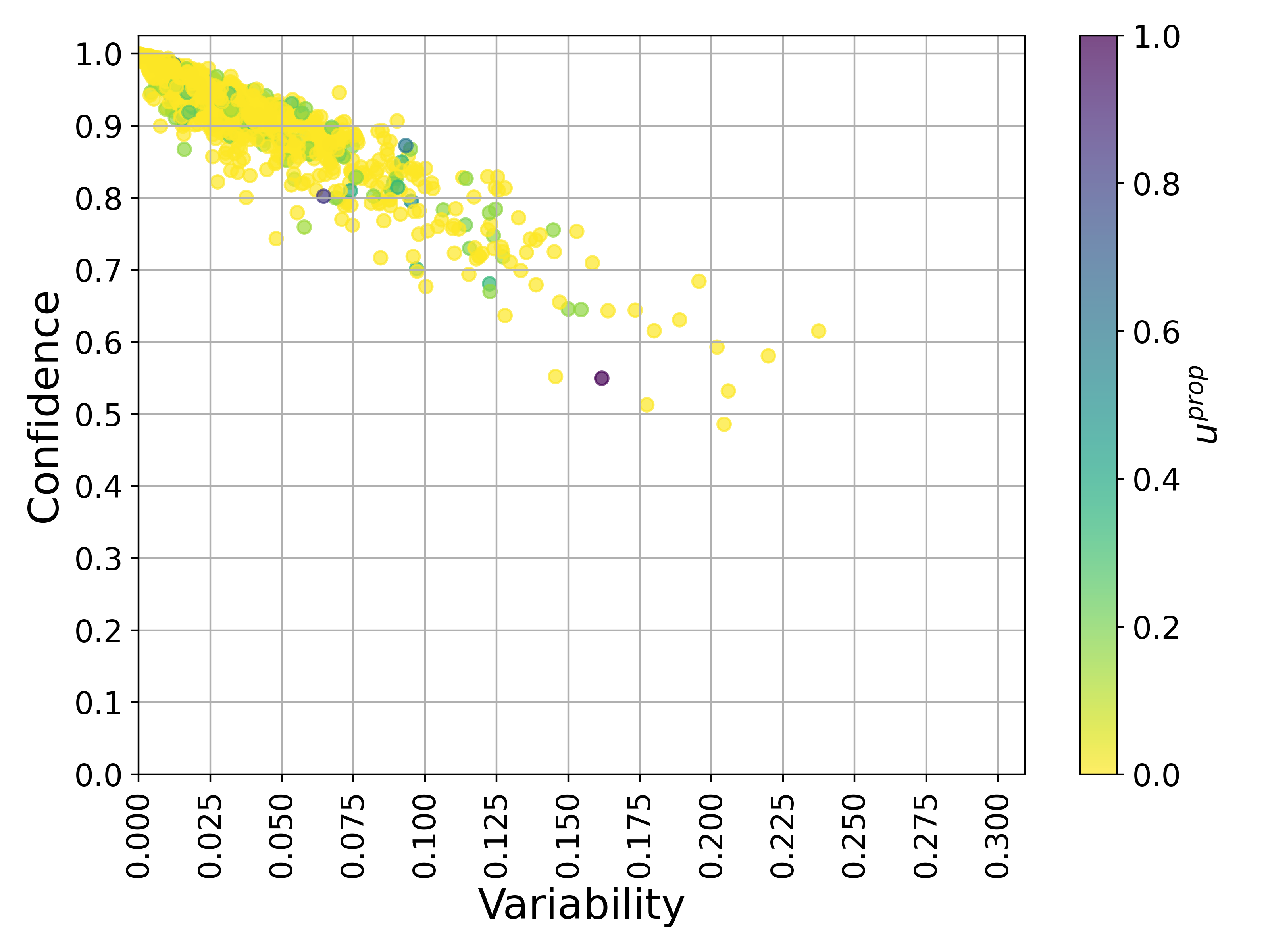}
        \vspace{-1.7em}
        \caption{MNIST - \textit{NoHLV} - $soft_{w}$ }
        \label{fig:row1_left}
    \end{subfigure}
    \begin{subfigure}[b]{0.49\textwidth}
        \centering
        \includegraphics[width=\textwidth]{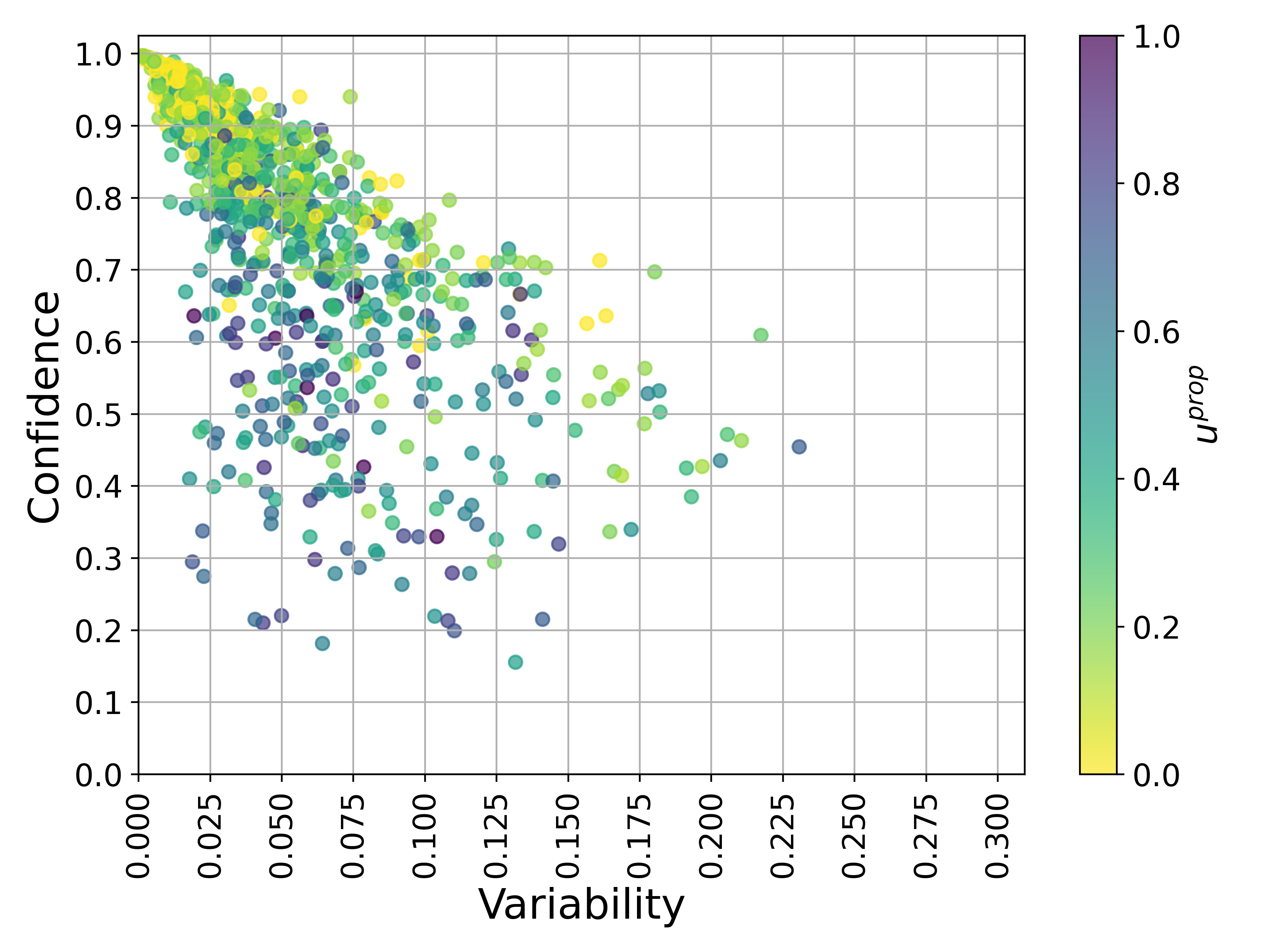}
        \vspace{-1.7em}
        \caption{MNIST - \textit{HLV} - $soft_{w}$}
        \label{fig:row2_left}
    \end{subfigure}

    \caption{
    Late stage training dynamics of SimpleFFN {\small (last 5 epochs)} averaged across 6 random seeds on MNIST data, with $soft_w$ and $maj._n$ contrasted;
    training on soft-labels indicates lower variability in HLV strata compared to training on majority label.}
    \label{fig:mnist_cartography_human_uncert_simple}
\end{figure}

\clearpage
\newpage

\subsection{Human Uncertainty}
\label{app:spearman_umean}
\begin{table}[ht!]
\caption{Spearman correlation between model training dynamics and human uncertainty ($u_{mean}$) across 6 seeds; patterns are consistent with $u_{prop}$. }
\vspace{0.5em}
\centering
\small
\renewcommand{\arraystretch}{1.0} 
\setlength{\tabcolsep}{2.2pt} 
\begin{tabular}{crcccccc}
\toprule
 & & \multicolumn{2}{c}{\textbf{SimpleFFN}} & \multicolumn{2}{c}{\textbf{DeeperFFN}} & \multicolumn{2}{c}{\textbf{LeNet}} \\
\cmidrule(lr){3-4} \cmidrule(lr){5-6} \cmidrule(lr){7-8}
\textbf{eval.} & \textbf{target} & {Conf.(${u_{mean}}$)} & {Var.(${u_{mean}}$)} & {Conf.(${u_{mean}}$)} & {Var.(${u_{mean}}$)}  & {Conf.(${u_{mean}}$)} & {Var.(${u_{mean}}$)} \\ 
\midrule
\multirow{2}{*}{\rotatebox[origin=c]{90}{MNIST}} & $soft_{w}$ & $-0.62_{(p < 0.001)}$ & $0.41_{(p < 0.001)}$ & $-0.71_{(p < 0.001)}$ & $0.59_{(p < 0.001)}$ & $-0.70_{(p < 0.001)}$ & $0.61_{(p < 0.001)}$ \\
& $maj._n$ & $-0.33_{(p < 0.001)}$ & $0.32_{(p < 0.001)}$ & $-0.35_{(p < 0.001)}$ & $0.35_{(p < 0.001)}$ & $-0.46_{(p < 0.001)}$ & $0.45_{(p < 0.001)}$ \\
\midrule
\multirow{3}{*}{\rotatebox[origin=c]{90}{Mukh.}} & $soft_{w}$ & $-0.66_{(p < 0.001)}$ & $0.25_{(p < 0.001)}$ & $-0.72_{(p < 0.001)}$ & $0.37_{(p < 0.001)}$ & $-0.67_{(p < 0.001)}$ & $0.37_{(p < 0.001)}$ \\
& $maj._n$ & $-0.37_{(p < 0.001)}$ & $0.23_{(p < 0.001)}$ & $-0.38_{(p < 0.001)}$ & $0.25_{(p < 0.001)}$ & $-0.44_{(p < 0.001)}$ & $0.35_{(p < 0.001)}$ \\
& $synth.$ & $-0.08_{(p < 0.001)}$ & $0.06_{(p < 0.05)}$ & $-0.10_{(p < 0.001)}$ & $0.08_{(p < 0.01)}$ & $-0.08_{(p < 0.001)}$ & $0.08_{(p < 0.01)}$ \\
\bottomrule
\end{tabular}
\end{table}

\subsection{CIFAR-S cartography}
\label{app:cifar-s}
To validate our findings on further datasets, we repeat our dataset cartography approach on the CIFAR-S dataset \cite{collins_eliciting_2022}. We selected CIFAR-S because its annotation methodology aligns more closely with our own data collection process than alternatives such as CIFAR-10H. Rather than capturing sample uncertainty via a single hard label per participant, annotators for CIFAR-S provided probability estimates for up to two of their most likely choices, while explicitly ruling out any definitively incorrect classes.

Using the resulting human probability distribution for each image, we train models against both the collapsed argmax (majority labels) and the full distribution (soft labels). To ensure our observations hold across different inductive biases and model scales, we evaluate four distinct architectures:
\begin{itemize}
    \item DeeperFFN, a fully-connected baseline with two hidden layers (3072→256→128→10) over flattened pixels.
    \item LeNet-style, a CNN with three 3×3 conv blocks (32→64→128 channels) with max-pooling, followed by two FC layers	
    \item Vision Transformer; a convolutional tokenizer (2 conv+pool blocks) produces 64 patch tokens + a CLS token, processed by 4 self-attention encoder layers
    \item ResNet, a ResNet-20, an initial conv followed by three stages of three residual blocks each (16/32/64 channels) with skip connections, global average pooling
\end{itemize}

Consistent with our primary experiments, models trained on soft-targets show lower variability in confidence across random seeds compared to those trained on majority (argmax) targets.
Furthermore, the cartography mappings mirror some of the observations from our digit dataset experiments. Low-entropy (lower human uncertainty) samples cluster in the high-confidence, low-variability region, while high-entropy samples end up in the low-confidence, low-variability region. With intermediate samples filling the remaining transitional space.
When training on the collapsed argmax targets, this structured mapping degrades: 
The distribution of samples across the cartographic space loses its visible alignment with human uncertainty, becoming more arbitrary. 

\begin{figure}[htbp]
    \centering
    \begin{subfigure}[b]{0.49\textwidth}
        \centering
        \includegraphics[width=\textwidth]{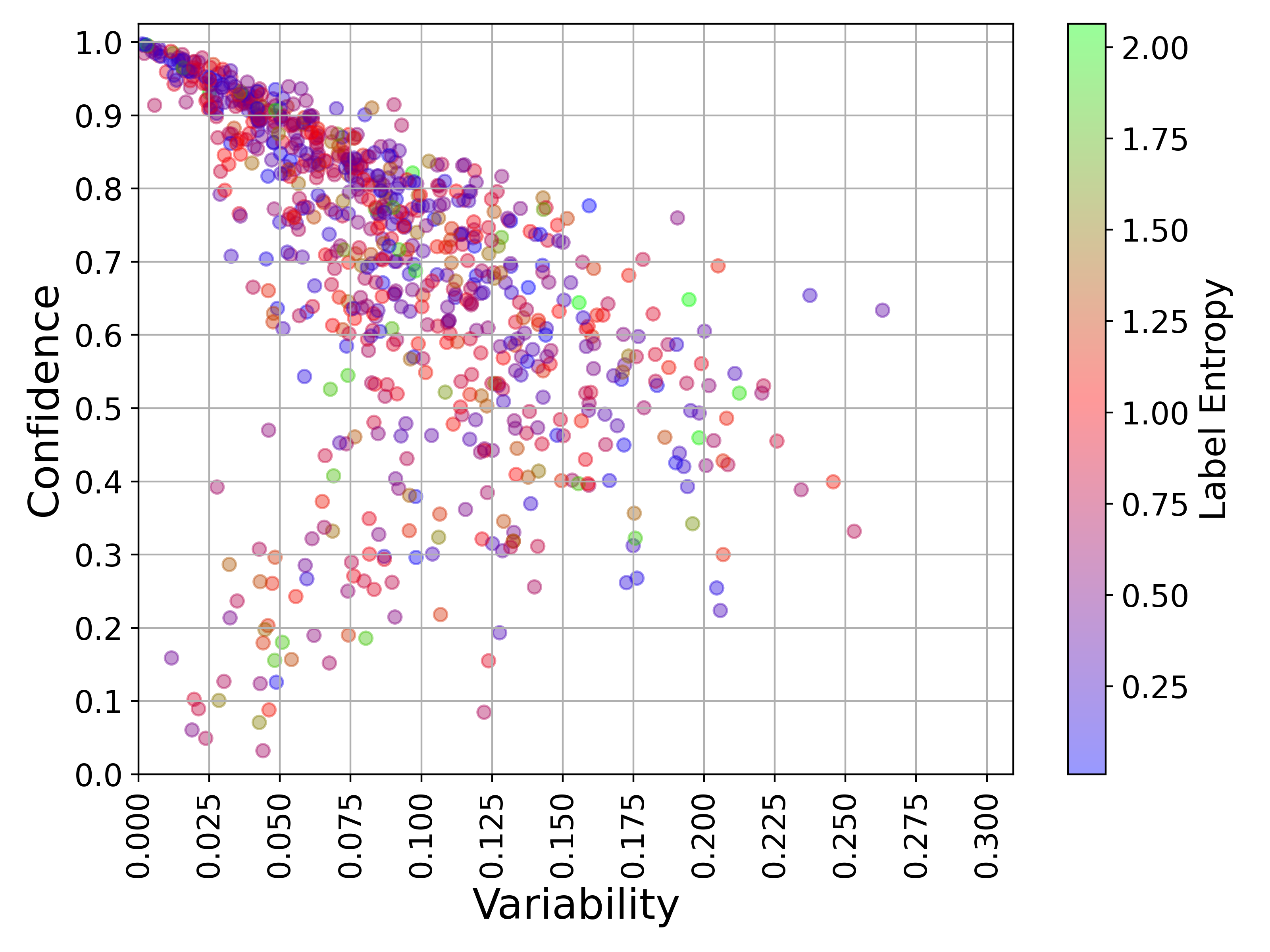}
        \vspace{-1.7em}
        \caption{DeeperFFN - \textit{HLV} - argmax }
        \label{fig:row1_left}
    \end{subfigure}
    \begin{subfigure}[b]{0.49\textwidth}
        \centering
        \includegraphics[width=\textwidth]{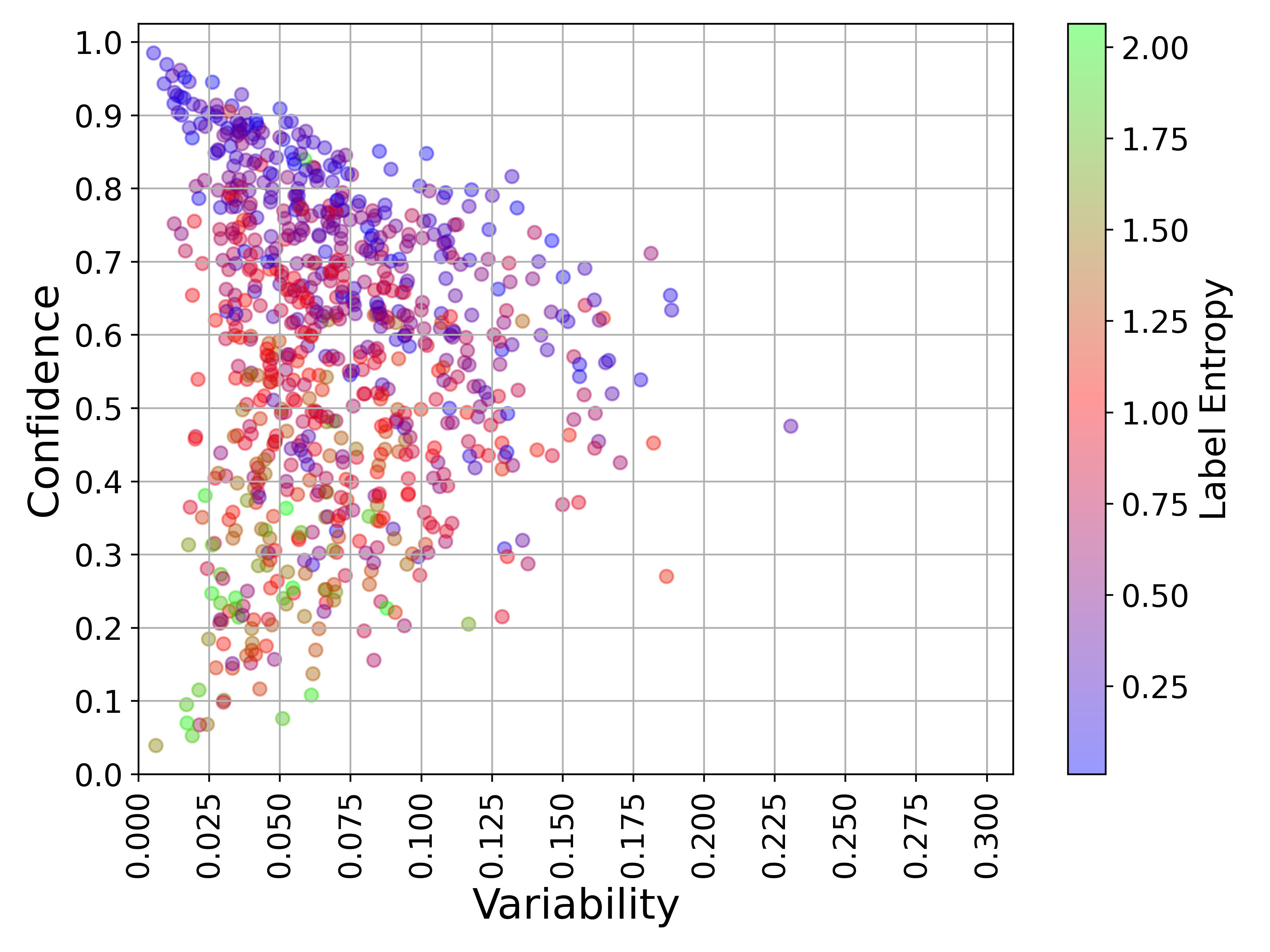}
        \vspace{-1.7em}
        \caption{DeeperFFN - \textit{HLV} - soft }
        \label{fig:row1_right}
    \end{subfigure}

    \vspace{.5ex} 

    \begin{subfigure}[b]{0.49\textwidth}
        \centering
        \includegraphics[width=\textwidth]{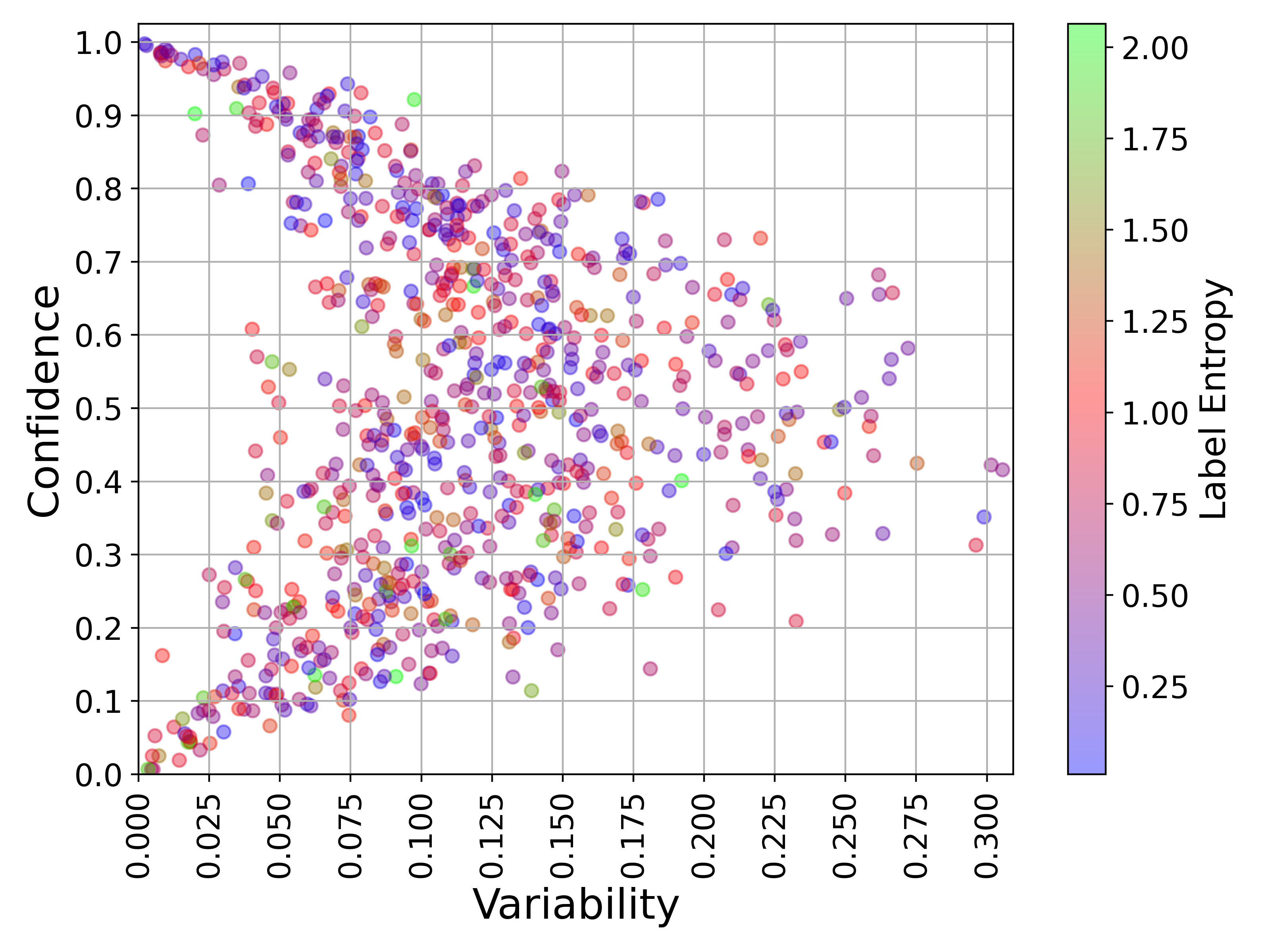}
        \vspace{-1.7em}
        \caption{LeNet - \textit{HLV} - argmax }
        \label{fig:row2_left}
    \end{subfigure}
    \begin{subfigure}[b]{0.49\textwidth}
        \centering
        \includegraphics[width=\textwidth]{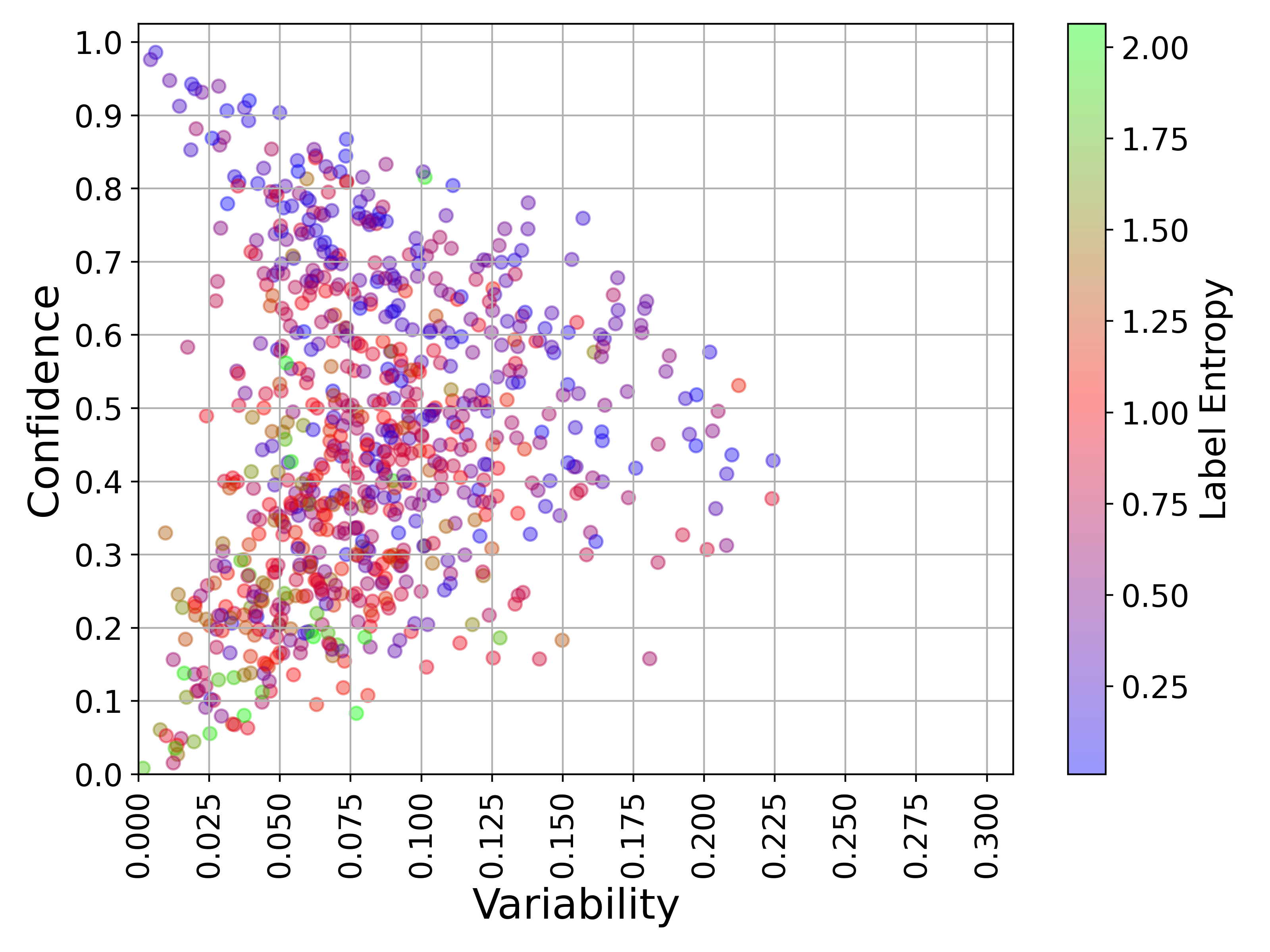}
        \vspace{-1.7em}
        \caption{LeNet - \textit{HLV} - soft }
        \label{fig:row2_right}
    \end{subfigure}

    \vspace{.5ex}

    \begin{subfigure}[b]{0.49\textwidth}
        \centering
        \includegraphics[width=\textwidth]{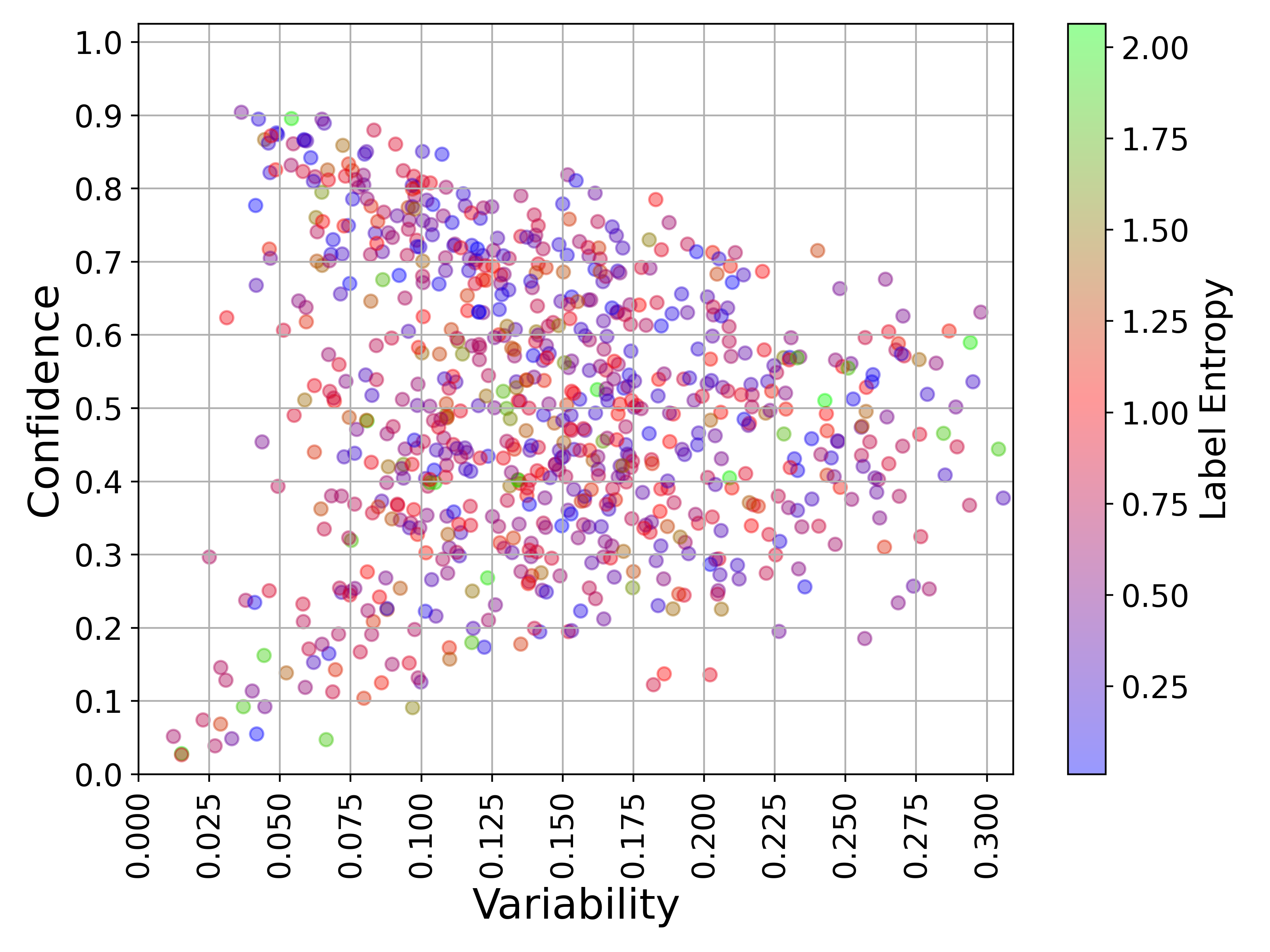}
        \vspace{-1.7em}
        \caption{ViT - \textit{HLV} - argmax }
        \label{fig:row3_left}
    \end{subfigure}
    \begin{subfigure}[b]{0.49\textwidth}
        \centering
        \includegraphics[width=\textwidth]{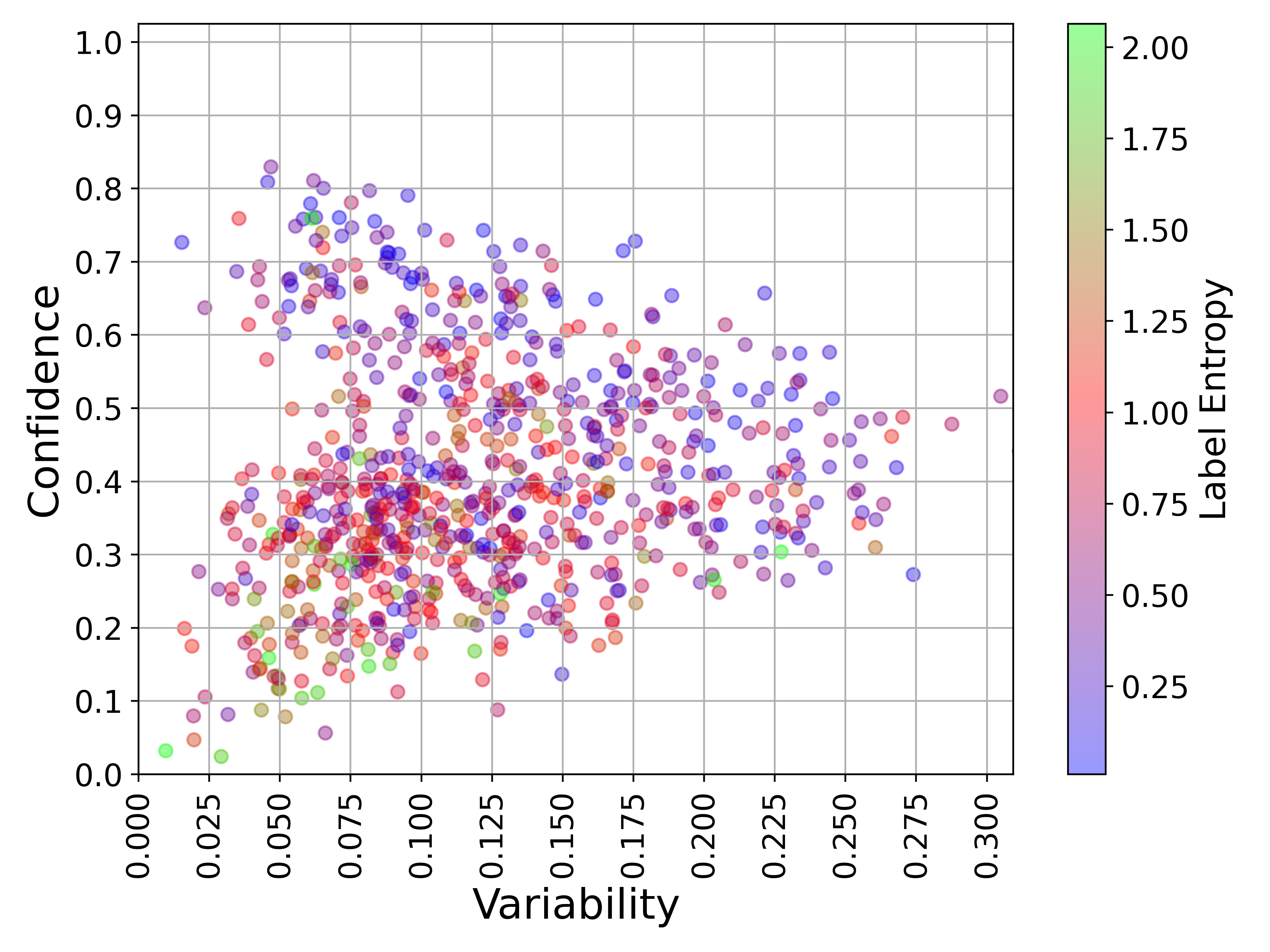}
        \vspace{-1.7em}
        \caption{ViT - \textit{HLV} - soft}
        \label{fig:row3_right}
    \end{subfigure}

    \begin{subfigure}[b]{0.49\textwidth}
        \centering
        \includegraphics[width=\textwidth]{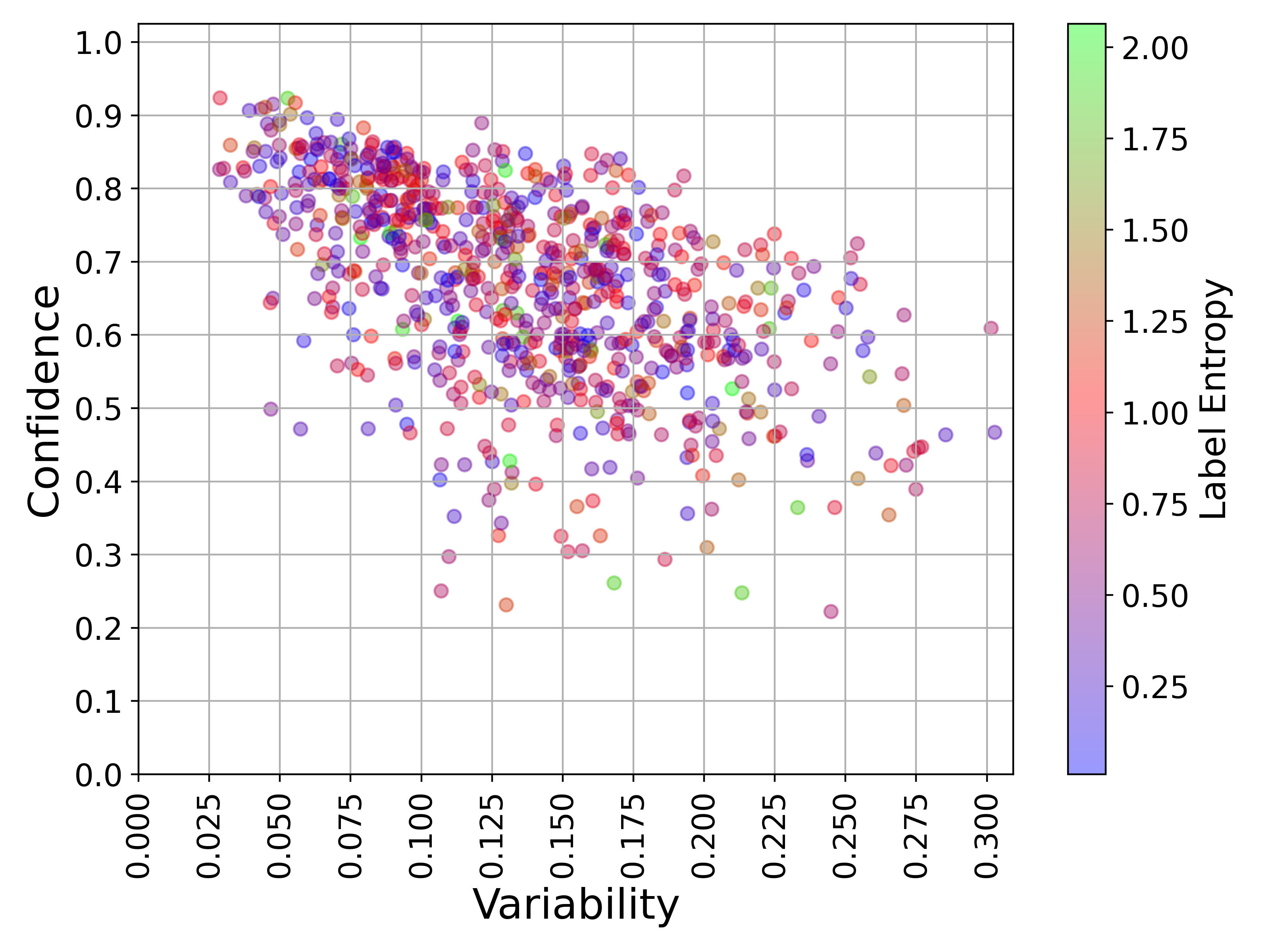}
        \vspace{-1.7em}
        \caption{ResNet - \textit{HLV} - argmax }
        \label{fig:row4_left}
    \end{subfigure}
    \begin{subfigure}[b]{0.49\textwidth}
        \centering
        \includegraphics[width=\textwidth]{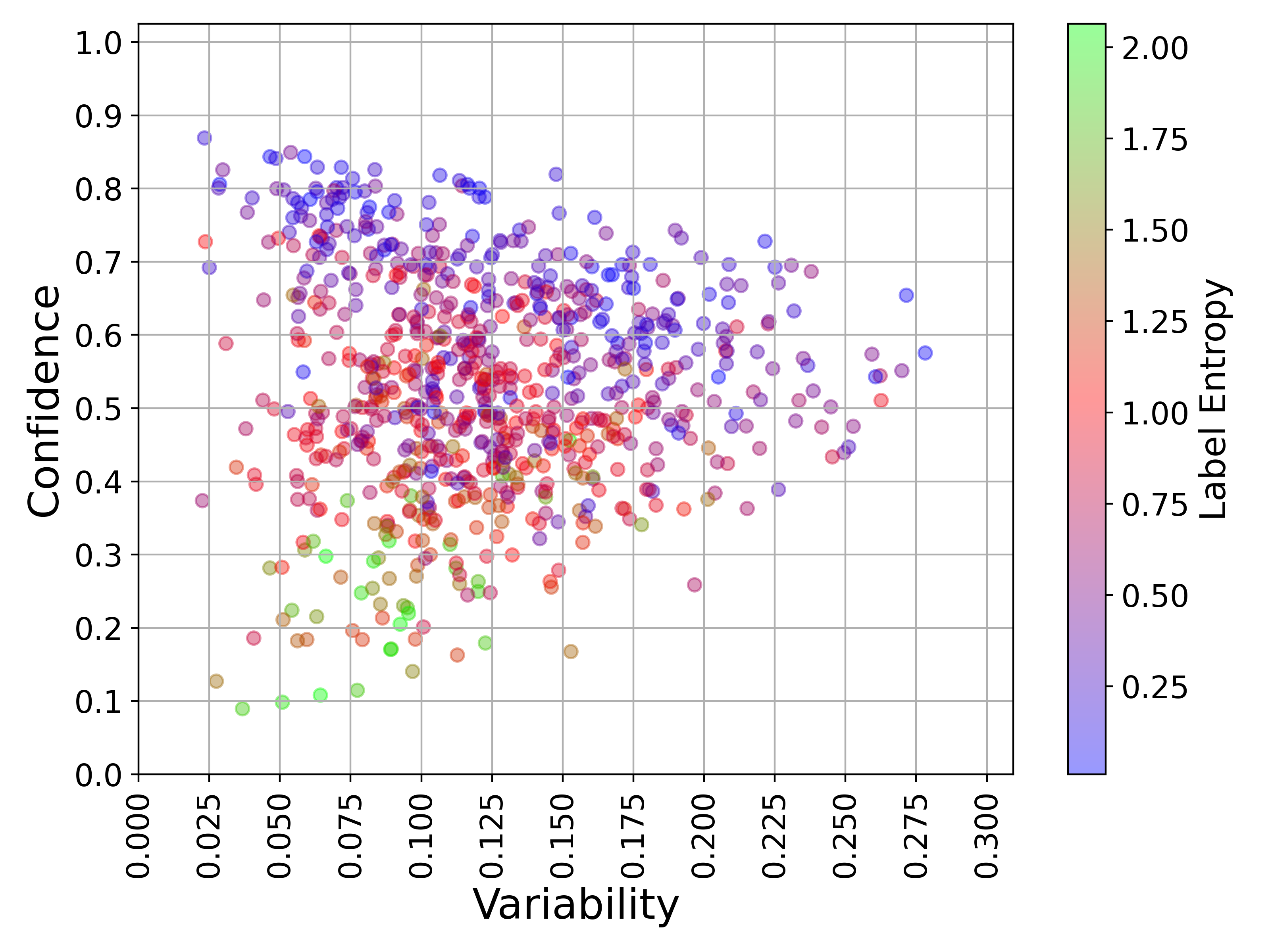}
        \vspace{-1.7em}
        \caption{ResNet - \textit{HLV} - soft }
        \label{fig:row4_right}
    \end{subfigure}

    \caption{CIFAR-S - DeeperFFN, LeNet, ViT - Late stage training dynamics averaged across 6 random seeds, with soft and argmax contrasted.
    }
    \label{fig:maps_cifar}
\end{figure}

\newpage

\section{Datasheet for soft-digits}
\label{app:datasheet}


\begin{figure}[htbp]
  \centering
  \begin{minipage}[b]{0.49\textwidth}
    \centering
    \begin{subfigure}[t]{0.14\textwidth}
        \centering
        \includegraphics[width=\textwidth]{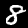}
    \end{subfigure}\hfill
    \begin{subfigure}[t]{0.14\textwidth}
        \centering
        \includegraphics[width=\textwidth]{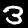}
    \end{subfigure}\hfill
    \begin{subfigure}[t]{0.14\textwidth}
        \centering
        \includegraphics[width=\textwidth]{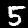}
    \end{subfigure}\hfill
    \begin{subfigure}[t]{0.14\textwidth}
        \centering
        \includegraphics[width=\textwidth]{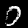}
    \end{subfigure}\hfill
    \begin{subfigure}[t]{0.14\textwidth}
        \centering
        \includegraphics[width=\textwidth]{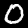}
    \end{subfigure}\hfill
    \begin{subfigure}[t]{0.14\textwidth}
        \centering
        \includegraphics[width=\textwidth]{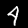}
    \end{subfigure}
    \vspace{-0.4em}
  \end{minipage}%
  \hfill
  \begin{minipage}[b]{0.49\textwidth}
    \centering
    \begin{subfigure}[t]{0.14\textwidth}
        \centering
        \includegraphics[width=\textwidth]{images/sample_digits/mukhoti/noise/low_var_high_conf/3803.png}
    \end{subfigure}\hfill
    \begin{subfigure}[t]{0.14\textwidth}
        \centering
        \includegraphics[width=\textwidth]{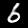}
    \end{subfigure}\hfill
    \begin{subfigure}[t]{0.14\textwidth}
        \centering
        \includegraphics[width=\textwidth]{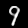}
    \end{subfigure}\hfill
    \begin{subfigure}[t]{0.14\textwidth}
        \centering
        \includegraphics[width=\textwidth]{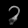}
    \end{subfigure}\hfill
    \begin{subfigure}[t]{0.14\textwidth}
        \centering
        \includegraphics[width=\textwidth]{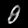}
    \end{subfigure}\hfill
    \begin{subfigure}[t]{0.14\textwidth}
        \centering
        \includegraphics[width=\textwidth]{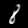}
    \end{subfigure}
    \vspace{-0.4em}
  \end{minipage}

  \vspace{0.7em} 

  \begin{minipage}[b]{0.49\textwidth}
    \centering
    \begin{subfigure}[t]{0.14\textwidth}
        \centering
        \includegraphics[width=\textwidth]{images/sample_digits/mnist/noise/low_var_low_conf/2928.png}
    \end{subfigure}\hfill
    \begin{subfigure}[t]{0.14\textwidth}
        \centering
        \includegraphics[width=\textwidth]{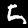}
    \end{subfigure}\hfill
    \begin{subfigure}[t]{0.14\textwidth}
        \centering
        \includegraphics[width=\textwidth]{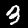}
    \end{subfigure}\hfill
    \begin{subfigure}[t]{0.14\textwidth}
        \centering
        \includegraphics[width=\textwidth]{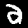}
    \end{subfigure}\hfill
    \begin{subfigure}[t]{0.14\textwidth}
        \centering
        \includegraphics[width=\textwidth]{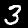}
    \end{subfigure}\hfill
    \begin{subfigure}[t]{0.14\textwidth}
        \centering
        \includegraphics[width=\textwidth]{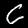}
    \end{subfigure}
    \vspace{-0.4em}
    \caption*{MNIST}
  \end{minipage}%
  \hfill
  \begin{minipage}[b]{0.49\textwidth}
    \centering
    \begin{subfigure}[t]{0.14\textwidth}
        \centering
        \includegraphics[width=\textwidth]{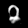}
    \end{subfigure}\hfill
    \begin{subfigure}[t]{0.14\textwidth}
        \centering
        \includegraphics[width=\textwidth]{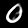}
    \end{subfigure}\hfill
    \begin{subfigure}[t]{0.14\textwidth}
        \centering
        \includegraphics[width=\textwidth]{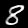}
    \end{subfigure}\hfill
    \begin{subfigure}[t]{0.14\textwidth}
        \centering
        \includegraphics[width=\textwidth]{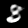}
    \end{subfigure}\hfill
    \begin{subfigure}[t]{0.14\textwidth}
        \centering
        \includegraphics[width=\textwidth]{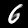}
    \end{subfigure}\hfill
    \begin{subfigure}[t]{0.14\textwidth}
        \centering
        \includegraphics[width=\textwidth]{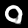}
    \end{subfigure}
    \vspace{-0.4em}
    \caption*{Mukhoti - ambiguous MNIST}
  \end{minipage}
  \caption{Samples from MNIST and Mukhoti (ambig. MNIST) for the soft-label digits}
\end{figure}

\subsection{Motivation}
The dataset was specifically developed to provide a controlled testbed for investigating the dynamics between human perceptual uncertainty and model-based uncertainty. By including human-level uncertainty for each class, the dataset enables a more detailed analysis of how machine learning models align with or diverge from human judgment in ambiguous scenarios.

It benefits researchers and students by providing a low-compute entry point for exploring model uncertainty, specifically regarding ambiguous edge-case recognition and noise handling.

High performance on this controlled dataset is not an indication for deployment readiness. To ensure fairness and transparency, the origins of the re-annotated source data are documented to communicate inherent geographic biases to users.

The dataset is explicitly communicated as an evaluation testbed rather than a production-grade training set.

Information regarding the authors and funding for this dataset has been omitted to preserve the anonymity of the review process.

\subsection{Composition}
Each instance in the dataset consists of a grayscale image of a handwritten digit (0–9) with a resolution of 28x28 pixels. There are 5,530 instances in the dataset. The dataset is a curated sample constructed from a combination of two existing public datasets. The selection process was designed to capture samples across all difficulties; a detailed description of the sampling methodology can be found in the main text of the accompanying paper.  

Each instance is associated with four distinct labels: 
\begin{itemize}
    \item Original Labels: The initial labels associated with the source data. For the subset derived from Mukhoti \cite{mukhoti_2023_ddu}, these are synthetically generated target distributions.
    \item Human Soft-Labels ($soft_e$): Probability distributions over the digit classes derived from human annotators.
    \item Human Soft-Labels ($soft_w$): An alternative set of human-derived soft-labels, giving less weight to the annotators uncertainty selections.
    \item New Majority Voted Labels ($maj.n$): A one-hot encoded label representing the class that received the highest consensus from the human annotators, derived directly from the soft-label distributions.
\end{itemize}

Detailed description of exact information accompanying each image:
\begin{itemize}
    \item {images:} $list\; size([1, 28, 28])$ - containing the grayscale image
    \item {original\_labels:} $list$ - containing the $original$ label
    \item {indices:} $int$ - indices
    \item {file\_name:} $str$ - containing file name (e.g. \textit{1.png})
    \item {human\_uncert\_mean:} $double$ - containing "mean uncertainty" ($u_{mean}$): the average magnitude of doubt across all annotators for an image
    \item {pct\_ann\_unsure:} $double$ - containing "unsure proportion" ($u_{prop}$): the fraction of annotators who expressed uncertainty at least once for a given image, capturing the consensus of doubt
    \item {soft\_label\_yes\_unc\_equal:} $list$ of $floats$ - equally-weighted ($soft_e$): Both "Yes" and "Unsure" selections receive a weight of 1 to form each annotator’s individual probability distribution. These individual distributions are then averaged to create the final image-level soft-label
    \item {soft\_label:} $list$ of $floats$ - uncertainty-weighted ($soft_w$):  "Unsure" selections are down-weighted to 0.5 at the annotator level to reflect a lower class-likelihood, while "Yes" remains 1, before averaging the distributions.
    \item {soft\_label\_argmax:} $list$ of $floats$ - new majority voted labels ($maj._n$): a one-hot encoded label representing the class that received the highest consensus from the human annotators, derived directly from the soft-label distributions 
    \item {split:} $string$ - containing the data split either "train", "val" or "test"
    \item {source:} $string$  - containing the origin source, either "mnist" or "mukhoti"
\end{itemize}

\subsection{Collection Process}
We explored four openly available digit datasets to identify samples that represent a broad spectrum of classification difficulty. We explored the standard MNIST \cite{lecun2002gradient}, the Swedish historical handwritten dataset ARDIS \cite{kusetogullari2020ardis}, and two synthetic variants of Ambiguous-MNIST \cite{mukhoti_2023_ddu, weiss2023generating}.
We first de-duplicated the datasets to ensure evaluation integrity. Following de-duplication, we performed cartography mapping using a simple one-hidden-layer feed-forward network. Guided by these steps, we selected MNIST and Mukhoti's distributional Ambiguous-MNIST as our primary sources. MNIST was chosen over ARDIS because its larger scale (70k vs 7,474 images) provides a richer tail of naturally occurring hard and ambiguous cases. Between the synthetic datasets, Mukhoti offered a more comprehensive distribution across the difficulty spectrum and better adheres to the natural data manifold compared to Weiss's variant. The resulting datasets were partitioned into training, validation, and test sets using stratified sampling to ensure the difficulty distribution remains consistent across splits. We enforced a constraint of at least 150 'easy' instances per digit in the training set to maintain sufficient class representation.

Our curated dataset inherits the same biases from its sources:
\begin{itemize}
    \item MNIST: shows demographic and geographic skew, sourced from U.S. Census employees and high school students. This results in a Western-centric handwriting bias and lacks global stylistic variations. As a "clean" curated subset of the original NIST databases, it under-represents the scale variance, noise and ambiguity found in real-world samples.
    \item Mukhoti: as a synthetically generated dataset designed to introduce ambiguity, it uses MNIST as a structural baseline. Therefore, inheriting the underlying Western-centric skews of the original MNIST data.
\end{itemize}
We source annotators from the US, France and Germany to re-annotate the subsets. Thus, our dataset reflects a highly Western-centric distribution of handwriting and interpretation, resulting in a model that is optimized for American and European understanding but lacks the diversity to generalize to global variations and real-world environments.

\subsection{Preprocessing/Cleaning/Labeling}
No extensive pre-processing or cleaning was performed on the human annotations. This was an intentional design choice to preserve the natural variance in annotations, thereby the release of the dataset allows for further research into separating informative human label variation from noise. Regarding data integrity, we have retained the vast majority of collected annotations to maintain a representative distribution. We only excluded a small subset of samples from a single annotator who reported a hardware display failure during their session.

\subsection{Distribution \& Licensing}

The soft-digits dataset will be distributed via Hugging Face and is a derivative of: 
\begin{itemize}
    \item the MNIST dataset (http://yann.lecun.com/exdb/mnist/), originally created by Yann LeCun and Corinna Cortes - (released under CC-BY-SA 3.0)
    \item Ambiguous-MNIST dataset (https://github.com/omegafragger/DDU), originally created by Jishnu Mukhoti, Andreas Kirsch, Joost van Amersfoort, Philip H.S. Torr, and Yarin Gal -  (released under MIT license)
\end{itemize}
The soft-digits dataset is released under the CC-BY-SA 4.0 license. The components derived from \cite{mukhoti_2023_ddu} remain subject to the terms of the MIT License.

\subsection{Maintenance}
The soft-digits dataset is hosted on the Hugging Face Hub, which serves as the primary platform for its distribution and long-term maintenance. The corresponding author is responsible for managing the repository, ensuring the data remains accessible, and performing any necessary updates or versioning.

Error Reporting and Community Feedback: To enable continuous improvement, we use the native infrastructure provided by Hugging Face. Users can report errors or suggest corrections through the "Discussions" tab of the dataset’s repository.


\end{document}